\documentclass{article} %
\usepackage{iclr2027_conference,times}

\usepackage{amsmath,amsfonts,bm}

\def\eqref#1{equation~\ref{#1}}

\def\1{\bm{1}}

\DeclareMathAlphabet{\mathsfit}{\encodingdefault}{\sfdefault}{m}{sl}
\SetMathAlphabet{\mathsfit}{bold}{\encodingdefault}{\sfdefault}{bx}{n}

\usepackage{float} %
\usepackage{hyperref}
\usepackage{url}
\usepackage{amsmath}
\usepackage{amssymb}
\usepackage{amsthm}
\usepackage{booktabs}
\usepackage{bm}
\usepackage{algorithm}
\usepackage{algorithmic}
\usepackage{enumitem}
\usepackage{graphicx}
\usepackage{makecell}
\usepackage{xcolor}

\newtheorem{theorem}{Theorem}
\newtheorem{lemma}{Lemma}
\newtheorem{proposition}{Proposition}
\newtheorem{corollary}{Corollary}
\newtheorem{definition}{Definition}
\newtheorem{remark}{Remark}

\newcommand{\envGPU}{NVIDIA~A100-SXM4-40GB}
\newcommand{\envGPUMem}{40{,}960\,MiB}
\newcommand{\envPython}{3.13.16}
\newcommand{\envTorch}{2.14.1{+}cu130}
\newcommand{\envCuda}{13.0}

\newcommand{\famSeeds}{5}
\newcommand{\famNThree}{800}

\newcommand{\famNMax}{1000}
\newcommand{\famSizesThree}{100,200,300,500,800,1000}
\newcommand{\famSizesFour}{20,40,60,80}
\newcommand{\famSmallDataThree}{7.63}

\newcommand{\famData}{3{,}906}

\newcommand{\famDataMaxGiB}{7.45}
\newcommand{\famFailNames}{naive HOOI, TensorLy, HOSVD and ST-HOSVD}

\newcommand{\famBestExistingName}{HOQRI}
\newcommand{\famSpeedBest}{1.12}
\newcommand{\famSpeedNaive}{15}
\newcommand{\famSpeedTensorly}{13}

\newcommand{\famWLargest}{95}
\newcommand{\famWMax}{121}

\newcommand{\famWLargestFour}{25}

\newcommand{\famWFirstThree}{1.2}
\newcommand{\famWFirstFour}{1.9}
\newcommand{\famNMinThree}{100}

\newcommand{\famTSlowRangeThree}{0.65--0.85}
\newcommand{\famTSlowMaxNThree}{500}
\newcommand{\famTSlowRangeFour}{0.37--0.55}

\newcommand{\famSlowThreeNames}{HOQRI}

\newcommand{\famMinErrRatioFaster}{7.6}

\newcommand{\famKryInitPeak}{5{,}614}
\newcommand{\famKryInitTime}{1.3}
\newcommand{\famCopyInitPeak}{7{,}848}
\newcommand{\famTreeGain}{1.26}
\newcommand{\famTreeGainFour}{1.07}
\newcommand{\famAllKryTime}{2.0}
\newcommand{\famWorkOurs}{41.4}
\newcommand{\famWorkBestMB}{3{,}921}
\newcommand{\famWorkOursMax}{63.2}
\newcommand{\famWorkBestMaxGiB}{7.47}

\newcommand{\benchSeedsMax}{5}
\newcommand{\benchNmax}{500}
\newcommand{\benchSizes}{100,200,300,500}
\newcommand{\rZeroT}{3.4}
\newcommand{\rZeroM}{4.0}

\newcommand{\rThreeT}{2.4}
\newcommand{\rThreeM}{2.0}

\newcommand{\rFourLowT}{3.7}
\newcommand{\rFourLowM}{8.4}

\newcommand{\rFourHighT}{1.3}
\newcommand{\rFourHighM}{2.3}

\newcommand{\rOneHighT}{1.2}
\newcommand{\rOneHighM}{2.3}

\newcommand{\rRawTwoM}{6.5}
\newcommand{\rRawFourM}{7.2}

\newcommand{\rRawZeroM}{4.0}

\newcommand{\rZeroFasterFrom}{200}

\newcommand{\rFourLowFasterFrom}{300}

\newcommand{\rFourHighFasterFrom}{500}

\newcommand{\rSlowSmallMax}{1.4}
\newcommand{\benchNmin}{100}
\newcommand{\gpuAgreeBench}{1.0\times10^{-12}}
\newcommand{\gpuAgreeBenchRuns}{120}
\newcommand{\gpuAgreeFam}{7.3\times10^{-9}}

\newcommand{\exactHOOI}{3.7\times10^{-15}}
\newcommand{\exactTight}{5.2\times10^{-14}}
\newcommand{\exactDefault}{9.5\times10^{-11}}

\newcommand{\slRawHiLo}{3.25}
\newcommand{\slRawHiHi}{3.28}
\newcommand{\slRawLoLo}{2.2}
\newcommand{\slRawLoHi}{2.3}

\newcommand{\slTermErr}{4}
\newcommand{\slShareLo}{49}
\newcommand{\slShareHi}{55}
\newcommand{\slNstarLo}{555}
\newcommand{\slNstarHi}{722}
\newcommand{\slGapFWLo}{0.62}
\newcommand{\slGapFWHi}{1.34}
\newcommand{\slOursLo}{1.32}
\newcommand{\slOursHi}{2.17}
\newcommand{\slOtherRawLo}{2.54}
\newcommand{\slOtherRawHi}{2.94}
\newcommand{\slHooiOursMs}{11}
\newcommand{\slCpuRawLo}{3.42}
\newcommand{\slCpuRawHi}{3.50}
\newcommand{\slLiteOneLowOurs}{3.10}
\newcommand{\slLiteOneLowRaw}{3.19}
\newcommand{\slLiteOneHighOurs}{1.14}
\newcommand{\slLiteOneHighRaw}{3.14}

\newcommand{\xoRuns}{120}
\newcommand{\xoSeeds}{3}
\newcommand{\xoNmax}{500}
\newcommand{\xoNstar}{200}
\newcommand{\xoNsmall}{100}
\newcommand{\xoFast}{32}
\newcommand{\xoFastFew}{30}
\newcommand{\xoSnMax}{0.05}
\newcommand{\xoGainLo}{4.4}
\newcommand{\xoGainHi}{12.7}
\newcommand{\xoSlowLo}{10}
\newcommand{\xoSlowHi}{51}
\newcommand{\xoMemFewLo}{5.5}
\newcommand{\xoMemFewHi}{8.2}
\newcommand{\xoMemHalfLo}{1.0}
\newcommand{\xoMemHalfHi}{1.1}
\newcommand{\xoSmallSlowLo}{6}
\newcommand{\xoSmallSlowHi}{16}
\newcommand{\xoAlgFixedLo}{9}
\newcommand{\xoAlgFixedHi}{22}
\newcommand{\xoSvdSmallLo}{1.7}
\newcommand{\xoSvdSmallHi}{9.5}

\newcommand{\mxRuns}{1{,}040}

\newcommand{\mxScored}{16}
\newcommand{\mxAfail}{6}
\newcommand{\mxHits}{8}

\newcommand{\mxNoGainCells}{28}
\newcommand{\mxNoGainFast}{0}
\newcommand{\mxGainCells}{30}
\newcommand{\mxGainFast}{11}
\newcommand{\mxExactRel}{5.5\times10^{-9}}
\newcommand{\mxExactAbs}{1.6\times10^{-12}}

\newcommand{\mxWordsObsExt}{15}
\newcommand{\mxWordsObs}{15}
\newcommand{\mxSparseMemLo}{2.9}
\newcommand{\mxSparseMemHi}{5.8}
\newcommand{\mxEtenMem}{0.93}
\newcommand{\mxPrgdMemLo}{1.7}
\newcommand{\mxPrgdMemHi}{5.8}

\newcommand{\mxRgSlowLo}{1.9}
\newcommand{\mxRgSlowHi}{2.6}
\newcommand{\mxTihtLo}{7.9}
\newcommand{\mxTihtHi}{13.8}
\newcommand{\mxHighQqLo}{0.81}
\newcommand{\mxHighQqHi}{1.00}
\newcommand{\mxHighQLo}{54}
\newcommand{\mxHighQHi}{62}
\newcommand{\mxAdmmErrLo}{0.72}
\newcommand{\mxAdmmErrHi}{0.95}

\newcommand{\mxSiqLo}{0.07}
\newcommand{\mxSiqHi}{0.09}
\newcommand{\mxNest}{eight}
\newcommand{\mxNestCap}{Eight}
\newcommand{\mxSparseN}{five}
\newcommand{\edgeFamTTBSlow}{2.6}

\newcommand{\edgeCtx}{0.52}
\newcommand{\edgeRedPubMin}{1.55}
\newcommand{\edgeRedPubMax}{4.83}
\newcommand{\edgeRedRawMin}{2.64}
\newcommand{\edgeRedRawMax}{5.98}
\newcommand{\edgeOtherPubMin}{1.52}
\newcommand{\edgeOtherPubMax}{6.47}
\newcommand{\edgeOtherRawMin}{2.54}
\newcommand{\edgeOtherRawMax}{6.86}
\newcommand{\edgeFWSlow}{32}

\newcommand{\edgeSepHall}{1}
\newcommand{\edgeSepHypso}{4}
\newcommand{\edgeFWPubPSNR}{21.56}
\newcommand{\edgeDenseGramPeak}{2.18}
\newcommand{\edgeDenseGramMs}{46}
\newcommand{\edgeAntPeakDiff}{0.3}
\newcommand{\edgeAntPSNRThree}{34.00}
\newcommand{\edgeAntPSNRZero}{34.84}
\newcommand{\edgeImgDiff}{8.8\times10^{-13}}

\newcommand{\edgeSameEstOther}{2.1\times10^{-12}}
\newcommand{\edgeSameEstFW}{4.6\times10^{-7}}

\title{Lightweight Representation Learning for Low Tucker Rank Tensors}

\author{%
Hiroki Hasegawa\\
Graduate School of Science and Technology, University of Tsukuba\\
Tsukuba, Japan\\
\texttt{hasegawa.hiroki.tkb\_en@u.tsukuba.ac.jp}\\[1em]
Yukihiko Okada\\
Center for Artificial Intelligence Research, Tsukuba Institute for Advanced Research,\\
University of Tsukuba, Tsukuba, Japan\\
Institute of Systems and Information Engineering, University of Tsukuba,\\
Tsukuba, Japan
}

\iclrfinalcopy
\begin{document}

\maketitle
\lhead{Preprint. Under review.}

\begin{abstract}
Iterative estimators for low Tucker rank tensors, such as HOOI, nuclear-norm proximal and ADMM methods, Frank--Wolfe and scaled gradient descent, form large objects in every iteration: unfoldings, Kronecker operators, full SVDs and ambient iterates. The update uses little of these objects, yet they make memory the limiting resource on large tensors and on edge-sized devices. Remedies for each of these costs are known. However, it is unclear when a rewriting keeps the iterates unchanged, and where it pays off against efficient implementations. We therefore present a \emph{trajectory-preserving lightweight evaluation}. First, we state exactness as an intertwining relation between the stored state and the reference iteration. Second, we show that three known components satisfy this relation for five estimators: computing only the singular triplets that an update needs, contracting without Kronecker products, and storing sparse-plus-factored states. As a result, the iterates and all guarantees of the reference estimator stay the same. On an A100 GPU, with the input data counted in the memory, lightweight HOOI needs the least memory of 15 HOOI-type methods at every size ($\famWLargest\times$ less beyond the input than the lightest exact implementation on $\famNThree^3$ tensors), and it is the fastest method that reaches the HOOI estimate on the largest tensors. Against efficient public code, ours is faster and lighter for three of five estimators. On a surveillance video and a satellite hyperspectral cube, ours is the only implementation that fits a \edgeSepHall\,GiB and a \edgeSepHypso\,GiB device. Finally, we show that the memory and time gains need separate conditions, and we identify where these conditions fail.
\end{abstract}

\section{Introduction}
\label{sec:intro}
Low Tucker rank tensors model multiway data such as video, spatio-temporal sensor records and neuroimaging. Tensor completion (partial sampling) and denoising (full sampling with noise) recover such a tensor by estimating mode-wise orthonormal bases and a core tensor \citep{delathauwer2000multilinear, kolda2009tensor, lee2020tensor}. The estimators for this task are iterative: higher-order orthogonal iteration (HOOI) \citep{de2000best}, proximal and alternating direction method of multipliers (ADMM) methods for nuclear-norm relaxations \citep{liu2012tensor, zheng2015interpolating}, Frank--Wolfe \citep{guo2017efficient} and factored gradient methods \citep{tong2022scaling}. Their updates are short to state. However, a literal implementation of one update does far more work than the update needs. It forms each mode-$k$ unfolding, multiplies the unfolding by an explicit Kronecker product of factors, computes a full singular value decomposition (SVD), and stores ambient iterates of size $\prod_k n_k$. Yet the update uses only a few singular triplets, a few contracted quantities and the factors.

\begin{figure}[t]
\centering
\includegraphics[width=\linewidth]{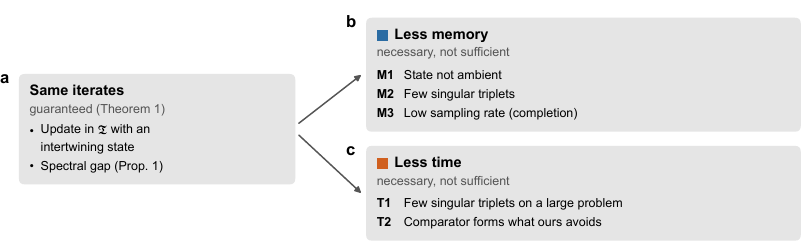}
\caption{What the lightweight evaluation needs. \textbf{a},~The iterates are unchanged under these conditions (Theorem~\ref{thm:exact_equivalence}). \textbf{b},~\textbf{c},~Necessary, not sufficient, conditions for less memory and for less time than the lightest or fastest implementation of the same iterates (Section~\ref{sec:conclusion}; tested on further estimators in Appendix~\ref{sec:appendix_mapext}).}
\label{fig:map}
\end{figure}

Each of these redundancies has a known remedy. Krylov and randomized methods compute partial SVDs \citep{lehoucq1998arpack, halko2011finding}; chains of mode products and the reuse of partial contractions avoid Kronecker products \citep{kolda2008scalable, vannieuwenhoven2012new, kaya2016high}; and factored or sparse-plus-low-rank states avoid ambient iterates \citep{mazumder2010spectral, bader2007efficient, yao2022low}. Efficient libraries such as TensorLy \citep{kossaifi2019tensorly} and the Tensor Toolbox \citep{bader2023tensortoolbox} already use some of these remedies, and randomized and sketched methods go further by changing the computed quantity \citep{minster2020randomized, malik2018low, sun2020lowrank, jang2020dtucker}. However, two things are missing. First, no condition tells, for a given estimator, when such a rewriting keeps the sequence of iterates unchanged; only then do all statistical and convergence guarantees carry over. Second, it is unclear where the rewriting saves time or memory against \emph{efficient} implementations, rather than against a literal implementation, once the input data is counted. Figure~\ref{fig:map} summarizes our answer to both questions.

\paragraph{Contributions.} The numerical components are known, and we cite each where we use it. Our contributions are the following.
\begin{enumerate}[leftmargin=*,itemsep=1pt,topsep=2pt]
\item \emph{A correctness condition} (Section~\ref{sec:method}). We state an intertwining relation between the stored state and the reference iteration. Three components satisfy this relation for five estimators, and for \mxNest{} more in Appendix~\ref{sec:appendix_mapext}. Therefore, the iterates and every guarantee of the reference estimator stay unchanged (Theorem~\ref{thm:exact_equivalence}).
\item \emph{One configuration.} We use one configuration per estimator and one rule for when Krylov extraction is used. Time and memory always come from the same runs.
\item \emph{A full comparison on HOOI} (Section~\ref{sec:hooi}). Among 15 HOOI-type methods on a GPU, lightweight HOOI computes the same iterates as naive HOOI and needs the least memory at every size. It is the fastest method on the largest tensors, but HOQRI is faster on small ones.
\item \emph{Where it pays off} (Sections~\ref{sec:five} and~\ref{sec:real}). Against public implementations of five estimators, ours is faster and lighter for three estimators on large problems, and lighter but slower on small ones; we give the cause of every case in which ours does not help. On two real data sets, only ours fits edge-sized devices.
\end{enumerate}
The learned representation is the Tucker representation of the reference estimator: mode-wise factors and a core tensor, which completion and denoising use directly. We propose no new representation model or objective: ``lightweight'' refers to how we compute the same representation (Theorem~\ref{thm:exact_equivalence}), at lower cost only where the conditions of Figure~\ref{fig:map} hold.

\section{Related Work}
\label{sec:related}

Computational accelerations of low-rank tensor estimation fall into five strategies (Table~\ref{tab:taxonomy_efficiency}). Strategies~I--III reduce the cost of a fixed iteration: they extract only the singular directions that an update needs, keep factor or sparse-plus-low-rank states instead of ambient arrays, or avoid forming large intermediates. In contrast, Strategies~IV and V change the iteration itself. Randomized and sketched methods trade accuracy for speed, and preconditioned or Riemannian methods reduce the iteration count. Statistical limits and sample complexity are studied separately \citep{tomioka2011statistical, mu2014square, zhang2018tensor, xia2021statistically}. We combine Strategies~I--III and keep the iteration unchanged.

\begin{table}[t]
\caption{Strategies for reducing the cost of low-rank tensor estimation. This work combines I--III and leaves the iteration unchanged. In contrast, IV changes the computed quantity, and V changes the iteration count. LMO: linear minimization oracle; MET: memory-efficient Tucker; ST-HOSVD: sequentially truncated higher-order SVD (HOSVD); PRGD: preconditioned Riemannian gradient descent; RGN: Riemannian Gauss--Newton.}
\label{tab:taxonomy_efficiency}
\centering
\scriptsize
\setlength{\tabcolsep}{4pt}
\renewcommand{\arraystretch}{1.05}
\vspace{3pt}
\begin{tabular}{@{}p{0.25\textwidth} p{0.53\textwidth} l@{}}
\toprule
\textbf{Strategy} & \textbf{Representative works} & \textbf{Target} \\
\midrule
\textbf{I. Selective computation} & Krylov/partial SVD \citep{lehoucq1998arpack}; rank-one LMO \citep{guo2017efficient}; truncated SVD \citep{deng2024fast} & FLOPs \\
\textbf{II. Compact representation} & Factor states \citep{bader2007efficient,nimishakavi2018dual}; sparse-plus-low-rank \citep{yao2022low,cai2023generalized} & Memory \\
\textbf{III. Implicit computation} & MET \citep{kolda2008scalable}; ST-HOSVD \citep{vannieuwenhoven2012new}; dimension trees \citep{kaya2016high}; single-mode retractions \citep{zhang2026single} & FLOPs + Memory \\
\textbf{IV. Approximate / randomized} & Randomized Tucker \citep{minster2020randomized}; TensorSketch \citep{malik2018low}; streaming \citep{sun2020lowrank}; D-Tucker \citep{jang2020dtucker}; multilinear Nystr\"om \citep{bucci2024multilinear} & FLOPs + Data \\
\textbf{V. Optimization acceleration} & Preconditioning \citep{kasai2016low}; ScaledGD \citep{tong2022scaling}; PRGD \citep{zhang2025preconditioned}; RGN \citep{luo2023low} & Iterations \\
\bottomrule
\end{tabular}
\end{table}

\paragraph{Efficient exact HOOI.} Several implementations already avoid explicit Kronecker products in HOOI. TensorLy \citep{kossaifi2019tensorly} evaluates each step by a chain of mode products and a truncated SVD. The Tensor Toolbox \texttt{tucker\_als} \citep{bader2023tensortoolbox} uses the same chain and computes the leading eigenvectors of the Gram matrix of the contracted unfolding. HOQRI \citep{sun2022hoqri} replaces the SVD by one QR step per mode and converges to the same fixed points. Therefore, these implementations, and not a literal implementation, are the relevant comparison for our method. Section~\ref{sec:hooi} compares ours with all of them and with the approximate methods above.

\section{Trajectory-Preserving Lightweight Evaluation}
\label{sec:method}
\label{sec:method_redundancy}

Our goal is to remove the redundant work of an iteration without changing its iterates. We first fix the notation and state the condition for this goal (Definition~\ref{def:admissible_class}). Section~\ref{sec:method_components} then gives three components that satisfy the condition, and Section~\ref{sec:method_integration} states the resulting exactness and cost. Appendices~\ref{sec:appendix_pseudocode} and~\ref{sec:appendix_proofs} give the pseudocode and the proofs.

\paragraph{Notation.} Let $\mathcal{X} \in \mathbb{R}^{n_1 \times \dots \times n_N}$ denote an order-$N$ tensor with ambient dimension $\bar{n} := \prod_{k=1}^N n_k$, complementary dimension $n_{-k} := \prod_{j \neq k} n_j$, and mode-$k$ unfolding $X_{(k)} \in \mathbb{R}^{n_k \times n_{-k}}$. A Tucker representation consists of orthonormal factors $U_k \in \mathbb{R}^{n_k \times r_k}$ and a core $\mathcal{G} \in \mathbb{R}^{r_1 \times \dots \times r_N}$; $(r_1, \dots, r_N)$ is the Tucker, or multilinear, rank, and $\bar{r} := \prod_k r_k$. The representation gives $\mathcal{X} = \mathcal{G} \times_1 U_1 \dots \times_N U_N$ and $X_{(k)} = U_k G_{(k)} U_{-k}^\top$, where $U_{-k} := U_N \otimes \dots \otimes U_{k+1} \otimes U_{k-1} \otimes \dots \otimes U_1 \in \mathbb{R}^{n_{-k} \times r_{-k}}$ is the reverse-order Kronecker product without mode $k$, and $r_{-k} := \prod_{j \ne k} r_j$ \citep{kolda2009tensor}. Given measurements $y = \mathcal{A}(\mathcal{X}^\star) + \varepsilon$ of a low Tucker rank tensor $\mathcal{X}^\star$, an estimator learns its Tucker representation by iterating a reference update $\mathcal{X}^{t+1} = \mathcal{T}(\mathcal{X}^t; y)$. For completion, $\Omega$ is the set of $|\Omega|$ observed entries, $P_\Omega$ zeroes all entries outside $\Omega$ ($(P_\Omega \mathcal{X})_i = \mathcal{X}_i$ for $i \in \Omega$), and $P_{\Omega^\perp} := \mathrm{Id} - P_\Omega$.

\paragraph{Condition.} A literal implementation of $\mathcal{T}$ wastes work in three ways. It computes full SVDs but uses only a few singular triplets. It forms Kronecker operators $U_{-k}$ but only applies them. It stores ambient iterates of size $\bar{n}$, although the update depends only on the factors, the core and a sparse residual. We therefore ask when this work can be removed without changing the iterates.

\begin{definition}[Lightweight realization; admissible class]
\label{def:admissible_class}
Let $\mathbb{V} := \mathbb{R}^{n_1 \times \dots \times n_N}$ and let $\mathcal{T}: \mathbb{V} \to \mathbb{V}$ be a reference iteration for fixed data $(\Omega, y)$. A \emph{lightweight realization} of $\mathcal{T}$ is a state set $\mathcal{Z}$ with a decoding map $D: \mathcal{Z} \to \mathbb{V}$ and a transition $\tilde{\mathcal{T}}: \mathcal{Z} \to \mathcal{Z}$ that satisfy the \emph{intertwining relation}
\begin{equation}
D \circ \tilde{\mathcal{T}} = \mathcal{T} \circ D \qquad \text{on } \mathcal{Z}.
\label{eq:intertwining}
\end{equation}
The \emph{admissible class} $\mathfrak{T}$ contains the iterations $\mathcal{T} = \mathcal{T}_U \circ \mathcal{T}_L \circ \mathcal{T}_S$ made of (S) a spectral primitive $\Phi \in \{\mathcal{D}_\lambda, \Pi_r, \mathrm{LMO}_\tau\}$ (\eqref{eq:spectral_identities}) applied to mode-$k$ unfoldings, (L) contractions with factor matrices, and (U) an update assembled from (S) and (L) by multilinear operations.
\end{definition}
In words, one step on the stored state, decoded, equals one reference step on the decoded tensor. Proximal methods for sums of nuclear norms and for the subspace norm, Frank--Wolfe, HOOI and ScaledGD belong to $\mathfrak{T}$, and Appendix~\ref{sec:appendix_alg_derivations} gives their states. Note that the definition does not bound the size of $\mathcal{Z}$. For example, the dual variables of ADMM methods make the state as large as the ambient iterate (Section~\ref{sec:five_limits}).

\subsection{Three exact components}
\label{sec:method_components}

Each component removes one kind of redundant work, and each keeps the output of its stage exact.

\paragraph{Only the required singular triplets.}
\label{sec:method_alg1}
An update applies a spectral primitive to an unfolding, but it uses only some of the singular triplets. The spectral primitives are singular value thresholding (SVT) $\mathcal{D}_\lambda$, rank-$r$ projection $\Pi_r$, and the linear minimization oracle over the nuclear-norm ball, $\mathrm{LMO}_\tau(M) := \arg\min_{\|X\|_* \le \tau} \langle M, X \rangle$. With the singular triplets $(\sigma_i, u_i, v_i)$ of $M$,
\begin{equation}
\mathcal{D}_\lambda(M) = \sum_{i \in I_\lambda(M)} (\sigma_i - \lambda) u_i v_i^\top, \quad \Pi_r(M) = \sum_{i \in I_r(M)} \sigma_i u_i v_i^\top, \quad \mathrm{LMO}_\tau(M) = -\tau u_1 v_1^\top.
\label{eq:spectral_identities}
\end{equation}
Each output therefore uses only the triplets indexed by $I_\lambda(M) := \{i : \sigma_i > \lambda\}$, $I_r(M) := \{1, \dots, r\}$ or $I_{\mathrm{LMO}}(M) := \{1\}$.
\begin{proposition}[Spectral sufficiency]
\label{prop:algorithm1_exactness}
Let $M \in \mathbb{R}^{n_k \times n_{-k}}$ have singular values $\sigma_1 \ge \sigma_2 \ge \dots \ge 0$. (i) $\mathcal{D}_\lambda(M)$ is unique, equals \eqref{eq:spectral_identities}, and depends only on the invariant subspaces $\operatorname{span}\{u_i : \sigma_i > \lambda\}$, $\operatorname{span}\{v_i : \sigma_i > \lambda\}$ and the singular values above $\lambda$ (not on the choice of singular vectors within repeated singular values). (ii) If $\sigma_r > \sigma_{r+1}$, then $\Pi_r(M)$ is unique and equals $P_{\mathcal{U}_r} M$ with $\mathcal{U}_r = \operatorname{span}\{u_1, \dots, u_r\}$; if $\sigma_1 > \sigma_2$, then $\mathrm{LMO}_\tau(M) = -\tau u_1 v_1^\top$ is the unique minimizer. (iii) Consequently, any procedure returning, in exact arithmetic, the singular triplets indexed by $I_\Phi(M)$ reproduces $\Phi(M)$ exactly.
\end{proposition}
\emph{How it is shown.} Part (i) follows from the closed form of soft-thresholding \citep{cai2010singular}. Written as a matrix function of $M^\top M$, this form does not depend on the basis chosen within an eigenspace. Part (ii) follows from the Eckart--Young--Mirsky theorem, because the gap forces every minimizer to have the column space $\mathcal{U}_r$; for the oracle, it follows from the duality of the nuclear and spectral norms. Part (iii) is then immediate. Appendix~\ref{sec:appendix_proofs_lemma1} gives the proof and also bounds the finite-precision error of Krylov extraction by the Davis--Kahan theorem.

Algorithm~\ref{alg:component_extraction} therefore computes only these triplets. It uses Golub--Kahan--Lanczos bidiagonalization with early stopping \citep{golub2013matrix, lehoucq1998arpack, larsen1998lanczos}, and it replaces a full SVD of cost $O(n_k n_{-k} \min(n_k, n_{-k}))$ by
\begin{equation}
\mathrm{FLOP}_{\mathrm{req}} = O\big(q\, c(M) + q^2 (n_k + n_{-k})\big),
\label{eq:component_complexity}
\end{equation}
where $s = |I|$ is the number of required triplets, $q \ge s$ is the Krylov dimension, and $c(M)$ is the cost of one product with $M$ and one with $M^\top$. However, this saves time and memory only if few triplets are needed and a product with $M$ costs much less than forming $M$ (Appendix~\ref{sec:appendix_crossover}). We therefore fix its use by one rule for all estimators: \emph{we use Algorithm~\ref{alg:component_extraction} if and only if $M$ is available only as an implicit operator} (a sparse matrix, or a sum of sparse and factored terms); a small explicit matrix gets a dense SVD. Under this rule, the subspace norm (E0) and HOOI (E3) use dense SVDs of small contracted matrices, HaLRTC (E1) and Frank--Wolfe (E2) use Krylov extraction, and ScaledGD (E4) needs no SVD (Table~\ref{tab:exp1_estimators} in Appendix~\ref{sec:appendix_exp_public}). We fixed the rule before the measurements, and Section~\ref{sec:five_limits} reports the one case in which it picks the slower option.

\paragraph{Contraction without Kronecker products.}
\label{sec:method_alg2}
An update needs only the action $X_{(k)} U_{-k}$ of the Kronecker operator, never $U_{-k}$ itself. Algorithm~\ref{alg:factorwise_contraction} therefore evaluates this action by a chain of mode products, as efficient Tucker codes do \citep{kolda2008scalable, bader2023tensortoolbox}. It uses the standard identities \citep{delathauwer2000multilinear, kolda2009tensor}
\begin{equation}
U_{-k}^\top U_{-k} = \bigotimes_{j \neq k} (U_j^\top U_j) = \mathrm{Id}_{r_{-k}}, \qquad X_{(k)} U_{-k} = \left(\mathcal{X} \times_{j \neq k} U_j^\top\right)_{(k)} = U_k G_{(k)}.
\label{eq:kron_a_b}
\end{equation}
\begin{proposition}[Kronecker-free contraction]
\label{prop:algorithm2_equivalence}
Let $\mathcal{Y} \in \mathbb{R}^{m_1 \times \dots \times m_N}$, $A_j \in \mathbb{R}^{m_j \times m'_j}$ for $j \neq k$, and $A_{-k} := A_N \otimes \dots \otimes A_{k+1} \otimes A_{k-1} \otimes \dots \otimes A_1$. For every contraction order $\pi \in \mathfrak{S}_{N-1}$, Algorithm~\ref{alg:factorwise_contraction} returns $Z_{(k)} = Y_{(k)} A_{-k}$ without forming $A_{-k}$ (whose storage is $m_{-k} m'_{-k}$). If $m'_j \le m_j$ for all $j \ne k$, every intermediate $\mathcal{W}^{(l)}$ has at most $\bar{m} := \prod_j m_j$ entries and the arithmetic cost is at most $2(N-1)\bar{m} \max_{j \ne k} m'_j$; if $\mathcal{Y}$ is sparse with support $\Omega$, the cost is $O(N|\Omega| m'_{-k})$. Moreover, for factors with orthonormal columns, $U_{-k}^\top U_{-k} = \mathrm{Id}_{r_{-k}}$ and $X_{(k)} U_{-k} = U_k G_{(k)} = (\mathcal{X} \times_{j \neq k} U_j^\top)_{(k)}$, i.e.\ \eqref{eq:kron_a_b} holds.
\end{proposition}
\emph{How it is shown.} The proof uses induction over the chain. Products along distinct modes commute, and the unfolding rule $(\mathcal{Y} \times_1 A_1 \cdots \times_N A_N)_{(k)} = A_k Y_{(k)} A_{-k}^\top$ \citep{kolda2009tensor} turns the chain into $Y_{(k)} A_{-k}$. The costs follow from counting the entries of the intermediates (Appendix~\ref{sec:appendix_proofs_lemma2}). Furthermore, when all modes are updated in turn against the same data, as in HOOI, a dimension tree \citep{kaya2016high} reuses partial contractions. It then needs two contractions with the full tensor per sweep instead of $N$. Because it keeps the same Gauss--Seidel order, it computes the same iterates (Appendix~\ref{sec:appendix_alg_hooi}).

\paragraph{A compact sparse-plus-factored state.}
\label{sec:method_alg3}
Finally, an update should not store the ambient iterate. Algorithm~\ref{alg:compact_state_update} therefore keeps the state $z = (\{U_k\}_{k=1}^N, \mathcal{G}, R_\Omega)$ with decoding $D(z) := \mathcal{G} \times_1 U_1 \dots \times_N U_N$ and sparse residual $R_\Omega := y - P_\Omega(D(z))$. With this state, it realizes the \emph{projected imputation iteration}
\begin{equation}
\begin{aligned}
\mathcal{T}_{\mathrm{proj}}:\quad & \mathcal{M}^{t+1} := P_\Omega(y) + P_{\Omega^\perp}(\mathcal{X}^t), \qquad
\mathcal{U}_k^{t+1} := \operatorname{span}\{ u_i(M^{t+1}_{(k)}) : i \in I_\Phi(M^{t+1}_{(k)}) \}, \\
& \mathcal{X}^{t+1} := \mathcal{M}^{t+1} \times_{k=1}^{N} P_{\mathcal{U}_k^{t+1}},
\end{aligned}
\label{eq:proj_iteration}
\end{equation}
where $P_{\mathcal{U}} := UU^\top$. With $\Phi = \Pi_{r_k}$, this iteration is the subspace-projected hard-impute step. Estimators that shrink singular values (SVT proximal steps) or accumulate rank-one atoms (Frank--Wolfe) instead keep the factored output of $\Phi$ as their state (Appendix~\ref{sec:appendix_alg_derivations}).
\begin{proposition}[Compact state intertwining]
\label{prop:algorithm3_commutation}
Let $z = (\{U_k\}, \mathcal{G}, R_\Omega)$ with $R_\Omega = y - P_\Omega(D(z))$ and let $z^+$ be the output of one iteration of Algorithm~\ref{alg:compact_state_update} with exact spectral extraction. Then $D(z^+) = \mathcal{T}_{\mathrm{proj}}(D(z))$ and $R_\Omega^+ = y - P_\Omega(D(z^+))$. Each matrix-vector product with $M^{t+1}_{(k)}$ or its transpose costs $O(|\Omega| + n_k r_k + \bar{r} + N n_{-k} r_{\max})$ and the persistent state occupies
\begin{equation}
\mathrm{MEM}_{\mathrm{cmp}} = \sum_{k=1}^N n_k r_k + \bar{r} + (N+1)|\Omega|, \qquad \zeta := 1 - \frac{\mathrm{MEM}_{\mathrm{cmp}}}{\bar{n}} \to 1 \ \text{ as } \ \frac{\sum_k n_k r_k + \bar{r} + |\Omega|}{\bar{n}} \to 0,
\label{eq:memory_compression}
\end{equation}
counted in words (values plus $N$ coordinates per observed entry).
\end{proposition}
\emph{How it is shown.} The input of the update splits as $\mathcal{M}^{t+1} = R_\Omega^t + D(z^t)$ \citep{mazumder2010spectral}. Therefore, products with its unfoldings need only the sparse residual and the factors, never an $n_k \times n_{-k}$ array. By Proposition~\ref{prop:algorithm1_exactness}, the extracted factors span the selected subspaces. By the mode-product rules, the core update equals $\mathcal{M}^{t+1}$ projected onto these subspaces. Decoding then gives $\mathcal{T}_{\mathrm{proj}}(D(z))$, and we recompute the residual from the new state (Appendix~\ref{sec:appendix_proofs_lemma3}).

\subsection{Exactness and cost}
\label{sec:method_integration}

Composing the three components gives a lightweight iteration $\tilde{\mathcal{T}} := \tilde{\mathcal{T}}_U \circ \tilde{\mathcal{T}}_L \circ \tilde{\mathcal{T}}_S$ on $\mathcal{Z}$. Theorem~\ref{thm:exact_equivalence} shows that this iteration computes the same iterates as the reference.

\begin{theorem}[Exact algebraic trajectory equivalence]
\label{thm:exact_equivalence}
Let $\mathcal{T} \in \mathfrak{T}$ be an iteration whose stage (U) has a realization on a state set $\mathcal{Z}$ that intertwines with it (Algorithm~\ref{alg:compact_state_update} for $\mathcal{T}_{\mathrm{proj}}$; Appendix~\ref{sec:appendix_alg_derivations} for the five estimators), and let $(\mathcal{Z}, D, \tilde{\mathcal{T}})$ be obtained by executing stage (S) with Algorithm~\ref{alg:component_extraction}, stage (L) with Algorithm~\ref{alg:factorwise_contraction}, and stage (U) with this realization. Assume exact arithmetic, that the spectral oracle returns the singular triplets indexed by $I_\Phi$, and, for $\Phi \in \{\Pi_r, \mathrm{LMO}_\tau\}$, the gap condition $\sigma_r > \sigma_{r+1}$ (resp.\ $\sigma_1 > \sigma_2$) for every matrix to which $\Phi$ is applied along the trajectory. If $D(z^0) = \mathcal{X}^0$, then
\begin{equation}
D(z^t) = \mathcal{X}^t \qquad \text{for all } t \ge 0,
\end{equation}
and any stopping rule that is a function of $\{\mathcal{X}^s\}_{s \le t}$ and $(\Omega, y)$ terminates both iterations at the same index $t$.
\end{theorem}
\emph{How it is shown.} Each stage intertwines with its reference counterpart: (S) by Proposition~\ref{prop:algorithm1_exactness}, (L) by Proposition~\ref{prop:algorithm2_equivalence}, and (U) by Proposition~\ref{prop:algorithm3_commutation} or, for the other estimators, by Propositions~\ref{prop:svt_state}--\ref{prop:scaledgd_state} (Appendix~\ref{sec:appendix_alg_derivations}). Composition preserves intertwining, and induction over $t$ turns it into equality of the whole trajectory (Appendix~\ref{sec:appendix_proofs_theorem1}).

\begin{corollary}[Inheritance of guarantees and compositionality]
\label{cor:guarantee_inheritance}
Under the assumptions of Theorem~\ref{thm:exact_equivalence}: (i) every property of the reference sequence $\{\mathcal{X}^t\}_{t \ge 0}$---convergence rates, condition-number independence, sample-complexity and minimax error bounds, and terminal error---holds verbatim for $\{D(z^t)\}_{t \ge 0}$; (ii) if two iterations $\mathcal{T}_1, \mathcal{T}_2$ admit lightweight realizations with a common $(\mathcal{Z}, D)$, then so does $\mathcal{T}_2 \circ \mathcal{T}_1$. In particular, a sketched (Strategy~IV) or preconditioned (Strategy~V) iteration that itself belongs to $\mathfrak{T}$ is evaluated exactly by $\tilde{\mathcal{T}}$. This concerns exactness only; whether it also saves time or memory depends on the implementation it is compared with.
\end{corollary}
Part (i) holds because the two sequences are the same sequence of tensors. Guarantees on running time, however, are not inherited; Theorem~\ref{thm:complexity} replaces them. Part (ii) is the composition rule used above (Appendix~\ref{sec:appendix_proofs_corollary1}).

\begin{theorem}[Per-iteration complexity bounds]
\label{thm:complexity}
Let $r_k$ bound the mode-$k$ rank of the compact state before and after the iteration, $r_{\max} := \max_k r_k$, $s := \max_k |I_k| \le r_{\max}$, and let $q$ with $s \le q \le \min_k \min(n_k, n_{-k})$ be the Krylov dimension of Algorithm~\ref{alg:component_extraction}. One iteration of Algorithm~\ref{alg:compact_state_update} costs
\begin{align}
\mathrm{FLOP}(\tilde{\mathcal{T}}) & = O\Big( q N |\Omega| + q\,(N r_{\max} + q) \sum_{k=1}^N (n_k + n_{-k}) + N |\Omega| \bar{r} + \sum_{k=1}^N n_k r_k^2 + \bar{r} \sum_{k=1}^N r_k \Big), \label{eq:thm_flop} \\
\mathrm{MEM}_{\mathrm{state}}(\tilde{\mathcal{T}}) & = \sum_{k=1}^N n_k r_k + \bar{r} + (N+1)|\Omega|, \label{eq:thm_mem} \\
\mathrm{MEM}_{\max}(\tilde{\mathcal{T}}) & = \mathrm{MEM}_{\mathrm{state}}(\tilde{\mathcal{T}}) + O\big( N|\Omega| + q \max_k (n_k + n_{-k}) \big), \notag
\end{align}
where $\mathrm{MEM}_{\mathrm{state}}$ counts the words of the persistent state carried between iterations and $\mathrm{MEM}_{\max}$ the peak including Krylov scratch space and the index structures of the sparse products ($O(N|\Omega|)$). The literal implementation (full SVD of each unfolding and explicit Kronecker operators) requires $\mathrm{FLOP} = \Omega\big(\bar{n} \sum_k \min(n_k, n_{-k})\big)$ and $\mathrm{MEM} = \Omega(\bar{n})$. For $n_k = n$, $r_k = r$, and fixed $N$, \eqref{eq:thm_flop} is $O\big(q(q + r) n^{N-1} + (q + r^N)|\Omega|\big)$ against $\Omega(n^{N+1})$ for the literal implementation. If $q$ does not grow with $n$, this is one polynomial order lower in $n$; if $q$ grows proportionally to $n$, the two are of the same order. The persistent state is $O(Nnr + r^N + |\Omega|)$ against $\Omega(n^N)$; at a fixed sampling rate $p$, $|\Omega| = p\,n^N$, so both are of order $n^N$, and the state is smaller than one ambient array only if $(N+1)p < 1$.
\end{theorem}
\emph{How it is shown.} We add, for every mode, the costs of Propositions~\ref{prop:algorithm1_exactness}--\ref{prop:algorithm3_commutation}: $q$ products with the sparse-plus-factored unfolding, the Krylov work, the core update and the residual (term by term in Appendix~\ref{sec:appendix_complexity}). With a fixed sampling rate, the term $(q+r^N)|\Omega|$ dominates. Therefore, the order in $n$ drops only where the reference computes full SVDs of unfoldings (HaLRTC, Frank--Wolfe) and $q$ stays bounded. This condition fails for HaLRTC at $p=0.05$, where $q\approx n$. For the other estimators, every implementation scales as $n^N$, and the gain is a constant factor. Note that these are operation counts, and measured times approach them only slowly: between $n=500$ and $700$, the slope of Raw HaLRTC and Frank--Wolfe is still $\slRawHiLo$--$\slRawHiHi$ instead of $4$ (Appendix~\ref{sec:appendix_complexity_dominant}).

\section{Experiments}
\label{sec:experiments}

We first check that all implementations compute the same iterates. We then ask three questions. First, is lightweight HOOI lighter than the other 14 HOOI-type methods (Section~\ref{sec:hooi}, Figure~\ref{fig:hooi_main})? Second, is the lightweight evaluation cheaper than efficient public code for five estimators (Section~\ref{sec:five}, Table~\ref{tab:five})? Third, does it fit an edge-sized device on real data (Section~\ref{sec:real}, Figure~\ref{fig:edge})?

\paragraph{Protocol.} All measurements follow one protocol (Appendix~\ref{sec:appendix_experimental_details}). We use one NVIDIA A100 GPU (40\,GiB) in float64 with TF32 disabled, and we start a fresh process for each run. We measure wall-clock time with the device synchronized before every reading. We measure memory as the peak GPU memory of the run, which includes the input data, all workspaces and all temporaries. We compare three implementations of each estimator. \emph{Raw} is the literal implementation (explicit Kronecker products, full SVDs, ambient iterates). \emph{Public} is a faithful port of the authors' code or of a standard library (Table~\ref{tab:exp1_estimators}). Ours uses one fixed configuration throughout.

\paragraph{Same iterates.}
\label{sec:exact}
\label{sec:five_exact}
We run Raw, Public and ours in lockstep for $T=20$ iterations ($n=30$, order 3, and $n=12$, order 4), and we record $\max_t\|D(z^t)-\mathcal{X}^t\|_F/\|\mathcal{X}^t\|_F$ against Raw (Table~\ref{tab:exactness}). These runs use our NumPy reference implementations in float64, which the GPU implementations port (Appendix~\ref{sec:appendix_experimental_details}). Over all five estimators, the largest discrepancy is $\exactTight$ with a tight Krylov tolerance and $\exactDefault$ with the default one. Public Frank--Wolfe follows a different trajectory, because its oracle is an inexact power method; the other public implementations reproduce Raw. On the GPU, the final error of ours agrees with that of Raw to a relative $\gpuAgreeBench$ in all $\gpuAgreeBenchRuns$ runs with E0, E1, E3 or E4.

\begin{table}[!ht]
\caption{Same iterates: largest relative discrepancy $\max_t\|D(z^t)-\mathcal{X}^t\|_F/\|\mathcal{X}^t\|_F$ from Raw over $T=20$ lockstep iterations (NumPy reference implementations, float64). Krylov tolerance tight ($10^{-14}$) or default ($10^{-10}$). $^\ddagger$Public Frank--Wolfe uses an inexact power-method oracle and therefore a different trajectory.}
\label{tab:exactness}
\centering
\footnotesize
\setlength{\tabcolsep}{3pt}
\begin{tabular}{lcccccc}
\toprule
 & \multicolumn{4}{c}{Ours vs.\ Raw} & \multicolumn{2}{c}{Public vs.\ Raw} \\
\cmidrule(lr){2-5}\cmidrule(lr){6-7}
 & \multicolumn{2}{c}{tight ($10^{-14}$)} & \multicolumn{2}{c}{default ($10^{-10}$)} & \multicolumn{2}{c}{} \\
Estimator & $\kappa=1$ & $\kappa=100$ & $\kappa=1$ & $\kappa=100$ & $\kappa=1$ & $\kappa=100$ \\
\midrule
E0 Subspace norm & $1.2\times10^{-14}$ & $5.2\times10^{-15}$ & $1.2\times10^{-14}$ & $5.2\times10^{-15}$ & 0 & 0 \\
E1 HaLRTC & $4.4\times10^{-15}$ & $6.7\times10^{-15}$ & $4.4\times10^{-15}$ & $2.5\times10^{-13}$ & $2.8\times10^{-15}$ & $1.1\times10^{-15}$ \\
E2 Frank--Wolfe & $5.2\times10^{-14}$ & $4.5\times10^{-14}$ & $9.5\times10^{-11}$ & $1.3\times10^{-11}$ & $1.4\times10^{0}$$^\ddagger$ & $4.8\times10^{-1}$$^\ddagger$ \\
E3 HOOI & $3.7\times10^{-15}$ & $1.9\times10^{-15}$ & $3.7\times10^{-15}$ & $1.9\times10^{-15}$ & $3.4\times10^{-15}$ & $3.8\times10^{-15}$ \\
E4 ScaledGD & $2.7\times10^{-16}$ & $3.7\times10^{-16}$ & $2.7\times10^{-16}$ & $3.7\times10^{-16}$ & $2.6\times10^{-16}$ & $3.0\times10^{-16}$ \\
\bottomrule
\end{tabular}

\end{table}

\subsection{HOOI among 15 HOOI-type methods}
\label{sec:hooi}
HOOI is the strictest test, because it is the most basic iterative Tucker estimator and has the most published accelerations. \emph{Setup.} We denoise $\mathcal{Y}=\mathcal{X}^\star+\mathcal{E}$ at an SNR of 20\,dB. The true tensor $\mathcal{X}^\star$ has Haar-random factors and a superdiagonal core whose mode spectra decay geometrically from 1 to $1/\kappa$ ($\kappa=100$). We use order 3 with $n\in\{\famSizesThree\}$ and $r=5$ (inputs of \famSmallDataThree\,MiB to \famDataMaxGiB\,GiB), and order 4 with $n\in\{\famSizesFour\}$ and $r=3$. We run \famSeeds{} seeds and 10 sweeps, and we time each run end to end. The 15 methods are five exact implementations (naive HOOI, TensorLy, the Tensor Toolbox with a dense or an iterative eigensolver, and ours), HOQRI, which reaches the same fixed points, and nine randomized, sketched or slice-compressed methods with their authors' defaults (Appendix~\ref{sec:appendix_hooi_family}).

\begin{figure}[t]
\centering
\includegraphics[width=\linewidth]{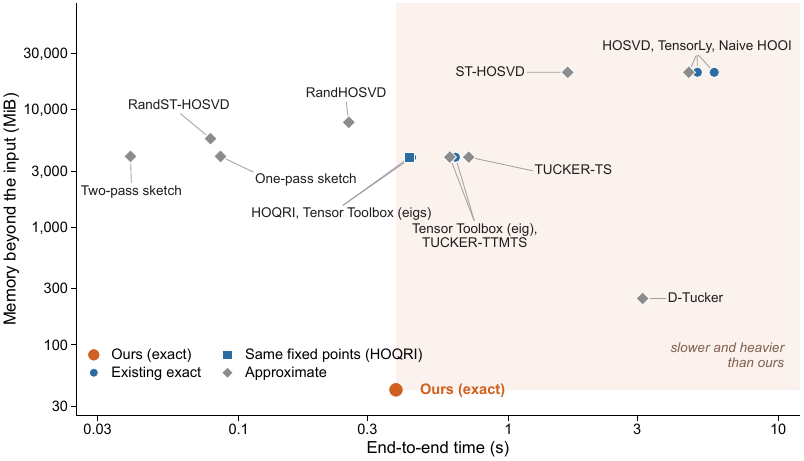}
\caption{Lightweight HOOI among all 15 HOOI-type methods on $\famNThree^3$ tensors ($r=5$, denoising, A100, mean of \famSeeds{} seeds), the largest size at which every method runs: end-to-end time against the GPU memory a method needs beyond the \famData\,MiB input (data-inclusive peak minus the input). Methods that nearly coincide share a label. The shaded region is slower and heavier than ours; it contains every exact method and HOQRI. Every faster method is approximate (errors: Figure~\ref{fig:hooi_family} in Appendix~\ref{sec:appendix_hooi_family}). All sizes: Table~\ref{tab:hooi_ratio} in Appendix~\ref{sec:appendix_hooi_family}.}
\label{fig:hooi_main}
\end{figure}

\emph{Results.} Ours is the lightest method at every size, and it runs about as fast as the fastest exact implementations (Figure~\ref{fig:hooi_main}).
\begin{itemize}[leftmargin=*,itemsep=1pt,topsep=2pt]
\item \emph{Memory.} On $\famNThree^3$ tensors, ours needs \famWorkOurs\,MiB beyond the \famData\,MiB input. In contrast, the lightest existing exact implementation, which holds a copy of the input, needs \famWorkBestMB\,MiB ($\famWLargest\times$ more). Against the lightest existing implementation at each size, the saving grows from $\famWFirstThree\times$ at $n=\famNMinThree$ to $\famWMax\times$ at $n=\famNMax$, and from $\famWFirstFour\times$ to $\famWLargestFour\times$ at order 4 (Table~\ref{tab:hooi_ratio} in Appendix~\ref{sec:appendix_hooi_family}). On $\famNMax^3$ tensors (\famDataMaxGiB\,GiB), ours needs \famWorkOursMax\,MiB against \famWorkBestMaxGiB\,GiB, and \famFailNames{} do not run (a cuSOLVER error in the SVD of the $1000\times10^6$ unfolding).
\item \emph{Time.} On $\famNThree^3$ tensors, every existing exact implementation is both slower and heavier than ours (Figure~\ref{fig:hooi_main}). The fastest of them, \famBestExistingName, takes $\famSpeedBest\times$ as long as ours; TensorLy takes $\famSpeedTensorly\times$, and naive HOOI $\famSpeedNaive\times$. On smaller tensors, however, \famSlowThreeNames{} is faster. It uses one QR step per mode and reaches the same fixed points along different iterates. There, the best existing implementation takes \famTSlowRangeThree$\times$ our time for $n\le\famTSlowMaxNThree$, and \famTSlowRangeFour$\times$ at order 4. We attribute this to fixed per-kernel costs, which dominate on small tensors.
\item \emph{Same estimate.} All exact methods and HOQRI reach the same relative error to $\mathcal{X}^\star$. On the GPU, their final errors agree with ours to a relative $\gpuAgreeFam$, and in lockstep, our iterates agree with naive HOOI to $\exactHOOI$ (Section~\ref{sec:five_exact}).
\item \emph{Approximate methods buy speed with accuracy, not memory.} Every method that is faster than ours on $\famNThree^3$ tensors is approximate, with an error at least $\famMinErrRatioFaster\times$ that of HOOI. Moreover, none of the nine approximate methods needs less memory at any size.
\end{itemize}

\begin{table}[t]
\caption{Ours at $n=\benchNmax$ (A100, order 3, $r=5$, mean of \benchSeedsMax{} seeds): time per iteration and data-inclusive peak GPU memory of Public and of Raw divided by ours; above $1\times$ favors ours (bold: at least $1.2\times$). Absolute values: Table~\ref{tab:five_full}; all sizes: Appendix~\ref{sec:appendix_tab_bench}.}
\label{tab:five}
\centering
\footnotesize
\setlength{\tabcolsep}{4pt}
\begin{tabular}{llrrrl}
\toprule
 & & \multicolumn{2}{c}{Public / Ours} & Raw / Ours & \\
\cmidrule(lr){3-4}\cmidrule(lr){5-5}
Estimator & $p$ & time & peak memory & peak memory & Why slower \\
\midrule
E0 Subspace norm & -- & \textbf{3.45$\times$} & \textbf{4.03$\times$} & \textbf{4.03$\times$} &  \\
E3 HOOI & -- & \textbf{2.41$\times$} & \textbf{1.98$\times$} & \textbf{3.98$\times$} &  \\
E4 ScaledGD & 0.05 & \textbf{3.69$\times$} & \textbf{8.37$\times$} & \textbf{7.19$\times$} &  \\
E4 ScaledGD & 0.2 & \textbf{1.30$\times$} & \textbf{2.31$\times$} & \textbf{2.01$\times$} &  \\
E1 HaLRTC & 0.2 & \textbf{1.21$\times$} & \textbf{2.27$\times$} & \textbf{1.96$\times$} &  \\
\midrule
E1 HaLRTC & 0.05 & 0.02$\times$ & \textbf{1.23$\times$} & 1.03$\times$ & nearly full rank ($q\approx n$) \\
E2 Frank--Wolfe & 0.05 & 0.07$\times$ & \textbf{2.07$\times$} & \textbf{6.54$\times$} & public code: inexact oracle \\
E2 Frank--Wolfe & 0.2 & 0.13$\times$ & \textbf{1.72$\times$} & \textbf{3.35$\times$} & public code: inexact oracle \\
\bottomrule
\end{tabular}

\end{table}

\subsection{Five estimators against efficient public code}
\label{sec:five}
\label{sec:five_cost}
\emph{Setup.} We run Raw, Public and ours with the same iteration, the same initialization and the same hyperparameters for E0 subspace norm \citep{zheng2015interpolating}, E1 HaLRTC \citep{liu2012tensor}, E2 Frank--Wolfe \citep{guo2017efficient}, E3 HOOI and E4 ScaledGD \citep{tong2022scaling}. We use order-3 tensors with $n\in\{\benchSizes\}$, $r=5$ and $\kappa=100$. Completion observes a uniformly random fraction $p\in\{0.05,0.2\}$ of the entries, and denoising adds noise at an SNR of 20\,dB. We run $T=10$ iterations and report the mean over \benchSeedsMax{} seeds of the median time per iteration.

\emph{Results.} At the largest size, ours is both faster and lighter than the public implementation for three estimators (Table~\ref{tab:five}; all sizes in Appendix~\ref{sec:appendix_tab_bench}): the subspace norm ($\rZeroT\times$ in time, $\rZeroM\times$ in peak memory), HOOI ($\rThreeT\times$, $\rThreeM\times$) and ScaledGD ($\rFourLowT\times$ and $\rFourLowM\times$ at $p=0.05$, $\rFourHighT\times$ and $\rFourHighM\times$ at $p=0.2$). On smaller problems, however, ours is not always faster. It is faster than the public code at every size for HOOI, from $n=\rZeroFasterFrom$ for the subspace norm, and from $n=\rFourLowFasterFrom$ ($p=0.05$) and $n=\rFourHighFasterFrom$ ($p=0.2$) for ScaledGD; at $n=\benchNmin$, it takes up to $\rSlowSmallMax\times$ as long (Appendix~\ref{sec:appendix_tab_bench}). We attribute this to the fixed cost of launching many small GPU kernels, which dominates an iteration on small problems. Against Raw, completion with $p=0.05$ needs $\rRawTwoM\times$ (Frank--Wolfe) and $\rRawFourM\times$ (ScaledGD) less peak memory with the data included. In contrast, denoising stays at about $\rRawZeroM\times$, because the dense input itself is part of the peak.

\emph{Why.}
\label{sec:five_limits}
Table~\ref{tab:map} (Appendix~\ref{sec:appendix_tab_bench}) lists every comparison in which ours is faster or slower, with its cause, and Figure~\ref{fig:map} gives the conditions behind them. Ours needs less memory in every case, though only marginally for HaLRTC, whose dual variables stay ambient (Figure~\ref{fig:map}b). Ours is faster when the comparator forms objects that ours never forms. In contrast, ours is slower for one of two reasons: the comparator computes something else, so that the two do not produce the same iterates, or a cause that can be predicted in advance. For example, for HaLRTC at $p=0.05$, the thresholded unfoldings stay nearly full rank (about 280 of 300 singular values per mode at $n=300$), and the dual variables are dense (Appendix~\ref{sec:appendix_alg_derivations}). We attribute the slowdown to this: Algorithm~\ref{alg:component_extraction} then extracts nearly all triplets, one small kernel at a time. At $p=0.2$, in contrast, ours is $\rOneHighT\times$ faster with $\rOneHighM\times$ less memory. A microbenchmark locates the crossover (Appendix~\ref{sec:appendix_crossover}). Algorithm~\ref{alg:component_extraction} beats a dense SVD only for $n\ge\xoNstar$ and $s\le\xoSnMax\,n$ ($\xoGainLo$--$\xoGainHi\times$ at $s=1$, $n=\xoNmax$, with $\xoMemFewLo$--$\xoMemFewHi\times$ less GPU memory for $s\le5$). For $s\ge n/2$, it is $\xoSlowLo$--$\xoSlowHi\times$ slower and needs about as much memory.

\begin{figure}[t]
\centering
\includegraphics[width=\linewidth]{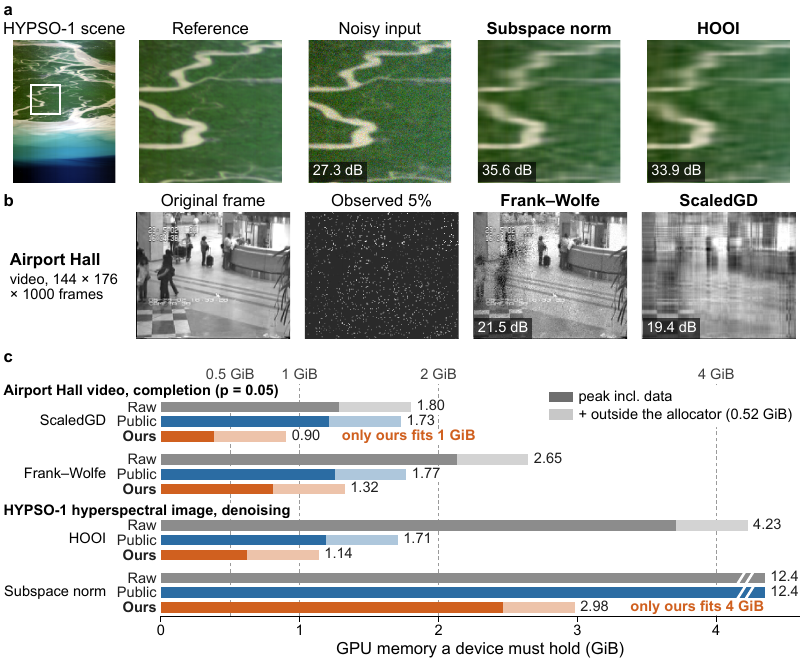}
\caption{Real data on edge-sized devices (A100, float64, $T=20$). \textbf{a},~HYPSO-1 hyperspectral cube ($956\times684\times120$): whole scene, and the marked region of the reference, the noisy input (SNR 20\,dB) and the cube denoised with $r=20$ (RGB composite); PSNR over the whole cube. \textbf{b},~Airport Hall video ($144\times176\times1000$): frame 100, its observed entries ($p=0.05$) and the completed frame ($r=10$); PSNR on all unobserved entries. Images: ours, seed 0; Raw gives the same. \textbf{c},~GPU memory a device must hold: data-inclusive peak (dark; median of 3 seeds) plus the \edgeCtx\,GiB held outside the PyTorch allocator (light), against device sizes (dashed).}
\label{fig:edge}
\end{figure}

\subsection{Real data on edge-sized devices}
\label{sec:real}
Recording devices often lack the storage and computation for low-rank recovery \citep{xiang2020edge}. We therefore repeat the comparison on two real data sets (Figure~\ref{fig:edge}): completion of the Airport Hall surveillance video \citep{li2004statistical} with ScaledGD and Frank--Wolfe, and denoising of a HYPSO-1 satellite hyperspectral cube \citep{justo2023open} with the subspace norm and HOOI. Ours agrees with Raw to a relative $\edgeSameEstOther$ at 512 fixed positions ($\edgeSameEstFW$ for Frank--Wolfe; same PSNR). It needs $\edgeRedPubMin$--$\edgeRedPubMax\times$ less peak memory than the public implementations and $\edgeRedRawMin$--$\edgeRedRawMax\times$ less than Raw (Table~\ref{tab:edge}). Counting the \edgeCtx\,GiB held outside the PyTorch allocator, only ours fits a \edgeSepHall\,GiB device for the video and a \edgeSepHypso\,GiB device for the cube. At allocator caps of these sizes, the other implementations run out of memory (Table~\ref{tab:edge_budget}). However, ours is not faster everywhere. Frank--Wolfe is $\edgeFWSlow\times$ slower than the public code, which uses an inexact oracle; HaLRTC is not lighter; and among HOOI-type methods on the cube, ours is $\edgeFamTTBSlow\times$ slower than the Tensor Toolbox (eig) (Table~\ref{tab:map}).

\section{Discussion and Conclusion}
\label{sec:conclusion}
\label{sec:discussion}
\label{sec:conclusion_main}

\paragraph{When to use it.} Same iterates, less memory and less time are separate properties (Figure~\ref{fig:map}). Same iterates hold for an iteration in $\mathfrak{T}$ whose update has an intertwining state (Theorem~\ref{thm:exact_equivalence}). Where only the required singular triplets are computed, a spectral gap is also needed (Proposition~\ref{prop:algorithm1_exactness}). Every guarantee of the reference estimator then applies (Corollary~\ref{cor:guarantee_inheritance}). In contrast, the other two properties have necessary conditions that are not sufficient (Appendix~\ref{sec:appendix_mapext}). \emph{Less memory} requires three things: the state does not stay ambient, as the dual variables of ADMM methods do; an update needs few singular triplets; and, for completion, the sampling rate is low enough. Because ours holds about \mxWordsObs{} words per observed entry, this threshold depends on how many dense arrays the comparator holds. \emph{Less time} than the fastest implementation of the same iterates requires two things: few singular triplets on a large problem, and a comparator that forms what ours avoids, such as large SVDs, unfoldings, Kronecker operators or dense imputed tensors. Where the first condition fails, ours was never faster (\mxNoGainFast{} of \mxNoGainCells{} cases). For the five estimators, ours saved time for three of them on large problems. It also saved memory in every synthetic comparison, though only marginally for HaLRTC, whose dual variables stay ambient; on the video, it saved none for HaLRTC (Section~\ref{sec:real}).

\paragraph{Limitations.} The framework changes how an iteration is computed, not what it computes. It therefore provides no new statistical or convergence result. Entrywise nonlinearities and non-spectral proximal maps lie outside $\mathfrak T$. Even within $\mathfrak T$, intertwining does not bound the state. For example, the Frank--Wolfe atom pool grows with $t$, as do HaLRTC's dual variables unless we keep them in ambient form (Appendix~\ref{sec:appendix_alg_derivations}), and periodic recompression would change the reference iteration. Our runs of these estimators stop after at most 20 iterations, so we do not measure this growth up to convergence. All measurements use one A100 in float64, and memory caps emulate edge-sized devices. Our GPU code ports our NumPy code and is not tuned for the GPU, so crossover points may move with other hardware, precision and kernels. Finally, our synthetic spectra decay geometrically with $\kappa=100$, except for a power law in Appendix~\ref{sec:appendix_crossover}; we leave other spectra to future work.

\paragraph{Conclusion.} Computing the same iterates without forming the large intermediates can save memory and time together. For example, lightweight HOOI is the lightest of 15 HOOI-type methods and, on large problems, as fast as the best exact implementations; three of five estimators are both faster and lighter than their public code. However, neither gain is guaranteed. Each gain needs the conditions of Figure~\ref{fig:map}, and these conditions fail on small problems, with dense variables or full-rank iterates, and against comparators that already avoid the large objects.

\section*{Reproducibility Statement}
We will release all code (estimators, public-code ports, measurement and table generation) upon acceptance. Scripts generate every number in Section~\ref{sec:experiments} from raw JSON logs, and Appendix~\ref{sec:appendix_experimental_details} gives the measurement protocol.

\section*{AI Use Statement}
In this work, we used generative AI tools (Claude) to assist with synthetic data generation, theoretical and conceptual model development, mathematical proof writing, research method implementation, translation, and interpretation of results. Generative AI tools were not used for other tasks requiring disclosure, where applicable.

Additionally, generative AI tools were used for language editing, proofreading, and code creation and editing.

All AI-assisted outputs were reviewed by the authors. AI-assisted code was tested for correctness where applicable, and experimental results and their interpretations were verified against the underlying experiments.

The authors take full responsibility for the final content of this work, including all text, claims, and artifacts produced with the assistance of generative AI.

\section*{Funding}
This work was supported by Japan Science and Technology Agency,
Support for Pioneering Research Initiated by the Next Generation
(no. JPMJSP2124).

\bibliography{iclr2027_conference}
\bibliographystyle{iclr2027_conference}

\clearpage
\appendix
\section*{Appendix Overview}
\begin{figure}[H]
\centering
\includegraphics[width=\linewidth]{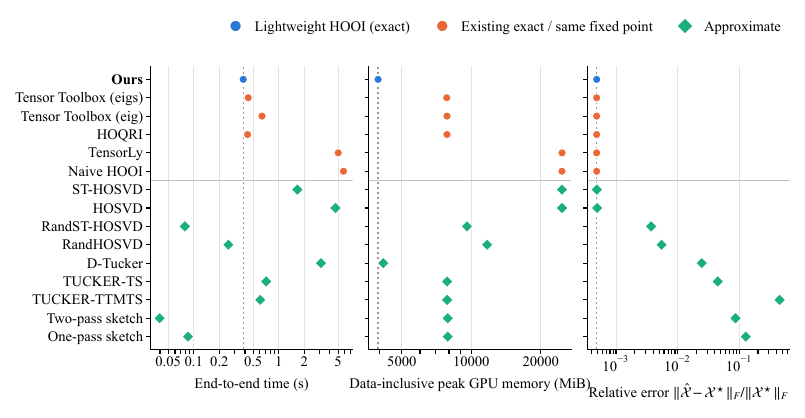}
\caption{All 15 HOOI-type methods on $\famNThree\times\famNThree\times\famNThree$ denoising ($r=5$, \famSeeds{} seeds): end-to-end time, data-inclusive peak GPU memory (A100) and relative error to the true tensor. Error bars show $\pm$ s.e.\ and are mostly smaller than the markers. Dashed lines mark our method. The input tensor alone occupies \famData\,MiB.}
\label{fig:hooi_family}
\end{figure}

\addcontentsline{toc}{section}{Appendix Overview}
Each item names an appendix, summarizes its content, and states which claim of the main text it supports.
\begin{itemize}[leftmargin=*,itemsep=3pt,topsep=3pt]
\item \textbf{Appendix~\ref{sec:appendix_pseudocode}: Pseudocode of the Three Components.} It gives the pseudocode of the three components (Algorithms~\ref{alg:component_extraction}--\ref{alg:compact_state_update}), which implement the method of Section~\ref{sec:method_components}.
\item \textbf{Appendix~\ref{sec:appendix_proofs}: Complete Proofs of Theorems and Propositions.} It proves Propositions~\ref{prop:algorithm1_exactness}--\ref{prop:algorithm3_commutation}, Theorem~\ref{thm:exact_equivalence} and Corollary~\ref{cor:guarantee_inheritance}, and it bounds the finite-precision error of Krylov extraction (Appendix~\ref{sec:appendix_proofs_lemma1}). These proofs support the exactness claims of Sections~\ref{sec:method_components} and~\ref{sec:method_integration}.
\item \textbf{Appendix~\ref{sec:appendix_complexity}: Detailed Complexity Derivations and Arithmetic Counts.} It derives the operation count and the peak memory of Theorem~\ref{thm:complexity} term by term, including the words per observed entry at the peak (Appendix~\ref{sec:appendix_complexity_mem}), and it compares computational paradigms (Table~\ref{tab:paradigm_comparison_appendix}). Its measured exponents (Table~\ref{tab:slopes}) support the remark in Section~\ref{sec:method_integration} that measured times approach the operation counts only slowly.
\item \textbf{Appendix~\ref{sec:appendix_alg_derivations}: Concrete Algorithmic Instantiations.} For the five estimators, it gives the state, the decoding and the intertwining relation (Propositions~\ref{prop:svt_state}--\ref{prop:scaledgd_state}), and the dimension tree of HOOI (Appendix~\ref{sec:appendix_alg_hooi}). It supports Theorem~\ref{thm:exact_equivalence} for these estimators and the remarks on the size of the state after Definition~\ref{def:admissible_class} and in Section~\ref{sec:conclusion}.
\item \textbf{Appendix~\ref{sec:appendix_implementation}: Implementation Details and Numerical Safeguards.} It describes how we implement Krylov extraction, copy-free contractions and sparse operations, and how we count the memory of the state. It supports the single configuration used in Section~\ref{sec:experiments}.
\item \textbf{Appendix~\ref{sec:appendix_experimental_details}: Experimental Protocol.} It specifies the hardware, the measurement of memory and time, the data, the memory caps and the hyperparameters, and Appendix~\ref{sec:appendix_exp_public} (Table~\ref{tab:exp1_estimators}) lists the public implementations. It supports the protocol of Section~\ref{sec:experiments}.
\item \textbf{Appendix~\ref{sec:appendix_tab_bench}: Complete Results of the Five-Estimator Comparison.} It gives the absolute values behind Table~\ref{tab:five} (Table~\ref{tab:five_full}), the cause of every gain and loss (Table~\ref{tab:map}), all sizes for each estimator (Tables~\ref{tab:bench_E0}--\ref{tab:bench_E4}) and the real-data numbers (Appendix~\ref{sec:appendix_real}, Tables~\ref{tab:edge}--\ref{tab:edge_budget}). It supports Sections~\ref{sec:five} and~\ref{sec:real} and Figure~\ref{fig:map}b--c.
\item \textbf{Appendix~\ref{sec:appendix_hooi_family}: HOOI-Type Methods: Parameters and Complete Results.} It describes the 15 HOOI-type methods and reports their errors (Figure~\ref{fig:hooi_family}), the memory ratios at all sizes (Table~\ref{tab:hooi_ratio}), an ablation of our components (Table~\ref{tab:hooi_ablation}), the absolute time and memory of the exact implementations (Table~\ref{tab:hooi_exact}) and all measurements (Tables~\ref{tab:hooi_all_3_100}--\ref{tab:hooi_all_4_80}). It supports Section~\ref{sec:hooi}.
\item \textbf{Appendix~\ref{sec:appendix_crossover}: When Algorithm~\ref{alg:component_extraction} Pays Off.} A microbenchmark against a dense SVD locates where Algorithm~\ref{alg:component_extraction} is faster and lighter (Table~\ref{tab:crossover}). It supports the fixed rule of Section~\ref{sec:method_components}, the explanation in Section~\ref{sec:five} (paragraph ``Why'') and the memory condition on singular triplets in Figure~\ref{fig:map}b.
\item \textbf{Appendix~\ref{sec:appendix_mapext}: Testing the Map on \mxNestCap{} Further Estimators.} A test with predictions fixed in advance rejects the map drawn from the five estimators as a predictor of gains (Table~\ref{tab:mapext}), whereas a count made after the test supports the conditions that rule out a gain, and all \mxNest{} estimators reproduce the reference iterates. It supports Figure~\ref{fig:map}b--c, the conditions in Section~\ref{sec:conclusion} and the first contribution (Section~\ref{sec:intro}).
\end{itemize}

\section{Pseudocode of the Three Components}
\label{sec:appendix_pseudocode}

\begin{algorithm}[H]
\caption{Extraction of the Required Singular Triplets}
\label{alg:component_extraction}
\begin{algorithmic}[1]
\REQUIRE $M: w \mapsto M w, v \mapsto M^\top v$; $\Phi \in \{\mathcal{D}_\lambda, \Pi_r, \mathrm{LMO}_\tau\}$; $\epsilon_{\mathrm{krylov}} > 0$.
\ENSURE $\{(\tilde{\sigma}_i, \tilde{u}_i, \tilde{v}_i)\}_{i \in I}$, $\Phi(M)$.
\IF{$\Phi = \mathcal{D}_\lambda$}
    \STATE $(\tilde{\sigma}_i, \tilde{u}_i, \tilde{v}_i)_{i=1}^s \leftarrow \operatorname{KrylovBidiag}(M, \epsilon_{\mathrm{krylov}}) \quad \text{s.t.} \quad \tilde{\sigma}_s > \lambda \ge \tilde{\sigma}_{s+1}$
    \STATE $I \leftarrow \{1, \dots, s\}$
    \STATE $\Phi(M) \leftarrow \sum_{i=1}^s (\tilde{\sigma}_i - \lambda) \tilde{u}_i \tilde{v}_i^\top$
\ELSIF{$\Phi = \Pi_r$}
    \STATE $(\tilde{\sigma}_i, \tilde{u}_i, \tilde{v}_i)_{i=1}^r \leftarrow \operatorname{KrylovBidiag}(M, r, \epsilon_{\mathrm{krylov}})$
    \STATE $I \leftarrow \{1, \dots, r\}$
    \STATE $\Phi(M) \leftarrow \sum_{i=1}^r \tilde{\sigma}_i \tilde{u}_i \tilde{v}_i^\top$
\ELSIF{$\Phi = \mathrm{LMO}_\tau$}
    \STATE $(\tilde{\sigma}_1, \tilde{u}_1, \tilde{v}_1) \leftarrow \operatorname{KrylovBidiag}(M, 1, \epsilon_{\mathrm{krylov}})$
    \STATE $I \leftarrow \{1\}$
    \STATE $\Phi(M) \leftarrow -\tau \tilde{u}_1 \tilde{v}_1^\top$
\ENDIF
\STATE \textbf{return} $\{(\tilde{\sigma}_i, \tilde{u}_i, \tilde{v}_i)\}_{i \in I}, \Phi(M)$
\end{algorithmic}
\end{algorithm}

\begin{algorithm}[H]
\caption{Kronecker-Free Contraction}
\label{alg:factorwise_contraction}
\begin{algorithmic}[1]
\REQUIRE $\mathcal{Y} \in \mathbb{R}^{m_1 \times \dots \times m_N}$, $\{A_j \in \mathbb{R}^{m_j \times m'_j}\}_{j \neq k}$, $k \in \{1, \dots, N\}$.
\ENSURE $Z_{(k)} = \left(\mathcal{Y} \times_{j \neq k} A_j^\top\right)_{(k)} \in \mathbb{R}^{m_k \times m'_{-k}}$.
\STATE $\pi \leftarrow \operatorname{argsort}_{j \in \{1, \dots, N\} \setminus \{k\}} (m_j / m'_j, \text{descending})$
\STATE $\mathcal{W}^{(0)} \leftarrow \mathcal{Y}$
\FOR{$l = 1, \dots, N-1$}
    \STATE $\mathcal{W}^{(l)} \leftarrow \mathcal{W}^{(l-1)} \times_{\pi_l} A_{\pi_l}^\top$
\ENDFOR
\STATE $Z_{(k)} \leftarrow W_{(k)}^{(N-1)}$
\STATE \textbf{return} $Z_{(k)}$
\end{algorithmic}
\end{algorithm}

\begin{algorithm}[H]
\caption{Sparse-plus-Factored State Update}
\label{alg:compact_state_update}
\begin{algorithmic}[1]
\REQUIRE $z^0 = (\{U_k^0\}_{k=1}^N, \mathcal{G}^0, R_\Omega^0)$, $(\Omega, y)$, $\epsilon_{\mathrm{tol}} > 0$.
\ENSURE $z^T = (\{U_k^T\}_{k=1}^N, \mathcal{G}^T, R_\Omega^T)$.
\FOR{$t = 0, 1, \dots, T-1$}
    \FOR{$k = 1, \dots, N$}
        \STATE $M_{(k)}^{t+1}: w \mapsto (R_\Omega^t)_{(k)} w + U_k^t G_{(k)}^t (U_{-k}^t)^\top w$
        \STATE $\{(\tilde{\sigma}_i, \tilde{u}_i, \tilde{v}_i)\}_{i \in I_k}, \Phi(M_{(k)}^{t+1}) \leftarrow \operatorname{Algorithm~\ref{alg:component_extraction}}(M_{(k)}^{t+1})$
        \STATE $U_k^{t+1} \leftarrow \operatorname{qr}([\tilde{u}_1, \dots, \tilde{u}_{s_k}])$
    \ENDFOR
    \STATE $\mathcal{G}^{t+1} \leftarrow R_\Omega^t \times_1 (U_1^{t+1})^\top \dots \times_N (U_N^{t+1})^\top + \mathcal{G}^t \times_{j=1}^N ((U_j^{t+1})^\top U_j^t)$
    \STATE $R_\Omega^{t+1} \leftarrow y - P_\Omega\left(\mathcal{G}^{t+1} \times_1 U_1^{t+1} \dots \times_N U_N^{t+1}\right)$
    \IF{$\|R_\Omega^{t+1} - R_\Omega^t\|_2 / \|y\|_2 \le \epsilon_{\mathrm{tol}}$}
        \STATE \textbf{break}
    \ENDIF
\ENDFOR
\STATE \textbf{return} $z^{t+1} = (\{U_k^{t+1}\}_{k=1}^N, \mathcal{G}^{t+1}, R_\Omega^{t+1})$
\end{algorithmic}
\end{algorithm}

\section{Complete Proofs of Theorems and Propositions}
\label{sec:appendix_proofs}

\subsection{Notation, Conventions, and Standard Multilinear Identities}
\label{sec:appendix_proofs_notation}

This subsection fixes the notation and collects the standard multilinear identities that the proofs use. Throughout, $[N] := \{1, \dots, N\}$ and $\mathbb{V} := \mathbb{R}^{n_1 \times \dots \times n_N}$, and $\bar{n}$, $n_{-k}$, $\bar{r}$, $r_{-k}$ are as in Section~\ref{sec:method_redundancy}. For $\mathcal{Y} \in \mathbb{R}^{m_1 \times \dots \times m_N}$, the mode-$k$ unfolding $Y_{(k)} \in \mathbb{R}^{m_k \times m_{-k}}$ places entry $(i_1, \dots, i_N)$ in row $i_k$ and column $1 + \sum_{j \neq k} (i_j - 1) \prod_{l < j,\, l \neq k} m_l$ \citep{kolda2009tensor}. The mode-$j$ product with $A \in \mathbb{R}^{p \times m_j}$ is
\begin{equation}
(\mathcal{Y} \times_j A)_{i_1 \dots i_{j-1}\, \alpha\, i_{j+1} \dots i_N} := \sum_{i_j = 1}^{m_j} \mathcal{Y}_{i_1 \dots i_N}\, A_{\alpha i_j}.
\end{equation}
For matrices $A_j$ ($j \in [N]$), we write $A_{-k} := A_N \otimes \dots \otimes A_{k+1} \otimes A_{k-1} \otimes \dots \otimes A_1$ for their Kronecker product in reverse order with $k$ omitted. We use the following standard facts, which hold for all conformable matrices \citep[Secs.~2.5 and 4.1]{kolda2009tensor}:
\begin{align}
&\text{(F1)} \quad (\mathcal{Y} \times_j A)_{(j)} = A\, Y_{(j)}; \nonumber \\
&\text{(F2)} \quad \mathcal{Y} \times_j A \times_m B = \mathcal{Y} \times_m B \times_j A \ \ (j \neq m), \qquad \mathcal{Y} \times_j A \times_j B = \mathcal{Y} \times_j (BA); \nonumber \\
&\text{(F3)} \quad \big( \mathcal{Y} \times_1 A_1 \times_2 A_2 \dots \times_N A_N \big)_{(k)} = A_k\, Y_{(k)}\, A_{-k}^\top; \nonumber \\
&\text{(F4)} \quad \operatorname{vec}\big( \mathcal{Y} \times_1 A_1 \dots \times_N A_N \big) = (A_N \otimes \dots \otimes A_1)\, \operatorname{vec}(\mathcal{Y}), \nonumber
\end{align}
where $\operatorname{vec}$ stacks entries with the first index varying fastest, so that $\operatorname{vec}(\mathcal{Y}) = \operatorname{vec}(Y_{(1)})$. Facts (F1) and (F2) follow directly from the definition, and (F3) is the unfolding rule for multilinear products. (F4) follows by combining (F3) for $k = 1$ with $\operatorname{vec}(AXB) = (B^\top \otimes A)\operatorname{vec}(X)$ and the associativity of $\otimes$. Mode products are linear in $\mathcal{Y}$. We abbreviate $\mathcal{Y} \times_{j=1}^{N} A_j := \mathcal{Y} \times_1 A_1 \times_2 A_2 \dots \times_N A_N$, and we write $\mathcal{Y} \times_{j \neq k} A_j$ for the same product with mode $k$ omitted; (F2) makes this product well defined. For a subspace $\mathcal{U} \subseteq \mathbb{R}^{n_k}$ with orthonormal basis matrix $U$, $P_{\mathcal{U}} := UU^\top$ denotes the orthogonal projector, which depends only on $\mathcal{U}$. The operator $P_\Omega$ zeroes all entries outside $\Omega$, and $P_{\Omega^\perp} := \mathrm{Id} - P_\Omega$. We write $P_\Omega(y) \in \mathbb{V}$ for the tensor that carries the observations $y$ on $\Omega$ and is zero elsewhere. Thus, $R_\Omega = y - P_\Omega(D(z))$ in Algorithm~\ref{alg:compact_state_update} is shorthand for the sparse tensor $P_\Omega(y) - P_\Omega(D(z))$. Finally, we use the mixed-product rule $(A \otimes B)(C \otimes D) = (AC) \otimes (BD)$ and the transpose rule $(A \otimes B)^\top = A^\top \otimes B^\top$.

\subsection{Proof of Proposition~\ref{prop:algorithm1_exactness} (Spectral Sufficiency) and Krylov Perturbation Analysis}
\label{sec:appendix_proofs_lemma1}

Let $M = \sum_{i=1}^{m} \sigma_i u_i v_i^\top$ be a singular value decomposition of $M \in \mathbb{R}^{n_k \times n_{-k}}$, where $m := \min(n_k, n_{-k})$, $\sigma_1 \ge \dots \ge \sigma_m \ge 0$, and $\{u_i\}$ and $\{v_i\}$ are orthonormal. We treat the three operators in turn and then the extraction step.

\paragraph{(i) Singular value thresholding.}
For $\lambda > 0$, the objective $X \mapsto \tfrac12 \|X - M\|_F^2 + \lambda \|X\|_*$ is strongly convex, so $\mathcal{D}_\lambda(M)$ is unique. By Theorem~2.1 of \cite{cai2010singular},
\begin{equation}
\mathcal{D}_\lambda(M) = \sum_{i=1}^{m} \max(0, \sigma_i - \lambda)\, u_i v_i^\top.
\label{eq:svt_full}
\end{equation}
Every index $i \notin I_\lambda(M) = \{i : \sigma_i > \lambda\}$ has $\sigma_i \le \lambda$ and hence $\max(0, \sigma_i - \lambda) = 0$. Therefore, \eqref{eq:svt_full} reduces to the first identity in \eqref{eq:spectral_identities}. Next, we show that $\mathcal{D}_\lambda(M)$ does not depend on the choice of singular bases. Put $g(\mu) := \max(0, 1 - \lambda/\sqrt{\mu})$ for $\mu > 0$ and $g(0) := 0$. Then $M\, g(M^\top M) = \sum_i \sigma_i\, g(\sigma_i^2)\, u_i v_i^\top = \sum_i \max(0, \sigma_i - \lambda)\, u_i v_i^\top = \mathcal{D}_\lambda(M)$, where $g(M^\top M)$ is the matrix function of the symmetric matrix $M^\top M$. A matrix function of a symmetric matrix is defined through the spectral projectors of that matrix, so it does not depend on the eigenbasis chosen within an eigenspace \citep[Ch.~9]{golub2013matrix}. Hence, $\mathcal{D}_\lambda(M)$ is a function of $M$ alone. Specifically, the singular values above $\lambda$, together with any orthonormal bases of the (unique) invariant subspaces $\mathcal{U}_\lambda := \operatorname{span}\{u_i : \sigma_i > \lambda\}$ of $MM^\top$ and $\mathcal{V}_\lambda := \operatorname{span}\{v_i : \sigma_i > \lambda\}$ of $M^\top M$, reproduce $\mathcal{D}_\lambda(M)$.

\paragraph{(ii) Rank-$r$ projection.}
By the Eckart--Young--Mirsky theorem \citep{eckart1936approximation}, $\sum_{i \le r} \sigma_i u_i v_i^\top$ minimizes $\|X - M\|_F$ over $\operatorname{rank}(X) \le r$, with $\|X - M\|_F^2 = \sum_{i > r} \sigma_i^2$. We now show that this minimizer is unique under $\sigma_r > \sigma_{r+1}$. Let $X$ be any minimizer, and let $\mathcal{C} := \operatorname{range}(X)$, so that $\dim \mathcal{C} \le r$. Among the matrices with column space in $\mathcal{C}$, $P_{\mathcal{C}} M$ is the closest to $M$. Therefore, $\|M - X\|_F^2 = \|P_{\mathcal{C}} M - X\|_F^2 + \|(I - P_{\mathcal{C}}) M\|_F^2 \ge \|M\|_F^2 - \operatorname{tr}(P_{\mathcal{C}} M M^\top)$. Write $w_i := \|P_{\mathcal{C}} u_i\|^2 \in [0, 1]$; these weights satisfy $\sum_i w_i \le \operatorname{tr}(P_{\mathcal{C}}) \le r$. Then $\operatorname{tr}(P_{\mathcal{C}} M M^\top) = \sum_i \sigma_i^2 w_i \le \sum_{i \le r} \sigma_i^2$. Moreover, $\sigma_i^2 \le \sigma_{r+1}^2 < \sigma_r^2$ for all $i > r$. Hence, any deficit $1 - w_i > 0$ at an index $i \le r$ can only be compensated by weight at indices with strictly smaller $\sigma_i^2$, and the last inequality is an equality only if $w_i = 1$ for all $i \le r$. Because $X$ is optimal, equality holds throughout, so $\|P_{\mathcal{C}} u_i\| = 1$, i.e.\ $u_i \in \mathcal{C}$, for $i \le r$. Since $\dim \mathcal{C} \le r$, it follows that $\mathcal{C} = \mathcal{U}_r := \operatorname{span}\{u_1, \dots, u_r\}$ and $X = P_{\mathcal{C}} M = P_{\mathcal{U}_r} M = \sum_{i \le r} \sigma_i u_i v_i^\top$. Thus, $\Pi_r(M) = P_{\mathcal{U}_r} M$ is unique and depends only on the invariant subspace $\mathcal{U}_r$, which is unique under the gap. In contrast, uniqueness fails if $\sigma_r = \sigma_{r+1}$ (e.g.\ $M = I_2$, $r = 1$). For this reason, the gap condition is part of the hypotheses of Theorem~\ref{thm:exact_equivalence}.

\paragraph{(ii') Linear minimization oracle.}
Let $\mathcal{B}_\tau := \{S : \|S\|_* \le \tau\}$, and let $G$ have the SVD $\sum_i \sigma_i(G) u_i v_i^\top$. The nuclear and spectral norms are dual, so $\langle S, G \rangle \ge -\|S\|_* \|G\|_2$ and hence $\min_{S \in \mathcal{B}_\tau} \langle S, G \rangle \ge -\tau \sigma_1(G)$. The matrix $S = -\tau u_1 v_1^\top$ attains this value, which proves \eqref{eq:spectral_identities}. We now show uniqueness under $\sigma_1(G) > \sigma_2(G)$. Let $S = \sum_j \nu_j a_j b_j^\top$ be an SVD of any minimizer ($\nu_j > 0$, $\sum_j \nu_j \le \tau$). Then $-\tau \sigma_1(G) = \sum_j \nu_j\, a_j^\top G b_j \ge -\sum_j \nu_j\, \sigma_1(G) \ge -\tau \sigma_1(G)$, so $\sum_j \nu_j = \tau$ and $a_j^\top G b_j = -\sigma_1(G)$ for every $j$. For unit vectors, $-a^\top G b \le \|G b\| \le \sigma_1(G)$. Equality holds only if $-a \parallel Gb$ and $b$ lies in the top right singular subspace, which is $\operatorname{span}\{v_1\}$ under the gap. Hence, $a_j b_j^\top = -u_1 v_1^\top$ for all $j$, and $S = -\tau u_1 v_1^\top$.

\paragraph{(iii) Exact extraction.}
By (i)--(ii'), $\Phi(M)$ depends only on the singular triplets indexed by $I_\Phi(M)$. For $\mathcal{D}_\lambda$ and $\Pi_r$, it depends even only on the corresponding invariant subspaces and singular values. Therefore, any procedure that returns these triplets exactly reproduces $\Phi(M)$, and the choice of procedure does not matter for Theorem~\ref{thm:exact_equivalence}. Algorithm~\ref{alg:component_extraction} is one such procedure, and its cost has three parts \citep[Ch.~10]{golub2013matrix}. First, $q$ steps of Golub--Kahan--Lanczos bidiagonalization cost $q$ matvec pairs $c(M)$ plus $O(q^2 (n_k + n_{-k}))$ for full reorthogonalization. Second, the Ritz triplets follow from the SVD of a $q \times q$ bidiagonal matrix in $O(q^3)$. Third, $s \le q$ Ritz vectors are assembled in $O(q s (n_k + n_{-k}))$. Together, these parts give \eqref{eq:component_complexity}.

\begin{remark}[Multiplicities and finite precision]
\label{rem:krylov}
(a) In exact arithmetic, a single-vector Krylov process reaches the invariant subspaces indexed by $I_\Phi$ after finitely many steps only when the required singular values are simple. A singular value of multiplicity $\mu > 1$ requires either a block Krylov process of block size at least $\mu$ or restarts with deflation, as implemented in standard partial-SVD routines. We understand Algorithm~\ref{alg:component_extraction} to include such deflation.
(b) In finite precision, let the returned Ritz values satisfy $|\tilde{\sigma}_i - \sigma_i| \le \epsilon$ for $i \le s + 1$, where $s = |I_\lambda(M)|$. If $\epsilon < \min(\sigma_s - \lambda,\ \lambda - \sigma_{s+1})$, then $\tilde{\sigma}_s > \lambda \ge \tilde{\sigma}_{s+1}$, so the index set $I_\lambda$ is identified exactly. Therefore, the stopping rule of Algorithm~\ref{alg:component_extraction} needs a tolerance below the distance of $\lambda$ to the spectrum, not a fixed fraction of $\lambda$.
(c) The Davis--Kahan $\sin\Theta$ theorem \citep{davis1970rotation} bounds the accuracy of the extracted subspace. Specifically, let $\tilde{U} \in \mathbb{R}^{n_k \times s}$ have orthonormal columns, and let $T := \tilde{U}^\top M M^\top \tilde{U}$ and $R_{\mathrm{Ritz}} := M M^\top \tilde{U} - \tilde{U} T$. If the eigenvalues of $T$ lie in $[a, b]$ and the eigenvalues of $M M^\top$ other than $\sigma_1^2, \dots, \sigma_s^2$ lie outside $(a - \delta, b + \delta)$ for some $\delta > 0$, then $\|\sin\Theta(\mathcal{U}_s, \operatorname{range} \tilde{U})\|_2 \le \|R_{\mathrm{Ritz}}\|_2 / \delta$. By Kaniel--Paige--Saad theory \citep{saad1980rates}, the Ritz residual $\|R_{\mathrm{Ritz}}\|_2$ decreases geometrically in the Krylov dimension $q$, at a rate governed by the relative spectral gap. Tightening $\epsilon_{\mathrm{krylov}}$ drives the residual perturbation observed for E2 in Table~\ref{tab:exactness} to machine precision, which we attribute to this geometric decrease.
\end{remark}

\subsection{Proof of Proposition~\ref{prop:algorithm2_equivalence} (Kronecker-Free Contraction)}
\label{sec:appendix_proofs_lemma2}

\paragraph{Correctness for every order $\pi$.}
By induction on $l$, $\mathcal{W}^{(l)} = \mathcal{Y} \times_{\pi_1} A_{\pi_1}^\top \dots \times_{\pi_l} A_{\pi_l}^\top$. By the commutation rule in (F2), the result of $N-1$ products along distinct modes does not depend on their order. Hence, $\mathcal{W}^{(N-1)} = \mathcal{Y} \times_{j \neq k} A_j^\top$ for every $\pi \in \mathfrak{S}_{N-1}$. Applying (F3) with $B_j := A_j^\top$ for $j \neq k$ and $B_k := I_{m_k}$, we obtain
\begin{equation}
\big( \mathcal{Y} \times_{j \neq k} A_j^\top \big)_{(k)} = Y_{(k)}\, \big( A_N^\top \otimes \dots \otimes A_{k+1}^\top \otimes A_{k-1}^\top \otimes \dots \otimes A_1^\top \big)^\top = Y_{(k)}\, A_{-k},
\end{equation}
where we used $(A \otimes B)^\top = A^\top \otimes B^\top$. This proves $Z_{(k)} = Y_{(k)} A_{-k}$. Moreover, the algorithm never forms any entry of $A_{-k}$.

\paragraph{Memory and arithmetic (dense input).}
Assume $m'_j \le m_j$ for all $j \ne k$. The intermediate $\mathcal{W}^{(l)}$ has $\prod_{j \in \{\pi_1, \dots, \pi_l\}} m'_j \prod_{j \notin \{\pi_1, \dots, \pi_l\}} m_j \le \bar{m}$ entries. Only $\mathcal{W}^{(l-1)}$ and $\mathcal{W}^{(l)}$ need to be stored at the same time, so the peak scratch is at most $2\bar{m}$ entries. Step $l$ computes $|\mathcal{W}^{(l-1)}|\, m'_{\pi_l} / m_{\pi_l}$ output entries, each an inner product of length $m_{\pi_l}$. Its cost is therefore $2 |\mathcal{W}^{(l-1)}| m'_{\pi_l} \le 2 \bar{m} \max_{j \ne k} m'_j$. Summing over $l = 1, \dots, N-1$ gives the bound $2(N-1)\bar{m} \max_{j \neq k} m'_j$. Algorithm~\ref{alg:factorwise_contraction} sorts the modes by descending $m_j / m'_j$. This order shrinks the intermediate as fast as possible in the first step, because it minimizes $|\mathcal{W}^{(1)}| = \bar{m}\, m'_{\pi_1} / m_{\pi_1}$. However, the correctness and the bounds above hold for every $\pi$.

\paragraph{Sparse input.}
If $\mathcal{Y}$ has non-zero support $\Omega$, we can accumulate the output $Z_{(k)} \in \mathbb{R}^{m_k \times m'_{-k}}$ directly. Specifically, for each $(i_1, \dots, i_N) \in \Omega$ with value $y$, we add $y \cdot \big( \bigotimes_{j \neq k} A_j[i_j, :] \big)$ (in the reverse order of $A_{-k}$) to row $i_k$ of $Z_{(k)}$. Forming the outer product of the $N-1$ row vectors incrementally costs $\sum_{l=2}^{N-1} \prod_{l' \le l} m'_{\pi_{l'}} \le (N-2) m'_{-k}$ multiplications, and the accumulation costs $2 m'_{-k}$ FLOPs. The total is therefore at most $2(N-1)|\Omega| m'_{-k} = O(N |\Omega| m'_{-k})$, with scratch $O(m'_{-k})$ beyond the input and the output. Correctness follows from the entrywise definition of the mode products, $(\mathcal{Y} \times_{j \neq k} A_j^\top)_{i_k, (\alpha_j)_{j \neq k}} = \sum_{(i_j)_{j \ne k}} \mathcal{Y}_{i_1 \dots i_N} \prod_{j \neq k} A_j[i_j, \alpha_j]$: only the entries $(i_1, \dots, i_N) \in \Omega$ contribute to this sum.

\paragraph{Orthogonality and unfolding identities.}
By the mixed-product rule, $U_{-k}^\top U_{-k} = \bigotimes_{j \neq k} (U_j^\top U_j)$ (reverse order). Since $U_j^\top U_j = \mathrm{Id}_{r_j}$, this gives $U_{-k}^\top U_{-k} = \mathrm{Id}_{r_{-k}}$. For $\mathcal{X} = \mathcal{G} \times_1 U_1 \dots \times_N U_N$, (F3) yields $X_{(k)} = U_k G_{(k)} U_{-k}^\top$, and hence $X_{(k)} U_{-k} = U_k G_{(k)}$. Finally, by (F2), $\mathcal{X} \times_{j \neq k} U_j^\top = \mathcal{G} \times_k U_k \times_{j \neq k} (U_j^\top U_j) = \mathcal{G} \times_k U_k$, and by (F1) the mode-$k$ unfolding of this tensor is $U_k G_{(k)}$. This proves \eqref{eq:kron_a_b}. \qed

\begin{lemma}[Implicit Kronecker matvecs]
\label{lem:kron_matvec}
Let $U_j \in \mathbb{R}^{n_j \times r_j}$ with $r_j \le n_j$ for $j \neq k$, and let $r_{\max} := \max_j r_j$. For $w \in \mathbb{R}^{n_{-k}}$ and $c \in \mathbb{R}^{r_{-k}}$, Algorithm~\ref{alg:factorwise_contraction} (applied to $w$ and $c$ reshaped as order-$(N-1)$ tensors) can evaluate the products $U_{-k}^\top w$ and $U_{-k} c$ without forming $U_{-k}$. Each product costs at most $2(N-1) n_{-k} r_{\max}$ FLOPs and needs $O(n_{-k})$ scratch.
\end{lemma}
\begin{proof}
First, consider $U_{-k}^\top w$. Reshape $w$ as $\mathcal{W} \in \mathbb{R}^{\times_{j \ne k} n_j}$, so that $\operatorname{vec}(\mathcal{W}) = w$. By (F4), $U_{-k}^\top w = \operatorname{vec}(\mathcal{W} \times_{j \neq k} U_j^\top)$, and Proposition~\ref{prop:algorithm2_equivalence} with $\bar{m} = n_{-k}$ gives the bound. Second, consider $U_{-k} c$. Reshape $c$ as $\mathcal{C} \in \mathbb{R}^{\times_{j \neq k} r_j}$; then $U_{-k} c = \operatorname{vec}(\mathcal{C} \times_{j \ne k} U_j)$ by (F4). We contract the modes in the order $\pi$. The $l$-th step then costs $2 \cdot |\text{current}| \cdot n_{\pi_l} = 2\, n_{-k}\, r_{\pi_l} \prod_{l' > l} (r_{\pi_{l'}} / n_{\pi_{l'}}) \le 2 n_{-k} r_{\max}$, and the largest intermediate has $n_{-k}$ entries.
\end{proof}

\subsection{Proof of Proposition~\ref{prop:algorithm3_commutation} (Compact State Intertwining)}
\label{sec:appendix_proofs_lemma3}

Let $z = (\{U_k\}_{k=1}^N, \mathcal{G}, R_\Omega)$ with $D(z) = \mathcal{G} \times_1 U_1 \dots \times_N U_N =: \mathcal{X}$ and $R_\Omega = P_\Omega(y) - P_\Omega(\mathcal{X})$. We write $z^+ = (\{U_k^+\}, \mathcal{G}^+, R_\Omega^+)$ for the output of one iteration (lines~2--8) of Algorithm~\ref{alg:compact_state_update}. We prove the claim in four steps and then count the storage.

\paragraph{Step 1: the operator input is sparse-plus-low-rank.}
The projected imputation input in \eqref{eq:proj_iteration} satisfies
\begin{equation}
\mathcal{M}^+ = P_\Omega(y) + P_{\Omega^\perp}(\mathcal{X}) = P_\Omega(y) - P_\Omega(\mathcal{X}) + \mathcal{X} = R_\Omega + D(z),
\label{eq:splr_input}
\end{equation}
so by (F3) and the linearity of unfolding, $M^+_{(k)} = (R_\Omega)_{(k)} + U_k G_{(k)} U_{-k}^\top$. Consequently, for $w \in \mathbb{R}^{n_{-k}}$ and $v \in \mathbb{R}^{n_k}$,
\begin{equation}
M^+_{(k)} w = (R_\Omega)_{(k)} w + U_k \big( G_{(k)} ( U_{-k}^\top w ) \big), \qquad (M^+_{(k)})^\top v = (R_\Omega)_{(k)}^\top v + U_{-k} \big( G_{(k)}^\top ( U_k^\top v ) \big),
\label{eq:splr_matvec}
\end{equation}
which is exactly the operator defined in line~3 of Algorithm~\ref{alg:compact_state_update}. We now count the cost of each term. The sparse terms cost $2|\Omega|$ FLOPs each (one multiply-add per stored entry). By Lemma~\ref{lem:kron_matvec}, $U_{-k}^\top w$ and $U_{-k}(\cdot)$ cost at most $2(N-1) n_{-k} r_{\max}$. The products $G_{(k)}(\cdot)$ and $G_{(k)}^\top(\cdot)$ cost $2\bar{r}$, and $U_k(\cdot)$ and $U_k^\top(\cdot)$ cost $2 n_k r_k$. Hence, one matvec pair costs $O(|\Omega| + n_k r_k + \bar{r} + N n_{-k} r_{\max})$, and no $n_k \times n_{-k}$ array is formed.

\paragraph{Step 2: the factor update yields the selected subspaces.}
By assumption, line~4 returns the exact singular triplets of $M^+_{(k)}$ indexed by $I_k := I_\Phi(M^+_{(k)})$. The vectors $\tilde{u}_1, \dots, \tilde{u}_{s_k}$ are orthonormal and span $\mathcal{U}_k^+ := \operatorname{span}\{u_i(M^+_{(k)}) : i \in I_k\}$. By Proposition~\ref{prop:algorithm1_exactness}, this subspace is unique (for $\Pi_r$ and $\mathrm{LMO}_\tau$ under the gap condition). Line~5 replaces this basis by the orthonormal factor of its QR decomposition, which spans the same subspace. Therefore, $\operatorname{range}(U_k^+) = \mathcal{U}_k^+$ and $U_k^+ (U_k^+)^\top = P_{\mathcal{U}_k^+}$, irrespective of the sign or basis convention of the QR routine. (If $I_k = \emptyset$, $U_k^+$ is the empty matrix, $P_{\mathcal{U}_k^+} = 0$, and both iterations produce $\mathcal{X}^+ = 0$.) As in \eqref{eq:proj_iteration}, all $N$ factors are extracted from the same tensor $\mathcal{M}^+$.

\paragraph{Step 3: the core update is the projected input.}
Let $W_j := (U_j^+)^\top U_j \in \mathbb{R}^{r_j^+ \times r_j}$. By the same-mode rule in (F2), $\mathcal{G} \times_j W_j = \mathcal{G} \times_j U_j \times_j (U_j^+)^\top$. By the commutation rule, the $N$ mode products can be interleaved, so
\begin{equation}
\mathcal{G} \times_{j=1}^{N} W_j = \big( \mathcal{G} \times_{j=1}^{N} U_j \big) \times_{j=1}^{N} (U_j^+)^\top = D(z) \times_{j=1}^{N} (U_j^+)^\top .
\end{equation}
Therefore, by the linearity of mode products and \eqref{eq:splr_input}, line~7 gives
\begin{equation}
\mathcal{G}^+ = \big( R_\Omega + D(z) \big) \times_1 (U_1^+)^\top \dots \times_N (U_N^+)^\top = \mathcal{M}^+ \times_1 (U_1^+)^\top \dots \times_N (U_N^+)^\top .
\end{equation}

\paragraph{Step 4: decoding.}
Using (F2) once more, we obtain
\begin{equation}
D(z^+) = \mathcal{G}^+ \times_{j=1}^{N} U_j^+ = \mathcal{M}^+ \times_{j=1}^{N} \big( U_j^+ (U_j^+)^\top \big) = \mathcal{M}^+ \times_{j=1}^{N} P_{\mathcal{U}_j^+} = \mathcal{T}_{\mathrm{proj}}(\mathcal{X}),
\end{equation}
which is the intertwining relation $D(\tilde{\mathcal{T}}(z)) = \mathcal{T}_{\mathrm{proj}}(D(z))$. Line~8 sets $R_\Omega^+ = P_\Omega(y) - P_\Omega(D(z^+))$, so the iteration maintains the invariant $R_\Omega = y - P_\Omega(D(z))$. As a result, the stopping test in line~9 depends only on $D(z^+)$, $D(z)$ and $y$. By construction, the factors $U_k^+$ have orthonormal columns.

\paragraph{Storage.}
The persistent state stores three objects: $\{U_k\}$ ($\sum_k n_k r_k$ words), $\mathcal{G}$ ($\bar{r}$ words), and $R_\Omega$. In coordinate format, $R_\Omega$ takes $|\Omega|$ values plus $N|\Omega|$ index words; the index part is shared with the data $(\Omega, y)$ and can be stored once. Together, these counts give \eqref{eq:memory_compression}. Line~8 needs $P_\Omega(D(z))$. We evaluate each queried entry by contracting the core sequentially with the rows $U_1[i_1, :], \dots, U_N[i_N, :]$, where the $m$-th contraction costs $2\bar{r} / \prod_{l < m} r_l \le 2\bar{r}$. Each entry therefore costs at most $2N\bar{r}$ FLOPs, for a total of $O(N|\Omega|\bar{r})$, and $D(z)$ is never materialized. The ratio $\zeta$ in \eqref{eq:memory_compression} tends to one exactly when $(\sum_k n_k r_k + \bar{r} + |\Omega|)/\bar{n} \to 0$. For $|\Omega| = p\bar{n}$, this requires $p \to 0$. Moreover, the sparse residual stops being advantageous when $(N+1)|\Omega| \ge \bar{n}$, i.e.\ when $p \ge 1/(N+1)$ in words. The threshold $p^*$ of Section~\ref{sec:discussion} refines this bound by the index/value byte ratio. \qed

\subsection{Proof of Theorem~\ref{thm:exact_equivalence} (Exact Algebraic Trajectory Equivalence)}
\label{sec:appendix_proofs_theorem1}

The proof rests on two abstract facts, which we state first.

\begin{lemma}[Intertwining implies trajectory equality]
\label{lem:intertwining_induction}
Let $(\mathcal{Z}, D, \tilde{\mathcal{T}})$ satisfy $D \circ \tilde{\mathcal{T}} = \mathcal{T} \circ D$ on a subset $\mathcal{Z}_0 \subseteq \mathcal{Z}$ with $\tilde{\mathcal{T}}(\mathcal{Z}_0) \subseteq \mathcal{Z}_0$. If $z^0 \in \mathcal{Z}_0$, $D(z^0) = \mathcal{X}^0$, $z^{t+1} = \tilde{\mathcal{T}}(z^t)$ and $\mathcal{X}^{t+1} = \mathcal{T}(\mathcal{X}^t)$, then $D(z^t) = \mathcal{X}^t$ for all $t \ge 0$.
\end{lemma}
\begin{proof}
We use induction on $t$. The base case is the hypothesis. If $D(z^t) = \mathcal{X}^t$ with $z^t \in \mathcal{Z}_0$, then $D(z^{t+1}) = D(\tilde{\mathcal{T}}(z^t)) = \mathcal{T}(D(z^t)) = \mathcal{T}(\mathcal{X}^t) = \mathcal{X}^{t+1}$, and $z^{t+1} \in \mathcal{Z}_0$ by forward invariance.
\end{proof}

\begin{lemma}[Compositionality]
\label{lem:compositionality}
If $D \circ \tilde{\mathcal{T}}_1 = \mathcal{T}_1 \circ D$ and $D \circ \tilde{\mathcal{T}}_2 = \mathcal{T}_2 \circ D$, then $D \circ (\tilde{\mathcal{T}}_2 \circ \tilde{\mathcal{T}}_1) = (\mathcal{T}_2 \circ \mathcal{T}_1) \circ D$.
\end{lemma}
\begin{proof}
$D \circ \tilde{\mathcal{T}}_2 \circ \tilde{\mathcal{T}}_1 = \mathcal{T}_2 \circ D \circ \tilde{\mathcal{T}}_1 = \mathcal{T}_2 \circ \mathcal{T}_1 \circ D$.
\end{proof}

\begin{proof}[Proof of Theorem~\ref{thm:exact_equivalence}]
Fix $\mathcal{T} \in \mathfrak{T}$ with its stage decomposition $\mathcal{T} = \mathcal{T}_U \circ \mathcal{T}_L \circ \mathcal{T}_S$, its state space $\mathcal{Z}$, and its decoding $D$. We verify the hypothesis of Lemma~\ref{lem:intertwining_induction} stage by stage on the forward-invariant set $\mathcal{Z}_0 := \{z \in \mathcal{Z} : R_\Omega = y - P_\Omega(D(z))\}$ (for states that carry a residual; otherwise $\mathcal{Z}_0 := \mathcal{Z}$).

\emph{Stage (S).} The reference stage evaluates $\Phi$ on the unfoldings $M_{(k)}$ of the input tensor $\mathcal{M} = \mathcal{M}(\mathcal{X}^t)$. The lightweight stage instead runs Algorithm~\ref{alg:component_extraction} on the implicit operator \eqref{eq:splr_matvec} (or on its factored analogue in Appendix~\ref{sec:appendix_alg_derivations}). By \eqref{eq:splr_input}, this operator equals $M_{(k)}$ as a linear map. By assumption, the oracle returns the singular triplets indexed by $I_\Phi(M_{(k)})$. By Proposition~\ref{prop:algorithm1_exactness}, these triplets determine $\Phi(M_{(k)})$ uniquely: for $\mathcal{D}_\lambda$ unconditionally, and for $\Pi_r$ and $\mathrm{LMO}_\tau$ under the gap condition. We assume this condition for every matrix on which $\Phi$ is evaluated along the reference trajectory; without it, the reference map itself is not single-valued. Hence, both iterations compute the same $\Phi(M_{(k)})$ (or the same selected subspaces $\mathcal{U}_k$).

\emph{Stage (L).} The reference stage contracts explicit unfoldings with Kronecker factors $U_{-k}$. The lightweight stage replaces every such contraction by Algorithm~\ref{alg:factorwise_contraction} or Lemma~\ref{lem:kron_matvec}. By Proposition~\ref{prop:algorithm2_equivalence}, the outputs coincide entrywise for every contraction order.

\emph{Stage (U).} For $\mathcal{T} = \mathcal{T}_{\mathrm{proj}}$, the compact update is Algorithm~\ref{alg:compact_state_update}. Proposition~\ref{prop:algorithm3_commutation} gives $D(\tilde{\mathcal{T}}(z)) = \mathcal{T}_{\mathrm{proj}}(D(z))$ together with the invariance of $\mathcal{Z}_0$. For the remaining members of $\mathfrak{T}$ used in this paper, Propositions~\ref{prop:svt_state}--\ref{prop:scaledgd_state} in Appendix~\ref{sec:appendix_alg_derivations} give the corresponding intertwining relations.

Each stage intertwines with its reference counterpart. Therefore, Lemma~\ref{lem:compositionality} (applied twice) yields $D \circ \tilde{\mathcal{T}} = \mathcal{T} \circ D$ on $\mathcal{Z}_0$, and Lemma~\ref{lem:intertwining_induction} gives $D(z^t) = \mathcal{X}^t$ for all $t \ge 0$. Finally, consider a stopping rule that is a function of $\{\mathcal{X}^s\}_{s \le t}$ and $(\Omega, y)$. This rule evaluates to the same Boolean value on $\{D(z^s)\}_{s \le t}$, so both iterations stop at the same index. In Algorithm~\ref{alg:compact_state_update}, the rule depends only on $R_\Omega^{t+1} - R_\Omega^t = P_\Omega(\mathcal{X}^t - \mathcal{X}^{t+1})$.
\end{proof}

\begin{remark}
Theorem~\ref{thm:exact_equivalence} is a statement about exact arithmetic. In finite precision, the discrepancy measured in Section~\ref{sec:five_exact} is the sum of two contributions: the rounding errors in the two implementations and the Ritz residual of Algorithm~\ref{alg:component_extraction}. This discrepancy is controlled as described in Remark~\ref{rem:krylov}. The gap conditions hold for the geometrically decaying spectra used in the experiments.
\end{remark}

\subsection{Proof of Corollary~\ref{cor:guarantee_inheritance} (Inheritance of Guarantees and Compositionality)}
\label{sec:appendix_proofs_corollary1}

(i) By Theorem~\ref{thm:exact_equivalence}, the decoded sequence $\{D(z^t)\}_{t \ge 0}$ and the reference sequence $\{\mathcal{X}^t\}_{t \ge 0}$ are the same sequence of tensors. Any guarantee of the reference estimator is a statement about this sequence and the data. Examples include $f(\mathcal{X}^t) - f^\star \le C\theta(t)$ for a convex objective $f$; a contraction $\|\mathcal{X}^{t+1} - \mathcal{X}^\star\|_F \le (1 - \mu)\|\mathcal{X}^t - \mathcal{X}^\star\|_F$ with $\mu$ independent of the condition number, as established for ScaledGD \citep{tong2022scaling} and PRGD \citep{zhang2025preconditioned}; and a minimax risk bound $\mathbb{E}\|\mathcal{X}^T - \mathcal{X}^\star\|_F^2 \le \psi(n, r, |\Omega|)$ under a sampling model. Therefore, every such guarantee holds verbatim with $\mathcal{X}^t$ replaced by $D(z^t)$. In contrast, guarantees about \emph{running time} are of course not inherited; Theorem~\ref{thm:complexity} replaces them.

(ii) This part is Lemma~\ref{lem:compositionality}. Concretely, let $\mathcal{T}' := \mathcal{T} \circ \mathcal{S}$, where $\mathcal{S}$ is a sketching step (Strategy~IV) or a preconditioning step (Strategy~V) that maps sparse-plus-factored states to sparse-plus-factored states. An example is the scaling by inverse Gram matrices in Appendix~\ref{sec:appendix_alg_prgd}. If $\mathcal{S}$ admits a lightweight realization $\tilde{\mathcal{S}}$ on the same $(\mathcal{Z}, D)$, then $\tilde{\mathcal{T}} \circ \tilde{\mathcal{S}}$ realizes $\mathcal{T}'$ exactly. We add two clarifications. First, the statement concerns the \emph{evaluation} of $\mathcal{T}'$. The trajectory of a sketched iteration differs from that of the unsketched one, and nothing here claims otherwise. We claim only that the lightweight evaluation of $\mathcal{T}'$ introduces no additional deviation. Second, these strategies and our algorithms save different resources. Sketching reduces the problem size, and preconditioning reduces the iteration count. In contrast, Algorithms~\ref{alg:component_extraction}--\ref{alg:compact_state_update} reduce the cost of executing a fixed iteration. The two savings can combine. However, whether they do depends on what the comparator already avoids: for PRGD, ours is lighter than the public code but not faster (Appendix~\ref{sec:appendix_mapext}). \qed

\section{Detailed Complexity Derivations and Arithmetic Counts}
\label{sec:appendix_complexity}

\subsection{Term-by-Term FLOP Accounting of Theorem~\ref{thm:complexity}}
\label{sec:appendix_complexity_flop}

We derive \eqref{eq:thm_flop} for one iteration (lines~2--8) of Algorithm~\ref{alg:compact_state_update}. Throughout, $r_k$ bounds the mode-$k$ rank of the state both before ($r_k^t$) and after ($r_k^{t+1} = s_k$) the iteration. Moreover, $r_{\max} = \max_k r_k$, $s = \max_k s_k \le r_{\max}$, and $s \le q \le \min_k \min(n_k, n_{-k})$. We count a multiply-add pair as two FLOPs and treat all $O(\cdot)$ constants as absolute.

\paragraph{Phase 1: sparse residual evaluation ($R_\Omega$).}
We obtain each queried entry of $D(z)$ by contracting the core with one row of each factor. Contracting mode $m$ after modes $1, \dots, m-1$ costs $2\bar{r} / \prod_{l < m} r_l \le 2\bar{r}$ FLOPs, so one entry costs at most $2N\bar{r}$ FLOPs (Appendix~\ref{sec:appendix_proofs_lemma3}). Across $|\Omega|$ entries,
\begin{equation}
\mathrm{FLOP}_{\mathrm{residual}} \le 2N|\Omega|\bar{r} = O(N|\Omega|\bar{r}).
\end{equation}

\paragraph{Phase 2: Krylov bidiagonalization with sparse-plus-low-rank matvecs.}
By Proposition~\ref{prop:algorithm3_commutation} and Lemma~\ref{lem:kron_matvec}, one matvec pair \eqref{eq:splr_matvec} for mode $k$ costs
\begin{equation}
c^{(k)} \le 4|\Omega| + 4 n_k r_k + 4\bar{r} + 4(N-1) n_{-k} r_{\max} = O\big(|\Omega| + n_k r_k + \bar{r} + N n_{-k} r_{\max}\big).
\end{equation}
Golub--Kahan--Lanczos bidiagonalization with $q$ steps and full reorthogonalization has four costs (Appendix~\ref{sec:appendix_proofs_lemma1}): $q$ matvec pairs; $O(q^2 (n_k + n_{-k}))$ FLOPs for reorthogonalization; $O(q^3)$ for the SVD of the $q \times q$ bidiagonal matrix; and $O(q s_k (n_k + n_{-k})) \le O(q^2 (n_k + n_{-k}))$ for assembling the $s_k$ Ritz vectors. Hence,
\begin{equation}
\mathrm{FLOP}_{\mathrm{Krylov}}^{(k)} = O\big( q\,(|\Omega| + n_k r_k + \bar{r} + N n_{-k} r_{\max}) + q^2 (n_k + n_{-k}) + q^3 \big).
\end{equation}
We sum over $k = 1, \dots, N$ and use $\bar{r} = r_k r_{-k} \le r_{\max} n_{-k}$ (as $r_j \le n_j$) and $q^3 \le q^2 n_{-k}$. This gives
\begin{equation}
\sum_{k=1}^N \mathrm{FLOP}_{\mathrm{Krylov}}^{(k)} = O\Big( qN|\Omega| + q\,(N r_{\max} + q) \sum_{k=1}^N (n_k + n_{-k}) \Big).
\label{eq:krylov_total}
\end{equation}
(With a block Krylov process of block size $b$ and $q/b$ block steps, the matvec term is unchanged, and the reorthogonalization term is also $O(q^2(n_k + n_{-k}))$. Hence, \eqref{eq:krylov_total} covers both variants.)

\paragraph{Phase 3: orthonormal factor retraction and core update.}
The QR factorization in line~5 costs $O(n_k s_k^2) \le O(n_k r_k^2)$ per mode. Line~7 has three parts. First, $R_\Omega \times_1 (U_1^{t+1})^\top \dots \times_N (U_N^{t+1})^\top$ is a sparse contraction over all $N$ modes, and by Proposition~\ref{prop:algorithm2_equivalence} it costs $O(N|\Omega|\bar{r})$. Second, the small matrices $W_j = (U_j^{t+1})^\top U_j^t$ cost $2 n_j r_j^{t+1} r_j^t \le 2 n_j r_j^2$. Third, $\mathcal{G}^t \times_1 W_1 \dots \times_N W_N$ costs $\sum_j 2 (\text{current size}) \cdot r_j^{t+1} \le 2\bar{r} \sum_j r_j$, because every intermediate has at most $\bar{r}$ entries. Line~8 is another Phase-1 evaluation. Altogether,
\begin{equation}
\mathrm{FLOP}_{\mathrm{update}} = O\Big( N|\Omega|\bar{r} + \sum_{k=1}^N n_k r_k^2 + \bar{r} \sum_{k=1}^N r_k \Big).
\end{equation}

Combining Phases~1--3 gives \eqref{eq:thm_flop}:
\begin{equation}
\mathrm{FLOP}(\tilde{\mathcal{T}}) = O\Big( qN|\Omega| + q\,(N r_{\max} + q) \sum_{k=1}^N (n_k + n_{-k}) + N|\Omega|\bar{r} + \sum_{k=1}^N n_k r_k^2 + \bar{r} \sum_{k=1}^N r_k \Big).
\end{equation}
For $n_k = n$, $r_k = r$ and fixed $N$, this bound is $O\big(q(q + r) n^{N-1} + (q + r^N)|\Omega|\big)$. In contrast, the literal implementation forms the unfolding $X_{(k)} \in \mathbb{R}^{n_k \times n_{-k}}$ and computes its full SVD at cost $\Theta(n_k n_{-k} \min(n_k, n_{-k}))$ \citep[Ch.~8]{golub2013matrix}. Per iteration, this cost is $\Theta(\bar{n} \sum_k \min(n_k, n_{-k}))$, i.e.\ $\Theta(N n^{N+1})$ for cubical tensors. Forming $U_{-k}$ and the product $X_{(k)} U_{-k}$ adds $\Theta(n_{-k} r_{-k} + \bar{n} r_{-k})$. Thus, for $q$ independent of $n$, the lightweight iteration is one polynomial order lower in $n$. However, for $q=\Theta(n)$, the two are of the same order. Appendix~\ref{sec:appendix_complexity_dominant} discusses which term dominates for a fixed sampling rate and reports measured exponents.

\subsection{Peak Memory Footprint Derivation of Equation~\eqref{eq:thm_mem}}
\label{sec:appendix_complexity_mem}

The memory of $\tilde{\mathcal{T}}$ splits into two parts: the persistent state carried between iterations and the transient scratch space.
\begin{enumerate}
    \item \textbf{Factor storage}: $\{U_k \in \mathbb{R}^{n_k \times r_k}\}_{k=1}^N$ consume $\sum_{k=1}^N n_k r_k$ words.
    \item \textbf{Core storage}: $\mathcal{G} \in \mathbb{R}^{r_1 \times \dots \times r_N}$ consumes $\bar{r}$ words.
    \item \textbf{Sparse residual storage}: one value and $N$ coordinates per observed entry, i.e.\ $(N+1)|\Omega|$ words. The coordinates coincide with those of the data and need not be duplicated.
    \item \textbf{Krylov scratch}: the $q$ left and $q$ right Lanczos vectors of lengths $n_k$ and $n_{-k}$, i.e.\ $q(n_k + n_{-k})$ words, plus $O(q^2)$ words for the bidiagonal matrix. These arrays are released after the mode-$k$ extraction.
    \item \textbf{Contraction scratch}: Lemma~\ref{lem:kron_matvec} needs $O(n_{-k})$ words, and the sparse contraction of line~7 needs $O(\bar{r})$ words.
    \item \textbf{Index structures}: the sparse products keep per-mode index patterns of the observed entries, which take $O(N|\Omega|)$ words. In our GPU implementation, the measured peak of ScaledGD at $n=\benchNmax$ is about \mxWordsObs{} words per observed entry, including the $N+1$ words of the data. Therefore, at a fixed sampling rate $p$, this peak is below one ambient array only for $p<1/\mxWordsObs$ (Appendix~\ref{sec:appendix_mapext}).
\end{enumerate}
Items~1--3 give $\mathrm{MEM}_{\mathrm{state}}(\tilde{\mathcal{T}}) = \sum_k n_k r_k + \bar{r} + (N+1)|\Omega|$, and items~4--6 give the peak
\begin{equation}
\mathrm{MEM}_{\max}(\tilde{\mathcal{T}}) = \mathrm{MEM}_{\mathrm{state}}(\tilde{\mathcal{T}}) + O\Big( N|\Omega| + q \max_k (n_k + n_{-k}) \Big),
\end{equation}
which is \eqref{eq:thm_mem}. We add two comments. First, the Krylov scratch $q \max_k n_{-k} = q\bar{n} / \min_k n_k$ is not negligible relative to $\bar{n}$ unless $q \ll \min_k n_k$. This scratch is transient, and it is what separates $\mathrm{MEM}_{\max}$ from $\mathrm{MEM}_{\mathrm{state}}$. Therefore, Section~\ref{sec:five} reports the measured peak, which also contains the input data, next to $\mathrm{MEM}_{\mathrm{state}}$. Second, the scratch can be reduced to $O(q n_k + s n_{-k})$ by keeping only the short Lanczos vectors and recovering the $s$ required right vectors as $\tilde{v}_i = M^\top \tilde{u}_i / \tilde{\sigma}_i$. The price is $s$ extra transposed matvecs. For comparison, the literal implementation stores the ambient iterate and its unfolding ($\Theta(\bar{n})$), the explicit Kronecker factor $U_{-k}$ ($n_{-k} r_{-k}$ words), and the dense SVD workspace ($\Theta(n_k n_{-k})$).

\subsection{Systematic Comparison across Computational Paradigms}
\label{sec:appendix_complexity_comparison}

Table~\ref{tab:paradigm_comparison_appendix} compares four paradigms: naive ambient execution, randomized sketching, Riemannian optimization, and the proposed lightweight method. The comparison assumes cubical order-$N$ tensors with $n_k = n$ and $r_k = r$, and $d$ denotes the sketch size.

\begin{table}[H]
\caption{Analytical comparison of arithmetic, spatial, and optimization properties across computing paradigms.}
\label{tab:paradigm_comparison_appendix}
\begin{center}
\small
\begin{tabular}{p{0.16\textwidth} p{0.26\textwidth} p{0.22\textwidth} p{0.11\textwidth} p{0.10\textwidth}}
\toprule
\textbf{Paradigm} & \textbf{FLOP / Iteration} & \textbf{Peak Memory} & \textbf{Algebraic Loss} & $\kappa$-Sensitivity \\
\midrule
\textbf{Naive Ambient} & $\Theta(N n^{N+1})$ & $\Theta(n^N)$ & None (Exact) & Unchanged \\
\textbf{Randomized Sketching} & $O(|\Omega| + N n d^2)$ & $O(N n d + d^N)$ & Approximate & Unchanged \\
\textbf{Riemannian / ScaledGD} & $O(|\Omega| r^N + N n r^2)$ & $O(n^N)$ or $O(N n r + r^N)$ & None (Exact) & Invariant \\
\textbf{Proposed Method} & $O(q(q + r) n^{N-1} + (q + r^N)|\Omega|)$ & state $O(N n r + r^N + |\Omega|)$; peak $+\,O(q n^{N-1})$ & None (Exact) & Invariant \\
\bottomrule
\end{tabular}
\end{center}
\end{table}

\subsection{Which Term Dominates, and Measured Exponents}
\label{sec:appendix_complexity_dominant}

\emph{Which term dominates.} The bound of Theorem~\ref{thm:complexity} covers two regimes. First, with a fixed sampling rate $p=|\Omega|/\bar n$, the term $(q+r^N)|\Omega|=(q+r^N)\,p\,n^N$ dominates. One iteration then costs $\Theta\big((q+r^N)\,p\,n^N\big)$, against $\Omega(n^{N+1})$ for the literal implementation. For $q$ independent of $n$, this cost is one order lower in $n$, with a gain of order $n/\big((q+r^N)p\big)$. However, for $q=\Theta(n)$, as with nearly full-rank iterates, there is no gain in order. Second, the larger gain $n^2/\big(q(q+r)\big)$ holds only when $|\Omega|=O(n^{N-1})$. For fully observed data (denoising with HOOI or the subspace norm), contracting the data tensor costs $\Theta(r\,n^N)$ per contraction, and this cost dominates in every implementation. Reusing partial contractions changes the number of such contractions per sweep, but not their order. All of these are operation counts for large $n$. Measured times also contain lower-order terms with large constants and fixed per-call costs. Therefore, measured exponents approach these orders only as $n$ grows (see below).

\emph{Which estimators gain an order.} The order is lowered only where the reference implementation computes the full SVD of an unfolding. This is the case for Raw HaLRTC and Raw Frank--Wolfe, which cost $\Theta(n^4)$ per iteration at order 3. The public HaLRTC code does the same. In contrast, the public Frank--Wolfe code and ours use a sparse power or Lanczos method, which costs $\Theta(n^3)$ at a fixed $p$. For HaLRTC, ours costs $\Theta(n^3)$ per Krylov step, and the number of steps grows like $n^\gamma$; the predicted exponent is therefore $3+\gamma$. At $p=0.05$, where $q\approx n$, the order is not lowered (Section~\ref{sec:five_limits}). For the subspace norm, HOOI and ScaledGD, $\Theta(n^3)$ operations on the data or on the observed entries dominate every implementation, and the implementations differ by constant factors.

\emph{Measured exponents.} Table~\ref{tab:slopes} gives the local slope of the time per iteration against $n$ for Raw and ours on the A100, for $n=200$, $300$, $500$ and $700$. We use the code, data and protocol of Section~\ref{sec:five}, but with $T=3$ iterations per run. Except for HaLRTC with ours, the time per iteration at $n\le500$ agrees with the $T=10$ runs of Appendix~\ref{sec:appendix_tab_bench} to within 20\%. We make three observations.
(i) In every row with a value, the slope between $n=500$ and $700$ is at least that between $200$ and $300$, but it is still at or below the predicted exponent. For Raw HaLRTC and Frank--Wolfe, the slope rises from $\slRawLoLo$--$\slRawLoHi$ to $\slRawHiLo$--$\slRawHiHi$. A fit $t=c_0+c_3n^3+c_4n^4$ to the four sizes (largest relative error \slTermErr\%) suggests why. According to this fit, at $n=700$ the $n^3$ operations on the dense iterate and the $n^4$ SVD contribute about equally (the $n^4$ term is \slShareLo--\slShareHi\% of the time), and the $n^4$ term overtakes only at $n\approx\slNstarLo$--$\slNstarHi$. Therefore, measured slopes approach 4 slowly, and slopes fitted over moderate sizes fall well below the exponent. On one CPU thread, where fixed costs are small, the slope of Raw HaLRTC between $n=200$ and 300 is already \slCpuRawLo--\slCpuRawHi.
(ii) Where the order is lowered and the slope can be measured, ours has the smaller slope. For Frank--Wolfe between $n=500$ and $700$, the slope of ours is smaller by \slGapFWHi{} at $p=0.05$ and by \slGapFWLo{} at $p=0.2$.
(iii) The slopes of ours for the subspace norm, HOOI and Frank--Wolfe at $p=0.05$ (\slOursLo--\slOursHi) stay below 3. We attribute this to fixed per-call costs, which still partly bind ours at these sizes: HOOI takes \slHooiOursMs\,ms per iteration at $n=700$. In contrast, for the estimators without an order reduction, Raw has slopes of \slOtherRawLo--\slOtherRawHi{} (HOOI, ScaledGD), close to 3.

\begin{table}[H]
\caption{Measured exponents: local slope $\log(t_2/t_1)/\log(n_2/n_1)$ of the time per iteration $t$ between consecutive sizes (A100, order 3, $r=5$, $\kappa=100$, mean of 3 seeds, $T=3$ iterations per run; bootstrap 95\% interval over seeds for $500\to700$). Predicted: exponent of the dominant term at a fixed $p$; $\gamma$ is the growth exponent of the number of Krylov steps. $^\ast$Out of memory at $n=700$ (40\,GiB). $^\dagger$The time per iteration of ours grows over the first iterations with the rank of the iterate, so $T=3$ gives no stationary value. In the $T=10$ runs of Appendix~\ref{sec:appendix_tab_bench} ($n\le500$), the slope between $n=300$ and 500 is \slLiteOneHighOurs{} for ours against \slLiteOneHighRaw{} for Raw at $p=0.2$, and \slLiteOneLowOurs{} against \slLiteOneLowRaw{} at $p=0.05$, where $q\approx n$.}
\label{tab:slopes}
\begin{center}
\small
\begin{tabular}{llllrrl}
\toprule
Estimator & $p$ & & Predicted & $200\to300$ & $300\to500$ & $500\to700$ [95\%] \\
\midrule
E0 Subspace norm & -- & Raw & 3 & 1.44 & 2.44 & OOM$^\ast$ \\
 &  & Ours & 3 & 0.52 & 1.29 & 2.17 [2.07, 2.31] \\
\addlinespace[2pt]
E1 HaLRTC & 0.05 & Raw & 4 & 2.25 & 3.19 & 3.25 [3.25, 3.26] \\
 &  & Ours & $3+\gamma$ & --$^\dagger$ & --$^\dagger$ & --$^\dagger$ \\
\addlinespace[2pt]
E1 HaLRTC & 0.2 & Raw & 4 & 2.34 & 3.17 & 3.27 [3.27, 3.28] \\
 &  & Ours & $3+\gamma$ & --$^\dagger$ & --$^\dagger$ & --$^\dagger$ \\
\addlinespace[2pt]
E2 Frank--Wolfe & 0.05 & Raw & 4 & 2.23 & 3.19 & 3.28 [3.27, 3.29] \\
 &  & Ours & 3 & 0.53 & 1.47 & 1.94 [1.75, 2.17] \\
\addlinespace[2pt]
E2 Frank--Wolfe & 0.2 & Raw & 4 & 2.17 & 3.19 & 3.27 [3.26, 3.28] \\
 &  & Ours & 3 & 1.19 & 2.28 & 2.65 [2.54, 2.72] \\
\addlinespace[2pt]
E3 HOOI & -- & Raw & 3 & 1.40 & 2.17 & 2.62 [2.25, 3.00] \\
 &  & Ours & 3 & 0.04 & 0.51 & 1.32 [1.02, 1.51] \\
\addlinespace[2pt]
E4 ScaledGD & 0.05 & Raw & 3 & 2.29 & 3.02 & 2.94 [2.67, 3.23] \\
 &  & Ours & 3 & $-$0.03 & 1.12 & 2.78 [2.43, 3.15] \\
\addlinespace[2pt]
E4 ScaledGD & 0.2 & Raw & 3 & 2.28 & 3.23 & 2.54 [2.53, 2.54] \\
 &  & Ours & 3 & 0.82 & 3.01 & 2.81 [2.72, 2.90] \\
\bottomrule
\end{tabular}

\end{center}
\end{table}

\section{Concrete Algorithmic Instantiations}
\label{sec:appendix_alg_derivations}

For each estimator of Table~\ref{tab:exp1_estimators}, we specify the state $\mathcal{Z}$, the decoding $D$, the reference iteration $\mathcal{T}$, and the intertwining relation \eqref{eq:intertwining} that Theorem~\ref{thm:exact_equivalence} requires. Proposition~\ref{prop:algorithm3_commutation} already covers the projected imputation iteration $\mathcal{T}_{\mathrm{proj}}$ of \eqref{eq:proj_iteration}. However, not every iterate in $\mathfrak{T}$ has a small Tucker core, so we need two further state formats.

\paragraph{Sparse-plus-mode-factored states.}
Let $\mathcal{Z}_{\mathrm{mf}}$ be the set of tuples $z = \big( \mathcal{S}, \{(A_k, B_k)\}_{k=1}^N \big)$ with $\mathcal{S}$ supported on $\Omega$, $A_k \in \mathbb{R}^{n_k \times s_k}$, $B_k \in \mathbb{R}^{n_{-k} \times s_k}$ ($s_k \ge 0$), decoded by
\begin{equation}
D(z) := \mathcal{S} + \sum_{k=1}^N \operatorname{fold}_k\big( A_k B_k^\top \big),
\label{eq:mf_decode}
\end{equation}
where $\operatorname{fold}_k$ is the inverse of the mode-$k$ unfolding. Storing such a state takes $(N+1)|\Omega| + \sum_k s_k (n_k + n_{-k})$ words.

\begin{proposition}[Mode-wise proximal steps on factored states]
\label{prop:svt_state}
Let $z \in \mathcal{Z}_{\mathrm{mf}}$ and $\mathcal{M} := D(z)$.
\begin{enumerate}[leftmargin=*,noitemsep]
\item[(a)] For every $k$, one matvec pair with $M_{(k)}$ and $M_{(k)}^\top$ costs $O\big(|\Omega| + \sum_{j=1}^N s_j (n_{-j} + n_{-k})\big)$ FLOPs without forming $M_{(k)}$.
\item[(b)] $\operatorname{fold}_k(\mathcal{D}_\lambda(M_{(k)})) = \operatorname{fold}_k(A B^\top)$ with $A := U_I (\Sigma_I - \lambda\,\mathrm{Id})$, $B := V_I$, where $(U_I, \Sigma_I, V_I)$ are the singular triplets of $M_{(k)}$ indexed by $I_\lambda(M_{(k)})$. Likewise, $\operatorname{fold}_k(\Pi_r(M_{(k)}))$ and $\operatorname{fold}_k(\mathrm{LMO}_\tau(M_{(k)}))$ are of the form $\operatorname{fold}_k(AB^\top)$ with $s_k = r$ and $s_k = 1$, respectively.
\item[(c)] $\mathcal{Z}_{\mathrm{mf}}$ is closed under linear combinations (by concatenating the factors of equal modes and adding the sparse parts), under $P_\Omega$ and $P_{\Omega^\perp}$ (by adding a sparse correction), and under the mode-wise primitives in (b). Consequently, any iteration built from these operations maps $\mathcal{Z}_{\mathrm{mf}}$ to itself, and its lightweight realization satisfies $D \circ \tilde{\mathcal{T}} = \mathcal{T} \circ D$ with exact spectral extraction. Such iterations include the proximal-gradient step for sum-of-mode-wise nuclear norms; the HaLRTC step of \cite{liu2012tensor} (an ADMM method), whose primal, auxiliary, and dual variables are all of the form \eqref{eq:mf_decode}; and the proximal step of the subspace norm \citep{zheng2015interpolating}, in which $B_k$ is additionally constrained to a fixed Kronecker subspace.
\end{enumerate}
\end{proposition}
\begin{proof}
(a) We have $M_{(k)} w = S_{(k)} w + \sum_{j} \operatorname{fold}_j(A_j B_j^\top)_{(k)} w$. The sparse term costs $2|\Omega|$. For $j = k$, the term is $A_k (B_k^\top w)$, which costs $2 s_k (n_{-k} + n_k)$. For $j \neq k$, write $A_j B_j^\top = \sum_{\ell=1}^{s_j} a_\ell b_\ell^\top$. The tensor $\operatorname{fold}_j(a_\ell b_\ell^\top)$ has entries $a_\ell[i_j]\, \mathcal{B}_\ell[(i_l)_{l \neq j}]$, where $\mathcal{B}_\ell$ is $b_\ell$ reshaped over the modes $l \ne j$. We reshape $w$ into $\mathcal{W}$ over the modes $l \neq k$. Its product with the mode-$k$ unfolding is then $\sum_{i_j} a_\ell[i_j] \sum_{(i_l)_{l \ne j,k}} \mathcal{B}_\ell[i_k, (i_l)_{l \ne j,k}]\, \mathcal{W}[i_j, (i_l)_{l \ne j,k}]$. We evaluate this sum in two contractions. First, we contract $\mathcal{W}$ with $a_\ell$ along mode $j$ ($2 n_{-k}$ FLOPs). Second, we contract the result with $\mathcal{B}_\ell$ along all modes $l \neq j, k$ ($2 n_{-j}$ FLOPs). Summing over $\ell$ and $j$ gives the bound, and the argument for the transposed product is symmetric. (b) This part is Proposition~\ref{prop:algorithm1_exactness}. (c) Linear combinations and sparse corrections follow immediately from \eqref{eq:mf_decode}, and the mode-wise primitives are covered by (b). Thus, each reference operation is reproduced on the decoded tensors, i.e.\ intertwining holds operation by operation. Lemma~\ref{lem:compositionality} then composes these operations.
\end{proof}

\begin{remark}
(i) In the HaLRTC update, the dual variable $\mathcal{Y}_k^{t+1} = \mathcal{Y}_k^t - \rho(\mathcal{M}_k^{t+1} - \mathcal{X}^{t+1})$ is a running sum of factored terms. Its factor count therefore grows with $t$, unless the dual variable is stored in ambient form. Either choice preserves exactness and changes only the memory profile. Our implementation stores the dual variables in ambient form, so the state of E1 cannot be smaller than $(N+1)\bar{n}$ words (Section~\ref{sec:five_limits}). (ii) The matvec cost in (a) is $\Theta(\sum_j s_j n_{-j})$. Hence, for spectra that are not sparse ($s_j \approx n_j$), this cost is of the same order as for a dense unfolding. This is the regime in which E1 shows no gain in running time at $p=0.05$ (Section~\ref{sec:five_limits}). However, the memory reduction from \eqref{eq:mf_decode} is retained.
\end{remark}

\subsection{Subspace Norm and HaLRTC (E0, E1)}
\label{sec:appendix_alg_subspace}

Proposition~\ref{prop:svt_state} handles two regularizers. The overlapped (sum-of-nuclear-norms) regularizer \citep{liu2012tensor} penalizes $\sum_{k=1}^N \lambda_k \|X_{(k)}\|_*$. The subspace norm \citep{zheng2015interpolating} penalizes mode-wise nuclear norms of components constrained to fixed Kronecker subspaces. Algorithm~\ref{alg:lightweight_subspace_prox} states the generic lightweight proximal-gradient iteration on $\mathcal{Z}_{\mathrm{mf}}$ with step size $1$. The ADMM variant differs only in the bookkeeping of the auxiliary and dual variables, and by Proposition~\ref{prop:svt_state}(c) all of these variables stay in $\mathcal{Z}_{\mathrm{mf}}$.

\begin{algorithm}[H]
\caption{Lightweight Mode-Wise Proximal Estimation on Sparse-plus-Mode-Factored States}
\label{alg:lightweight_subspace_prox}
\begin{algorithmic}[1]
\REQUIRE Observed data $(\Omega, y)$, thresholds $\{\lambda_k\}_{k=1}^N$, initial state $z^0 \in \mathcal{Z}_{\mathrm{mf}}$, max iterations $T$.
\ENSURE State $z^T \in \mathcal{Z}_{\mathrm{mf}}$ with $D(z^T) = \mathcal{X}^T$.
\FOR{$t = 0, 1, \dots, T-1$}
    \STATE Sparse residual $R_\Omega^t \leftarrow P_\Omega(y) - P_\Omega(D(z^t))$ (each entry of $D(z^t)$ on $\Omega$ costs $O(\sum_j s_j)$).
    \STATE Gradient-step input $\mathcal{M}^{t} \leftarrow D(z^t) + R_\Omega^t$, kept in factored form (sparse part $\mathcal{S}^t + R_\Omega^t$).
    \FOR{each mode $k = 1, \dots, N$}
        \STATE Implicit matvec $\mathcal{M}_k(w) := M^t_{(k)} w$ evaluated as in Proposition~\ref{prop:svt_state}(a).
        \STATE $\{(\sigma_i, u_i, v_i)\}_{i \in I_{\lambda_k}} \leftarrow$ Algorithm~\ref{alg:component_extraction}$(\mathcal{M}_k, \mathcal{D}_{\lambda_k})$.
        \STATE $A_k^{t+1} \leftarrow [u_i (\sigma_i - \lambda_k)]_{i \in I_{\lambda_k}}$, \quad $B_k^{t+1} \leftarrow [v_i]_{i \in I_{\lambda_k}}$.
    \ENDFOR
    \STATE Assemble $z^{t+1}$ from $\{(A_k^{t+1}, B_k^{t+1})\}$ according to the estimator's combination rule (e.g.\ averaging over modes and re-imposing $P_\Omega(\cdot) = y$ through the sparse part).
    \STATE Check convergence $\|R_\Omega^{t+1} - R_\Omega^t\|_2 / \|y\|_2 \le \epsilon_{\mathrm{tol}}$.
\ENDFOR
\end{algorithmic}
\end{algorithm}

\subsection{Frank--Wolfe with Greedy Atom Pursuit (E2)}
\label{sec:appendix_alg_fw}

Consider nuclear-norm-constrained completion, $\min_{\|\mathcal{X}\|_* \le \tau} \tfrac12 \|P_\Omega(\mathcal{X}) - y\|_F^2$, where $\|\cdot\|_*$ is a mode-wise or averaged nuclear norm. Frank--Wolfe iterates $\mathcal{X}^{t+1} = (1 - \eta_t)\mathcal{X}^t + \eta_t \mathcal{A}^t$ with $\mathcal{A}^t := \operatorname{fold}_{k_t}(\mathrm{LMO}_\tau(\nabla f(\mathcal{X}^t)_{(k_t)}))$, for a selected mode $k_t$ and step size $\eta_t \in [0, 1]$ \citep{guo2017efficient}. The gradient $\nabla f(\mathcal{X}^t) = P_\Omega(\mathcal{X}^t) - y = -R_\Omega^t$ is supported on $\Omega$.

\begin{proposition}[Atom-pool state]
\label{prop:fw_state}
Let $\mathcal{Z}_{\mathrm{fw}}$ consist of atom pools $z = (\{(c_a, k_a, u_a, v_a)\}_{a=1}^{m}, R_\Omega)$ with $D(z) := \sum_a c_a \operatorname{fold}_{k_a}(u_a v_a^\top)$ and $R_\Omega = P_\Omega(y) - P_\Omega(D(z))$. The lightweight step has three parts. (i) It computes the LMO by Algorithm~\ref{alg:component_extraction} on the sparse operator $(R_\Omega)_{(k_t)}$, whose matvec costs $2|\Omega|$. (ii) It rescales all coefficients by $(1 - \eta_t)$ and appends the atom $(\eta_t \tau, k_t, u_1, v_1)$, where $(u_1, v_1)$ is the top singular pair of $(R_\Omega^t)_{(k_t)}$, so that $\mathcal{A}^t = \tau \operatorname{fold}_{k_t}(u_1 v_1^\top)$. (iii) It updates $R_\Omega \leftarrow (1 - \eta_t) R_\Omega + \eta_t \big( P_\Omega(y) - P_\Omega(\mathcal{A}^t) \big)$ in $O(|\Omega|)$. This step satisfies $D \circ \tilde{\mathcal{T}} = \mathcal{T} \circ D$ under $\sigma_1 > \sigma_2$ for $(R_\Omega^t)_{(k_t)}$. Moreover, $\mathcal{Z}_{\mathrm{fw}}$ is closed under the fully-corrective or line-search variants, which only re-weight the pool.
\end{proposition}
\begin{proof}
(i) Since $R_\Omega^t = P_\Omega(y) - P_\Omega(\mathcal{X}^t)$, we have $\nabla f(\mathcal{X}^t)_{(k_t)} = -(R_\Omega^t)_{(k_t)}$, so the LMO input is the sparse unfolding. By Proposition~\ref{prop:algorithm1_exactness}(ii'), $\mathrm{LMO}_\tau$ is $-\tau$ times the top singular pair of $-(R_\Omega^t)_{(k_t)}$. Since the top pair of $-R$ is $(u_1, -v_1)$, this output equals $\tau u_1 v_1^\top$ for the top pair $(u_1, v_1)$ of $(R^t_\Omega)_{(k_t)}$, and it is unique under the gap. (ii) $D$ is linear in the coefficients, so $D(z^{t+1}) = (1 - \eta_t) D(z^t) + \eta_t \tau \operatorname{fold}_{k_t}(u_1 v_1^\top) = \mathcal{X}^{t+1}$. (iii) We have $P_\Omega(y) - P_\Omega(\mathcal{X}^{t+1}) = (1 - \eta_t)(P_\Omega(y) - P_\Omega(\mathcal{X}^t)) + \eta_t (P_\Omega(y) - P_\Omega(\mathcal{A}^t))$, and $P_\Omega(\mathcal{A}^t)$ needs one product $u_1[i_{k_t}] v_1[\cdot]$ per observed entry.
\end{proof}

\subsection{Higher-Order Orthogonal Iteration (HOOI, E3)}
\label{sec:appendix_alg_hooi}

For a fully observed tensor $\mathcal{Y}$, HOOI \citep{de2000best,kolda2009tensor} iterates, for $k = 1, \dots, N$,
\begin{equation}
\mathcal{W}_k^{t} := \mathcal{Y} \times_{j < k} (U_j^{t+1})^\top \times_{j > k} (U_j^t)^\top, \qquad U_k^{t+1} := \text{leading } r_k \text{ left singular vectors of } (W_k^t)_{(k)},
\label{eq:hooi_step}
\end{equation}
and then sets $\mathcal{G}^{t+1} := \mathcal{Y} \times_1 (U_1^{t+1})^\top \dots \times_N (U_N^{t+1})^\top$. The data $\mathcal{Y}$ is an input, not part of the state.

\begin{proposition}[Factor state for HOOI]
\label{prop:hooi_state}
Let $z = (\{U_k\}, \mathcal{G})$ with $D(z) = \mathcal{G} \times_1 U_1 \dots \times_N U_N$. The lightweight HOOI step evaluates $(W_k^t)_{(k)} = Y_{(k)} \widehat{U}_{-k}$ by Algorithm~\ref{alg:factorwise_contraction}, where $\widehat{U}_{-k}$ is the reverse-order Kronecker product of the current factors. It then extracts the leading $r_k$ left singular vectors. By the rule of Section~\ref{sec:method_components}, this $n_k \times r_{-k}$ matrix is explicit and small, so a dense SVD is used. Assume $\sigma_{r_k} > \sigma_{r_k+1}$ for each $(W_k^t)_{(k)}$. Then the decoded iterates coincide with those of \eqref{eq:hooi_step}, i.e.\ $D(z^{t+1}) = \mathcal{Y} \times_1 P_{\mathcal{U}_1^{t+1}} \dots \times_N P_{\mathcal{U}_N^{t+1}}$ for both implementations. Moreover, the largest intermediate of the contraction chain has $\bar{n}\, r_{\pi_1} / n_{\pi_1}$ entries, whereas the literal implementation allocates $\widehat{U}_{-k} \in \mathbb{R}^{n_{-k} \times r_{-k}}$ and the product $Y_{(k)} \widehat{U}_{-k}$ through an $n_k \times n_{-k}$ operand.
\end{proposition}
\begin{proof}
By Proposition~\ref{prop:algorithm2_equivalence}, $(\mathcal{Y} \times_{j \neq k} \widehat{U}_j^\top)_{(k)} = Y_{(k)} \widehat{U}_{-k}$ for any contraction order. Hence, both implementations feed the same matrix to the spectral step. Under the gap, the leading $r_k$-dimensional left singular subspace $\mathcal{U}_k^{t+1}$ of this matrix is unique (Proposition~\ref{prop:algorithm1_exactness}(ii)). Next, suppose that a factor is replaced by another orthonormal basis $U_j Q_j$ of the same subspace. Then $(W_k^t)_{(k)}$ changes by right-multiplication with an orthogonal Kronecker matrix, which leaves its left singular subspaces and singular values unchanged. Hence, the sequence of subspaces $\{\mathcal{U}_k^t\}$ is basis-independent, and by (F2) so is $D(z^{t+1}) = \mathcal{Y} \times_j (U_j^{t+1} (U_j^{t+1})^\top) = \mathcal{Y} \times_j P_{\mathcal{U}_j^{t+1}}$. Finally, the intermediate sizes follow from Proposition~\ref{prop:algorithm2_equivalence}. After contracting mode $\pi_l$, the intermediate has $\bar{n} \prod_{l' \le l} (r_{\pi_{l'}} / n_{\pi_{l'}})$ entries. This number is largest for $l = 1$, and contracting first the mode with the smallest $r_j / n_j$ minimizes it, as Algorithm~\ref{alg:factorwise_contraction} does.
\end{proof}

\paragraph{Reuse of partial contractions (dimension tree).} We reuse partial contractions to reduce the number of contractions with the full tensor. One sweep of \eqref{eq:hooi_step} updates $U_1,\dots,U_N$ in order. The update of $U_k$ needs $\mathcal{Y}$ contracted with the new factors of the modes before $k$ and with the old factors of the modes after $k$. We split the modes into $J_1=\{1,\dots,h\}$ and $J_2=\{h+1,\dots,N\}$. All updates in $J_1$ use the old factors of $J_2$. Therefore, we compute $\mathcal{Y}_{J_2}:=\mathcal{Y}\times_{j\in J_2}(U_j^t)^\top$ once and update the modes in $J_1$ recursively from $\mathcal{Y}_{J_2}$. Then we compute $\mathcal{Y}_{J_1}:=\mathcal{Y}\times_{j\in J_1}(U_j^{t+1})^\top$ with the new factors of $J_1$ and update the modes in $J_2$ recursively from it. Every update receives exactly the matrix $(W_k^t)_{(k)}$ of \eqref{eq:hooi_step}, so Proposition~\ref{prop:hooi_state} applies unchanged. Meanwhile, the number of contractions with the full tensor per sweep falls from $N$ to $2$. This scheme applies the dimension-tree idea of \citet{kaya2016high}, developed for sparse tensors, to dense data. We use the balanced binary split; \citet{kaya2016high} discuss alternatives to it.

\subsection{Preconditioned Riemannian Gradient Descent (PRGD) and ScaledGD (E4)}
\label{sec:appendix_alg_prgd}

ScaledGD \citep{tong2022scaling}, and PRGD \citep{zhang2025preconditioned} up to the retraction, optimize directly over factors $z = (\{U_k\}, \mathcal{G})$ with $D(z) = \mathcal{G} \times_1 U_1 \dots \times_N U_N$. For $f(\mathcal{X}) = \tfrac{1}{2p} \|P_\Omega(\mathcal{X}) - y\|_F^2$, they perform updates of the form
\begin{equation}
U_k^{+} = U_k - \eta\, \big( \nabla f(\mathcal{X}) \big)_{(k)} \breve{U}_k \big( \breve{U}_k^\top \breve{U}_k \big)^{-1}, \qquad
\mathcal{G}^{+} = \mathcal{G} - \eta\, \nabla f(\mathcal{X}) \times_{j=1}^{N} \big( (U_j^\top U_j)^{-1} U_j^\top \big),
\label{eq:scaledgd}
\end{equation}
where $\mathcal{X} = D(z)$ and $\breve{U}_k := U_{-k} G_{(k)}^\top \in \mathbb{R}^{n_{-k} \times r_k}$ (so that $X_{(k)} = U_k \breve{U}_k^\top$ by (F3)); the factors need not be orthonormal. Here, the reference iteration is already defined on the factor state. However, the literal implementation materializes $\mathcal{X}$, $\breve{U}_k$ and $U_{-k}$, whereas the lightweight one does not.

\begin{proposition}[Kronecker-free ScaledGD]
\label{prop:scaledgd_state}
With $\nabla f(\mathcal{X}) = \tfrac{1}{p}(P_\Omega(\mathcal{X}) - y) = -\tfrac1p R_\Omega$ supported on $\Omega$, the quantities in \eqref{eq:scaledgd} satisfy
\begin{equation}
\big( \nabla f(\mathcal{X}) \big)_{(k)} \breve{U}_k = \big( \nabla f(\mathcal{X}) \times_{j \neq k} U_j^\top \big)_{(k)} G_{(k)}^\top, \qquad
\breve{U}_k^\top \breve{U}_k = \big( \mathcal{G} \times_{j \neq k} (U_j^\top U_j) \big)_{(k)} G_{(k)}^\top,
\end{equation}
and $\nabla f(\mathcal{X}) \times_j (U_j^\top U_j)^{-1} U_j^\top$ is a sparse contraction. Evaluating these three quantities costs $O(N|\Omega| r_{-k} + n_k r_k r_{-k})$, $O(\sum_j n_j r_j^2 + \bar{r} \sum_j r_j + r_k^2 r_{-k})$, and $O(N|\Omega|\bar{r} + \sum_j n_j r_j^2 + \sum_j r_j^3)$, respectively, and $R_\Omega$ is obtained in $O(N|\Omega|\bar{r})$. The resulting factor iterates are identical to those of the literal implementation. Hence, $\tilde{\mathcal{T}}$ intertwines with \eqref{eq:scaledgd} trivially: the state is the same object, and only the evaluation differs.
\end{proposition}
\begin{proof}
By (F3) with mode-$k$ factor $\mathrm{Id}_{n_k}$, $(\nabla f \times_{j \neq k} U_j^\top)_{(k)} = (\nabla f)_{(k)} U_{-k}$. Multiplying by $G_{(k)}^\top$ gives the first identity. For the second identity, the mixed-product rule gives $\breve{U}_k^\top \breve{U}_k = G_{(k)} U_{-k}^\top U_{-k} G_{(k)}^\top = G_{(k)} \big( \bigotimes_{j \neq k} U_j^\top U_j \big) G_{(k)}^\top$. Moreover, by (F3), $G_{(k)} C_{-k}^\top = (\mathcal{G} \times_{j \ne k} C_j)_{(k)}$ with $C_j := U_j^\top U_j = C_j^\top$. The costs follow from Proposition~\ref{prop:algorithm2_equivalence} (sparse contraction over $N-1$ or $N$ modes), the Gram computations $U_j^\top U_j$ ($2 n_j r_j^2$), the $r_j \times r_j$ inversions, and dense products of the stated sizes. In exact arithmetic, both implementations compute the same matrices \eqref{eq:scaledgd}. Hence, they produce the same factor iterates and the same $D(z^t)$. The inherited condition-number-independent rate of \cite{tong2022scaling} is then an instance of Corollary~\ref{cor:guarantee_inheritance}(i).
\end{proof}

\section{Implementation Details and Numerical Safeguards}
\label{sec:appendix_implementation}

\paragraph{Krylov extraction.} We implement Algorithm~\ref{alg:component_extraction} as Golub--Kahan--Lanczos bidiagonalization with full reorthogonalization. We add the following safeguards, none of which changes the returned triplets in exact arithmetic. (i) We take the start vector in the row space ($v_0=M^\top z$ for a seeded random $z$), so that no Lanczos steps are spent on the null space of a wide unfolding. (ii) On breakdown, i.e.\ when a Lanczos coefficient falls below $10^{-11}$ times the running norm estimate of $M$, the process restarts with a new vector orthogonal to the current basis. (iii) For thresholding, the process continues until the smallest Ritz value is certified below $\lambda$. Short verification chains (four steps from a fresh start vector) then detect singular values repeated beyond the current basis. (iv) We extract the final triplets by an SVD of the projected short-side matrix, which removes the drift of long-side vectors. The tolerances are $10^{-10}$ (default) and $10^{-14}$ (tight).

\paragraph{Copy-free contractions and Gram matrices.} We never copy an unfolding. All unfoldings are strided views of one array, and we evaluate mode products as matrix products on reshaped views in the Kolda ordering \citep{kolda2009tensor}. We accumulate the Gram matrix $Y_{(k)}Y_{(k)}^\top$ slice by slice from matrix products of views of the reshaped data (\texttt{addmm} on the GPU, \texttt{dsyrk} in the NumPy reference implementation). This accumulation needs $O(n_k^2)$ extra memory.

\paragraph{Sparse operations.} We store the observations once, as values and coordinates. The sparse contractions of Proposition~\ref{prop:algorithm2_equivalence} contract the first mode with a sparse matrix product and the remaining modes with dense products of size $|\Omega|\times r$.

\paragraph{Measured quantities.} We count $\mathrm{MEM}_{\mathrm{state}}$ from the arrays carried between iterations. We measure the data-inclusive peak with the GPU memory allocator (Appendix~\ref{sec:appendix_experimental_details}).

\section{Experimental Protocol}
\label{sec:appendix_experimental_details}
\label{app:experiments}

\paragraph{Environment.} We use one \envGPU{} GPU (\envGPUMem) on Google Colab, with Python~\envPython, PyTorch~\envTorch{} with CUDA~\envCuda{} (cuBLAS, cuSOLVER, cuSPARSE), and TensorLy~0.9.0 with its PyTorch backend. All computations use float64, and TF32 is disabled for all matrix products. Every run starts as a fresh process. Before anything is timed, the process creates the cuBLAS, cuSOLVER and cuSPARSE handles in the same way for every method. The GPU implementations are ports of our NumPy reference implementations and use the same algorithms, operation order and hyperparameters. Before the measurements, a self-test on the same GPU compares every GPU implementation with its NumPy counterpart on small problems: the results are identical up to rounding or, for randomized methods, follow the same error distribution. We generate problems and shared initializations on the host with NumPy, so that a seed defines the same problem for every implementation. This environment replaces the one described in v1, which did not correspond to the reported measurements.

\paragraph{Memory measurement.} We measure peak GPU memory with the PyTorch caching allocator. Immediately before the input is moved to the GPU, the process resets the peak statistics of the allocator and records the allocated memory. After the run, it reads the peak (\texttt{torch.cuda.max\_memory\_allocated}). We report the difference between the two readings. This difference contains the input data and every workspace and temporary allocated on the GPU, including the cuBLAS and cuSOLVER workspaces that PyTorch allocates. If a run exceeds the GPU memory, we report it as OOM and skip larger sizes of the same method. Among the HOOI-type methods on $\famNMax^3$ tensors, \famFailNames{} stop with a cuSOLVER error (\texttt{CUSOLVER\_STATUS\_INVALID\_VALUE}) in the SVD of the $1000\times10^6$ unfolding (\texttt{torch.linalg.svd}); we report these runs as failures.

\paragraph{Time measurement.} We measure wall-clock time and call \texttt{torch.cuda.synchronize()} before every clock reading. In Section~\ref{sec:hooi}, we report the end-to-end time of each method, including its initialization. In Section~\ref{sec:five}, every method starts from a shared initialization, and we take the median of the time per iteration over the $T$ iterations; we report the mean and standard error (s.e.) of this median over seeds. We report a difference smaller than two standard errors as no difference. In addition, each run pays a fixed cost of about 3\,s for process start-up and CUDA initialization; no reported time includes this cost.

\paragraph{Synthetic data.} The true tensor is $\mathcal{X}^\star=\mathcal{G}^\star\times_1U_1^\star\cdots\times_NU_N^\star$. Its core is superdiagonal with entries that decay geometrically from $1$ to $1/\kappa$, and its factors are orthonormal and Haar-distributed. We scale $\mathcal{X}^\star$ to unit root-mean-square entry. For completion, we draw $\Omega$ uniformly without replacement with $|\Omega|=p\bar n$. For denoising, we add Gaussian noise at an SNR of 20\,dB. We use seeds $0$--$4$; the seed also fixes the Krylov start vectors.

\paragraph{Real data.} We use two real data sets. HYPSO-1 \citep{justo2023open}: we take the raw L1A counts of two scenes, Bangladesh (2022-10-26) and Antarctica (2022-08-27), of size $956\times684\times120$ (lines, samples, bands), and divide them by their 99.9th percentile. This reference carries the sensor's own noise and stripes. We add Gaussian noise at an SNR of 20\,dB and report the error, the PSNR and the mean spectral angle against the reference; $r=20$, $T=20$, seeds $0$--$2$. Airport Hall \citep{li2004statistical}: we take frames 2900--3899 ($144\times176$ pixels) as distributed with the OLSTEC code (\url{https://github.com/hiroyuki-kasai/OLSTEC}) and used by \citet{kasai2016low}, and divide them by 255. The observed set $\Omega$ is a uniform fraction $p\in\{0.02,0.05,0.1\}$ of all entries, and every other entry is a test entry; $r=10$, $T=20$, seeds $0$--$2$. We run HaLRTC only at $p=0.05$.

\paragraph{Memory caps.} We cap the memory that the caching allocator may reserve with \texttt{torch.cuda.set\_per\_process\_memory\_fraction}. We set the cap before the input is moved to the GPU and lift it before the evaluation. We apply caps of 0.5, 1, 2 and 4\,GiB on seed 0, from the largest down; after an out-of-memory error, we do not run the smaller caps. However, the cap does not cover memory outside the allocator, mainly the CUDA context (\edgeCtx\,GiB in the error messages).

\paragraph{Hyperparameters.} All implementations of an estimator share the same hyperparameters. E0: we choose $\lambda$ to minimize the relative error over a logarithmic grid on an independent calibration problem (seed 999) of the same size; for each HYPSO-1 scene, we instead use an independent noise realization (seed 999) and measure the error against the reference. E1: we set $\rho_0$ so that the first threshold $\alpha_1/\rho_0$ equals half of the largest singular value of the zero-filled mode-1 unfolding, and we use $\rho_{t+1}=\min(1.05\rho_t,10^5)$ as in the public code. E2: the radius is $\tau=0.5\cdot\mathrm{RMS}(y)\cdot\sqrt{\bar n}$. E3: HOSVD initialization. E4: spectral initialization and step size $\eta=0.1$. The public code of E4 uses $0.3$, but this step size diverged for some seeds at $\kappa=100$; therefore, all implementations use $0.1$.

\subsection{Public Implementations}
\label{sec:appendix_exp_public}

\begin{table}[H]
\caption{Estimators and the components that each one uses under the fixed rule of Section~\ref{sec:method_components}. Public: the implementation that we compare against.}
\label{tab:exp1_estimators}
\centering
\scriptsize
\setlength{\tabcolsep}{4pt}
\begin{tabular}{cllll}
\toprule
\textbf{ID} & \textbf{Estimator} & \textbf{Task} & \textbf{Public implementation} & \textbf{Components} \\
\midrule
E0 & Subspace norm, ADMM \citep{zheng2015interpolating} & denoising & authors' code (port) & Alg.~2, 3 \\
E1 & HaLRTC \citep{liu2012tensor} & completion & \texttt{transdim} HaLRTC (port) & Alg.~1, 3 \\
E2 & Frank--Wolfe (FFW) \citep{guo2017efficient} & completion & authors' code (port) & Alg.~1, 3 \\
E3 & HOOI \citep{de2000best} & denoising & TensorLy \texttt{partial\_tucker} & Alg.~2, 3 \\
E4 & ScaledGD \citep{tong2022scaling} & completion & authors' code (port) & Alg.~2, 3 \\
\bottomrule
\end{tabular}
\end{table}

MATLAB was not available to us. Therefore, we ported the public MATLAB code line by line to Python (NumPy, then PyTorch for the GPU) and checked the ports on small problems against the formulas of the papers. The code for each estimator is as follows. E0: \texttt{tensor\_subspace\_norm.m} of \citet{zheng2015interpolating} (with \texttt{softth.m} and \texttt{pca.m}). E1: the HaLRTC implementation of the \texttt{transdim} repository (\url{https://github.com/xinychen/transdim}), because the original package of \citet{liu2012tensor} is no longer online. E2: \texttt{acctenfw.m} of FFWTensor (\url{https://github.com/quanmingyao/FFWTensor}), including its power-method oracle with 3 iterations per mode and 8 warm-started iterations for the selected mode. E3: TensorLy \texttt{partial\_tucker}, used as installed with the PyTorch backend. E4: \texttt{TC.m} of ScaledGD (\url{https://github.com/Titan-Tong/ScaledGD}). The dense-Gram variant of HaLRTC is our own. It squares the condition number of the unfolding, so we use it for timing only. Two details of the GPU ports differ from the MATLAB code. First, in E0, the randomized branch of \texttt{pca.m} uses a column-pivoted QR, which CUDA does not provide. We instead normalize the columns and use an unpivoted Householder QR, which spans the same subspace. This change affects the time only, because when this branch is entered, \texttt{softth.m} discards its result and recomputes it with an exact SVD. Second, in E2, we draw the random start vectors of the power method on the GPU.

\section{Complete Results of the Five-Estimator Comparison}
\label{sec:appendix_tab_bench}
\label{sec:appendix_tabulations}

\begin{table}[H]
\caption{Absolute values behind Table~\ref{tab:five}: time per iteration and data-inclusive peak GPU memory at $n=\benchNmax$ on an A100 (order 3, $r=5$, $\kappa=100$; mean $\pm$ s.e.\ over \benchSeedsMax{} seeds). Ratios are Public (or dense Gram) divided by ours; bold: at least $1.2\times$ in our favor; $^\circ$: within two s.e. $^\dagger$The fastest dense HaLRTC variant that we know; it squares the condition number, so we use it for timing only.}
\label{tab:five_full}
\centering
\scriptsize
\setlength{\tabcolsep}{2.4pt}
\resizebox{\linewidth}{!}{\begin{tabular}{llrrrrrrrr}
\toprule
 & & \multicolumn{3}{c}{Time per iteration (s)} & \multicolumn{3}{c}{Data-incl.\ peak GPU memory (MiB)} & \multicolumn{2}{c}{Public / Ours} \\
\cmidrule(lr){3-5}\cmidrule(lr){6-8}\cmidrule(lr){9-10}
Estimator & $p$ & Raw & Public & Ours & Raw & Public & Ours & time & peak \\
\midrule
E0 Subspace norm & -- & 0.111{\tiny$\pm$0.000081} & 0.114{\tiny$\pm$0.000051} & 0.0330{\tiny$\pm$0.00047} & 15{,}409{\tiny$\pm$0} & 15{,}409{\tiny$\pm$0} & 3{,}827{\tiny$\pm$0} & \textbf{3.45$\times$} & \textbf{4.03$\times$} \\
E1 HaLRTC & 0.05 & 1.08{\tiny$\pm$0.00081} & 1.08{\tiny$\pm$0.00037} & 59.8{\tiny$\pm$0.067} & 13{,}208{\tiny$\pm$0} & 15{,}806{\tiny$\pm$0} & 12{,}819{\tiny$\pm$13.8} & 0.02$\times$ & \textbf{1.23$\times$} \\
\quad vs.\ dense Gram SVD$^\dagger$ &  &  & 0.157{\tiny$\pm$0.00066} & 59.8{\tiny$\pm$0.067} &  & 10{,}806{\tiny$\pm$0} & 12{,}819{\tiny$\pm$13.8} & 0.003$\times$ & 0.84$\times$ \\
E1 HaLRTC & 0.2 & 1.05{\tiny$\pm$0.00035} & 1.08{\tiny$\pm$0.00033} & 0.894{\tiny$\pm$0.037} & 13{,}779{\tiny$\pm$0} & 15{,}948{\tiny$\pm$0} & 7{,}031{\tiny$\pm$24.6} & \textbf{1.21$\times$} & \textbf{2.27$\times$} \\
\quad vs.\ dense Gram SVD$^\dagger$ &  &  & 0.125{\tiny$\pm$0.00013} & 0.894{\tiny$\pm$0.037} &  & 11{,}377{\tiny$\pm$0} & 7{,}031{\tiny$\pm$24.6} & 0.14$\times$ & \textbf{1.62$\times$} \\
E2 Frank--Wolfe & 0.05 & 0.981{\tiny$\pm$0.0013} & 0.0120{\tiny$\pm$0.00032} & 0.172{\tiny$\pm$0.0071} & 10{,}591{\tiny$\pm$191} & 3{,}350{\tiny$\pm$0} & 1{,}620{\tiny$\pm$4.98} & 0.07$\times$ & \textbf{2.07$\times$} \\
E2 Frank--Wolfe & 0.2 & 0.996{\tiny$\pm$0.0010} & 0.0426{\tiny$\pm$0.00053} & 0.324{\tiny$\pm$0.0022} & 12{,}636{\tiny$\pm$0.00028} & 6{,}492{\tiny$\pm$0} & 3{,}773{\tiny$\pm$4.69} & 0.13$\times$ & \textbf{1.72$\times$} \\
E3 HOOI & -- & 0.0276{\tiny$\pm$0.000089} & 0.0160{\tiny$\pm$0.00026} & 0.00664{\tiny$\pm$0.000036} & 3{,}864{\tiny$\pm$0} & 1{,}918{\tiny$\pm$0} & 970{\tiny$\pm$0} & \textbf{2.41$\times$} & \textbf{1.98$\times$} \\
E4 ScaledGD & 0.05 & 0.0292{\tiny$\pm$0.0013} & 0.0231{\tiny$\pm$0.0000028} & 0.00627{\tiny$\pm$0.00028} & 5{,}197{\tiny$\pm$0} & 6{,}046{\tiny$\pm$0} & 722{\tiny$\pm$0} & \textbf{3.69$\times$} & \textbf{8.37$\times$} \\
E4 ScaledGD & 0.2 & 0.0272{\tiny$\pm$0.0000089} & 0.0257{\tiny$\pm$0.00098} & 0.0198{\tiny$\pm$0.00076} & 5{,}768{\tiny$\pm$0} & 6{,}617{\tiny$\pm$0} & 2{,}870{\tiny$\pm$0} & \textbf{1.30$\times$} & \textbf{2.31$\times$} \\
\bottomrule
\end{tabular}
}
\end{table}

\begin{table}[H]
\caption{Where the lightweight evaluation is faster or slower on the A100, and why; Figure~\ref{fig:map} summarizes the conditions. Values are ranges over the sizes and sampling rates of Section~\ref{sec:experiments}. $n$: order 3 unless stated. Memory: data-inclusive peak, or memory beyond the input ($^\ast$). $^\dagger$Against Raw. TTB: Tensor Toolbox.}
\label{tab:map}
\centering
\scriptsize
\setlength{\tabcolsep}{2pt}
\begin{tabular}{@{}lrrl@{}}
\toprule
Comparison (A100) & Time of ours & Memory of ours & Cause \\
\midrule
E0, E3, E4 vs.\ public code ($n=500$) & 1.3--3.7$\times$ faster & 2.0--8.4$\times$ less & Public code forms Kronecker products, copies, dense iterates \\
HOOI vs.\ TensorLy, TTB ($n\ge200$) & 1.05--1.33$\times$ faster & 6.8--121$\times$ less$^\ast$ & Both Kronecker-free; ours avoids the unfolding copy \\
\midrule
HOOI vs.\ HOQRI ($n\le500$; order 4) & 1.2--2.7$\times$ slower & 1.4--47$\times$ less$^\ast$ & \emph{Other iteration}: one QR step per mode; small sweeps \\
E2 Frank--Wolfe vs.\ public code & 7.6--14$\times$ slower & 1.7--2.1$\times$ less & \emph{Other oracle}: inexact power method \\
E1 HaLRTC, $p=0.05$ ($n=100$--$500$) & 32--58$\times$ slower & 1.03$\times$ less$^\dagger$ & \emph{Predictable}: near full rank, so $q\ll n$ fails (Alg.~\ref{alg:component_extraction}) \\
E0 and E4 at $n=100$ vs.\ public code & 1.1--1.4$\times$ slower & 1.3--4.6$\times$ less & \emph{Predictable}: too small, GPU kernel costs dominate \\
HOOI, $r=20$ (HYPSO-1) vs.\ TTB (eig) & 2.6$\times$ slower & 12$\times$ less$^\ast$ & \emph{Implementation}: SVD where Gram eig.\ is faster \\
\bottomrule
\end{tabular}

\end{table}

\begin{table}[H]\centering\scriptsize\setlength{\tabcolsep}{3pt}
\caption{E0 Subspace norm: all sizes and implementations (order 3, $r=5$, $\kappa=100$, $T=10$, A100; mean $\pm$ s.e.\ over seeds).}
\label{tab:bench_E0}
\begin{tabular}{rllrrrrr}
\toprule
$n$ & $p$ & impl. & seeds & time / iter.\ (s) & peak (MiB) & state (MiB) & rel.\ error \\
\midrule
100 & -- & Raw & 5 & 0.0131{\tiny$\pm$0.00024} & 128{\tiny$\pm$0.00037} & 66.9 & 0.0050 \\
100 & -- & Public & 5 & 0.0125{\tiny$\pm$0.00015} & 128{\tiny$\pm$0} & 66.9 & 0.0050 \\
100 & -- & Ours & 5 & 0.0140{\tiny$\pm$0.00033} & 31.2{\tiny$\pm$0.00037} & 7.81 & 0.0050 \\
\addlinespace[2pt]
200 & -- & Raw & 5 & 0.0160{\tiny$\pm$0.00043} & 1{,}015{\tiny$\pm$0.00059} & 511 & 0.0022 \\
200 & -- & Public & 5 & 0.0169{\tiny$\pm$0.00040} & 1{,}015{\tiny$\pm$0} & 511 & 0.0022 \\
200 & -- & Ours & 5 & 0.0134{\tiny$\pm$0.00013} & 250{\tiny$\pm$0.00059} & 61.4 & 0.0022 \\
\addlinespace[2pt]
300 & -- & Raw & 5 & 0.0320{\tiny$\pm$0.00075} & 3{,}351{\tiny$\pm$0} & 1{,}700 & 0.0014 \\
300 & -- & Public & 5 & 0.0307{\tiny$\pm$0.00042} & 3{,}351{\tiny$\pm$0} & 1{,}700 & 0.0014 \\
300 & -- & Ours & 5 & 0.0162{\tiny$\pm$0.00051} & 828{\tiny$\pm$0} & 207 & 0.0014 \\
\addlinespace[2pt]
500 & -- & Raw & 5 & 0.111{\tiny$\pm$0.000081} & 15{,}409{\tiny$\pm$0} & 7{,}773 & 0.0008 \\
500 & -- & Public & 5 & 0.114{\tiny$\pm$0.000051} & 15{,}409{\tiny$\pm$0} & 7{,}773 & 0.0008 \\
500 & -- & Ours & 5 & 0.0330{\tiny$\pm$0.00047} & 3{,}827{\tiny$\pm$0} & 955 & 0.0008 \\
\addlinespace[2pt]
\bottomrule
\end{tabular}
\end{table}

\begin{table}[H]\centering\scriptsize\setlength{\tabcolsep}{3pt}
\caption{E1 HaLRTC: all sizes and implementations (order 3, $r=5$, $\kappa=100$, $T=10$, A100; mean $\pm$ s.e.\ over seeds).}
\label{tab:bench_E1}
\begin{tabular}{rllrrrrr}
\toprule
$n$ & $p$ & impl. & seeds & time / iter.\ (s) & peak (MiB) & state (MiB) & rel.\ error \\
\midrule
100 & 0.05 & Raw & 5 & 0.0224{\tiny$\pm$0.00080} & 122{\tiny$\pm$0} & 62.0 & 0.8022 \\
100 & 0.05 & Public & 5 & 0.0226{\tiny$\pm$0.00087} & 144{\tiny$\pm$0} & 82.8 & 0.8022 \\
100 & 0.05 & Dense Gram$^\dagger$ & 5 & 0.00836{\tiny$\pm$0.00014} & 86.6{\tiny$\pm$0.050} & 62.0 & 0.8022 \\
100 & 0.05 & Ours & 5 & 0.718{\tiny$\pm$0.0039} & 122{\tiny$\pm$0.18} & 53.9 & 0.8022 \\
\addlinespace[2pt]
200 & 0.05 & Raw & 5 & 0.0772{\tiny$\pm$0.000053} & 904{\tiny$\pm$0} & 496 & 0.6937 \\
200 & 0.05 & Public & 5 & 0.0774{\tiny$\pm$0.000099} & 1{,}071{\tiny$\pm$0} & 662 & 0.6937 \\
200 & 0.05 & Dense Gram$^\dagger$ & 5 & 0.0149{\tiny$\pm$0.00029} & 702{\tiny$\pm$0} & 496 & 0.6937 \\
200 & 0.05 & Ours & 5 & 3.05{\tiny$\pm$0.058} & 892{\tiny$\pm$0.62} & 428 & 0.6937 \\
\addlinespace[2pt]
300 & 0.05 & Raw & 5 & 0.211{\tiny$\pm$0.00024} & 2{,}924{\tiny$\pm$0} & 1{,}674 & 0.6133 \\
300 & 0.05 & Public & 5 & 0.212{\tiny$\pm$0.00011} & 3{,}486{\tiny$\pm$0} & 2{,}235 & 0.6133 \\
300 & 0.05 & Dense Gram$^\dagger$ & 5 & 0.0375{\tiny$\pm$0.0011} & 2{,}333{\tiny$\pm$0} & 1{,}674 & 0.6133 \\
300 & 0.05 & Ours & 5 & 12.2{\tiny$\pm$0.096} & 2{,}885{\tiny$\pm$2.72} & 1{,}428 & 0.6133 \\
\addlinespace[2pt]
500 & 0.05 & Raw & 5 & 1.08{\tiny$\pm$0.00081} & 13{,}208{\tiny$\pm$0} & 7{,}749 & 0.4901 \\
500 & 0.05 & Public & 5 & 1.08{\tiny$\pm$0.00037} & 15{,}806{\tiny$\pm$0} & 10{,}347 & 0.4901 \\
500 & 0.05 & Dense Gram$^\dagger$ & 5 & 0.157{\tiny$\pm$0.00066} & 10{,}806{\tiny$\pm$0} & 7{,}749 & 0.4901 \\
500 & 0.05 & Ours & 5 & 59.8{\tiny$\pm$0.067} & 12{,}819{\tiny$\pm$13.8} & 6{,}417 & 0.4901 \\
\addlinespace[2pt]
100 & 0.2 & Raw & 5 & 0.0244{\tiny$\pm$0.00074} & 127{\tiny$\pm$0} & 62.0 & 0.1694 \\
100 & 0.2 & Public & 5 & 0.0235{\tiny$\pm$0.00074} & 144{\tiny$\pm$0} & 79.3 & 0.1694 \\
100 & 0.2 & Dense Gram$^\dagger$ & 5 & 0.00865{\tiny$\pm$0.00019} & 91.0{\tiny$\pm$0} & 62.0 & 0.1694 \\
100 & 0.2 & Ours & 5 & 0.408{\tiny$\pm$0.025} & 113{\tiny$\pm$0.15} & 32.2 & 0.1694 \\
\addlinespace[2pt]
200 & 0.2 & Raw & 5 & 0.0791{\tiny$\pm$0.000049} & 941{\tiny$\pm$0} & 496 & 0.1507 \\
200 & 0.2 & Public & 5 & 0.0793{\tiny$\pm$0.00014} & 1{,}080{\tiny$\pm$0} & 635 & 0.1507 \\
200 & 0.2 & Dense Gram$^\dagger$ & 5 & 0.0144{\tiny$\pm$0.00022} & 738{\tiny$\pm$0} & 496 & 0.1507 \\
200 & 0.2 & Ours & 5 & 0.487{\tiny$\pm$0.045} & 710{\tiny$\pm$1.85} & 255 & 0.1507 \\
\addlinespace[2pt]
300 & 0.2 & Raw & 5 & 0.212{\tiny$\pm$0.000095} & 3{,}051{\tiny$\pm$0} & 1{,}674 & 0.1327 \\
300 & 0.2 & Public & 5 & 0.216{\tiny$\pm$0.00015} & 3{,}519{\tiny$\pm$0} & 2{,}142 & 0.1327 \\
300 & 0.2 & Dense Gram$^\dagger$ & 5 & 0.0329{\tiny$\pm$0.00095} & 2{,}460{\tiny$\pm$0} & 1{,}674 & 0.1327 \\
300 & 0.2 & Ours & 5 & 0.499{\tiny$\pm$0.055} & 1{,}961{\tiny$\pm$5.68} & 856 & 0.1327 \\
\addlinespace[2pt]
500 & 0.2 & Raw & 5 & 1.05{\tiny$\pm$0.00035} & 13{,}779{\tiny$\pm$0} & 7{,}749 & 0.1209 \\
500 & 0.2 & Public & 5 & 1.08{\tiny$\pm$0.00033} & 15{,}948{\tiny$\pm$0} & 9{,}918 & 0.1209 \\
500 & 0.2 & Dense Gram$^\dagger$ & 5 & 0.125{\tiny$\pm$0.00013} & 11{,}377{\tiny$\pm$0} & 7{,}749 & 0.1209 \\
500 & 0.2 & Ours & 5 & 0.894{\tiny$\pm$0.037} & 7{,}031{\tiny$\pm$24.6} & 3{,}951 & 0.1209 \\
\addlinespace[2pt]
\bottomrule
\end{tabular}
\end{table}

\begin{table}[H]\centering\scriptsize\setlength{\tabcolsep}{3pt}
\caption{E2 Frank--Wolfe: all sizes and implementations (order 3, $r=5$, $\kappa=100$, $T=10$, A100; mean $\pm$ s.e.\ over seeds).}
\label{tab:bench_E2}
\begin{tabular}{rllrrrrr}
\toprule
$n$ & $p$ & impl. & seeds & time / iter.\ (s) & peak (MiB) & state (MiB) & rel.\ error \\
\midrule
100 & 0.05 & Raw & 5 & 0.0204{\tiny$\pm$0.00076} & 103{\tiny$\pm$0.00022} & 16.2 & 0.7167 \\
100 & 0.05 & Public & 5 & 0.00587{\tiny$\pm$0.000088} & 101{\tiny$\pm$0} & 1.16 & 0.7099 \\
100 & 0.05 & Ours & 5 & 0.0821{\tiny$\pm$0.0017} & 41.8{\tiny$\pm$0.12} & 1.15 & 0.7167 \\
\addlinespace[2pt]
200 & 0.05 & Raw & 5 & 0.0698{\tiny$\pm$0.00023} & 743{\tiny$\pm$0.00020} & 130 & 0.4773 \\
200 & 0.05 & Public & 5 & 0.00582{\tiny$\pm$0.00010} & 436{\tiny$\pm$0} & 6.13 & 0.4578 \\
200 & 0.05 & Ours & 5 & 0.0751{\tiny$\pm$0.0034} & 189{\tiny$\pm$2.77} & 6.12 & 0.4773 \\
\addlinespace[2pt]
300 & 0.05 & Raw & 5 & 0.191{\tiny$\pm$0.00044} & 2{,}400{\tiny$\pm$0.00059} & 438 & 0.3730 \\
300 & 0.05 & Public & 5 & 0.00591{\tiny$\pm$0.000056} & 1{,}054{\tiny$\pm$0} & 17.2 & 0.3191 \\
300 & 0.05 & Ours & 5 & 0.0768{\tiny$\pm$0.0021} & 478{\tiny$\pm$3.16} & 17.2 & 0.3730 \\
\addlinespace[2pt]
500 & 0.05 & Raw & 5 & 0.981{\tiny$\pm$0.0013} & 10{,}591{\tiny$\pm$191} & 2{,}027 & 0.1852 \\
500 & 0.05 & Public & 5 & 0.0120{\tiny$\pm$0.00032} & 3{,}350{\tiny$\pm$0} & 66.8 & 0.1851 \\
500 & 0.05 & Ours & 5 & 0.172{\tiny$\pm$0.0071} & 1{,}620{\tiny$\pm$4.98} & 66.8 & 0.1852 \\
\addlinespace[2pt]
100 & 0.2 & Raw & 5 & 0.0217{\tiny$\pm$0.00093} & 118{\tiny$\pm$0.00036} & 16.2 & 0.1744 \\
100 & 0.2 & Public & 5 & 0.00571{\tiny$\pm$0.000063} & 126{\tiny$\pm$0} & 2.31 & 0.1742 \\
100 & 0.2 & Ours & 5 & 0.0645{\tiny$\pm$0.0011} & 58.4{\tiny$\pm$0.77} & 2.30 & 0.1744 \\
\addlinespace[2pt]
200 & 0.2 & Raw & 5 & 0.0726{\tiny$\pm$0.00011} & 862{\tiny$\pm$0.00012} & 130 & 0.0696 \\
200 & 0.2 & Public & 5 & 0.00594{\tiny$\pm$0.00010} & 637{\tiny$\pm$0} & 15.3 & 0.0702 \\
200 & 0.2 & Ours & 5 & 0.0602{\tiny$\pm$0.00088} & 314{\tiny$\pm$0.35} & 15.3 & 0.0696 \\
\addlinespace[2pt]
300 & 0.2 & Raw & 5 & 0.195{\tiny$\pm$0.00028} & 2{,}810{\tiny$\pm$0.00027} & 438 & 0.0460 \\
300 & 0.2 & Public & 5 & 0.0105{\tiny$\pm$0.00029} & 1{,}747{\tiny$\pm$0} & 48.1 & 0.0438 \\
300 & 0.2 & Ours & 5 & 0.0912{\tiny$\pm$0.0014} & 931{\tiny$\pm$2.08} & 48.1 & 0.0460 \\
\addlinespace[2pt]
500 & 0.2 & Raw & 5 & 0.996{\tiny$\pm$0.0010} & 12{,}636{\tiny$\pm$0.00028} & 2{,}027 & 0.0296 \\
500 & 0.2 & Public & 5 & 0.0426{\tiny$\pm$0.00053} & 6{,}492{\tiny$\pm$0} & 210 & 0.0297 \\
500 & 0.2 & Ours & 5 & 0.324{\tiny$\pm$0.0022} & 3{,}773{\tiny$\pm$4.69} & 210 & 0.0296 \\
\addlinespace[2pt]
\bottomrule
\end{tabular}
\end{table}

\begin{table}[H]\centering\scriptsize\setlength{\tabcolsep}{3pt}
\caption{E3 HOOI: all sizes and implementations (order 3, $r=5$, $\kappa=100$, $T=10$, A100; mean $\pm$ s.e.\ over seeds).}
\label{tab:bench_E3}
\begin{tabular}{rllrrrrr}
\toprule
$n$ & $p$ & impl. & seeds & time / iter.\ (s) & peak (MiB) & state (MiB) & rel.\ error \\
\midrule
100 & -- & Raw & 5 & 0.00536{\tiny$\pm$0.00011} & 32.5{\tiny$\pm$0} & 7.64 & 0.0039 \\
100 & -- & Public & 5 & 0.0110{\tiny$\pm$0.00016} & 15.7{\tiny$\pm$0} & 0.0124 & 0.0039 \\
100 & -- & Ours & 5 & 0.00532{\tiny$\pm$0.00012} & 14.2{\tiny$\pm$0} & 0.0124 & 0.0039 \\
\addlinespace[2pt]
200 & -- & Raw & 5 & 0.00611{\tiny$\pm$0.00020} & 256{\tiny$\pm$0} & 61.1 & 0.0019 \\
200 & -- & Public & 5 & 0.0103{\tiny$\pm$0.00015} & 126{\tiny$\pm$0} & 0.0238 & 0.0019 \\
200 & -- & Ours & 5 & 0.00533{\tiny$\pm$0.00011} & 69.9{\tiny$\pm$0} & 0.0238 & 0.0019 \\
\addlinespace[2pt]
300 & -- & Raw & 5 & 0.00934{\tiny$\pm$0.00020} & 842{\tiny$\pm$0} & 206 & 0.0013 \\
300 & -- & Public & 5 & 0.0126{\tiny$\pm$0.00014} & 416{\tiny$\pm$0} & 0.0353 & 0.0013 \\
300 & -- & Ours & 5 & 0.00516{\tiny$\pm$0.000022} & 216{\tiny$\pm$0} & 0.0353 & 0.0013 \\
\addlinespace[2pt]
500 & -- & Raw & 5 & 0.0276{\tiny$\pm$0.000089} & 3{,}864{\tiny$\pm$0} & 954 & 0.0008 \\
500 & -- & Public & 5 & 0.0160{\tiny$\pm$0.00026} & 1{,}918{\tiny$\pm$0} & 0.0582 & 0.0008 \\
500 & -- & Ours & 5 & 0.00664{\tiny$\pm$0.000036} & 970{\tiny$\pm$0} & 0.0582 & 0.0008 \\
\addlinespace[2pt]
\bottomrule
\end{tabular}
\end{table}

\begin{table}[H]\centering\scriptsize\setlength{\tabcolsep}{3pt}
\caption{E4 ScaledGD: all sizes and implementations (order 3, $r=5$, $\kappa=100$, $T=10$, A100; mean $\pm$ s.e.\ over seeds).}
\label{tab:bench_E4}
\begin{tabular}{rllrrrrr}
\toprule
$n$ & $p$ & impl. & seeds & time / iter.\ (s) & peak (MiB) & state (MiB) & rel.\ error \\
\midrule
100 & 0.05 & Raw & 5 & 0.00197{\tiny$\pm$0.0000090} & 45.7{\tiny$\pm$0} & 16.2 & 0.2353 \\
100 & 0.05 & Public & 5 & 0.00255{\tiny$\pm$0.000069} & 48.7{\tiny$\pm$0} & 16.2 & 0.2353 \\
100 & 0.05 & Ours & 5 & 0.00326{\tiny$\pm$0.000034} & 10.5{\tiny$\pm$0} & 0.394 & 0.2353 \\
\addlinespace[2pt]
200 & 0.05 & Raw & 5 & 0.00218{\tiny$\pm$0.000032} & 348{\tiny$\pm$0} & 130 & 0.1181 \\
200 & 0.05 & Public & 5 & 0.00229{\tiny$\pm$0.000016} & 393{\tiny$\pm$0} & 130 & 0.1181 \\
200 & 0.05 & Ours & 5 & 0.00323{\tiny$\pm$0.000013} & 83.2{\tiny$\pm$0} & 3.08 & 0.1181 \\
\addlinespace[2pt]
300 & 0.05 & Raw & 5 & 0.00654{\tiny$\pm$0.00018} & 1{,}139{\tiny$\pm$0} & 438 & 0.1055 \\
300 & 0.05 & Public & 5 & 0.00544{\tiny$\pm$0.00015} & 1{,}307{\tiny$\pm$0} & 438 & 0.1055 \\
300 & 0.05 & Ours & 5 & 0.00337{\tiny$\pm$0.000030} & 238{\tiny$\pm$0} & 10.3 & 0.1055 \\
\addlinespace[2pt]
500 & 0.05 & Raw & 5 & 0.0292{\tiny$\pm$0.0013} & 5{,}197{\tiny$\pm$0} & 2{,}027 & 0.0689 \\
500 & 0.05 & Public & 5 & 0.0231{\tiny$\pm$0.0000028} & 6{,}046{\tiny$\pm$0} & 2{,}027 & 0.0689 \\
500 & 0.05 & Ours & 5 & 0.00627{\tiny$\pm$0.00028} & 722{\tiny$\pm$0} & 47.7 & 0.0689 \\
\addlinespace[2pt]
100 & 0.2 & Raw & 5 & 0.00205{\tiny$\pm$0.000025} & 50.8{\tiny$\pm$0} & 16.2 & 0.1101 \\
100 & 0.2 & Public & 5 & 0.00232{\tiny$\pm$0.000025} & 53.3{\tiny$\pm$0} & 16.2 & 0.1101 \\
100 & 0.2 & Ours & 5 & 0.00319{\tiny$\pm$0.000022} & 40.2{\tiny$\pm$0} & 1.54 & 0.1101 \\
\addlinespace[2pt]
200 & 0.2 & Raw & 5 & 0.00228{\tiny$\pm$0.000054} & 385{\tiny$\pm$0} & 130 & 0.0620 \\
200 & 0.2 & Public & 5 & 0.00231{\tiny$\pm$0.000029} & 430{\tiny$\pm$0} & 130 & 0.0620 \\
200 & 0.2 & Ours & 5 & 0.00347{\tiny$\pm$0.00011} & 257{\tiny$\pm$0} & 12.2 & 0.0620 \\
\addlinespace[2pt]
300 & 0.2 & Raw & 5 & 0.00632{\tiny$\pm$0.00015} & 1{,}266{\tiny$\pm$0} & 438 & 0.0385 \\
300 & 0.2 & Public & 5 & 0.00502{\tiny$\pm$0.000052} & 1{,}434{\tiny$\pm$0} & 438 & 0.0385 \\
300 & 0.2 & Ours & 5 & 0.00513{\tiny$\pm$0.00016} & 629{\tiny$\pm$0} & 41.2 & 0.0385 \\
\addlinespace[2pt]
500 & 0.2 & Raw & 5 & 0.0272{\tiny$\pm$0.0000089} & 5{,}768{\tiny$\pm$0} & 2{,}027 & 0.0332 \\
500 & 0.2 & Public & 5 & 0.0257{\tiny$\pm$0.00098} & 6{,}617{\tiny$\pm$0} & 2{,}027 & 0.0332 \\
500 & 0.2 & Ours & 5 & 0.0198{\tiny$\pm$0.00076} & 2{,}870{\tiny$\pm$0} & 191 & 0.0332 \\
\addlinespace[2pt]
\bottomrule
\end{tabular}
\end{table}

\subsection{Real Data}
\label{sec:appendix_real}

\paragraph{Edge-sized devices.} Table~\ref{tab:edge} gives the numbers behind Figure~\ref{fig:edge}, and Table~\ref{tab:edge_budget} gives the runs under a cap on the PyTorch allocator. At $p=0.02$ and $p=0.1$, ours needs $\edgeOtherPubMin$--$\edgeOtherPubMax\times$ less peak memory than the public implementations of ScaledGD and Frank--Wolfe, and $\edgeOtherRawMin$--$\edgeOtherRawMax\times$ less than Raw. At rank 20, the denoised cubes are smooth in space (Figure~\ref{fig:edge}a). The completed videos recover the static background but not the people who move (Figure~\ref{fig:edge}b). Figure~\ref{fig:edge}a shows the bands at 641, 550 and 469\,nm with one stretch for all panels, in a region of $192\times192$ pixels of the Bangladesh scene. In Figure~\ref{fig:edge}b, $\Omega$ is drawn uniformly over the whole video. The A100 runs record metrics, not estimates. We therefore recomputed the images on the CPU with the same code, problem generator, seed and hyperparameters as the A100 runs. The recomputed estimates reproduce the A100 estimates to a relative $\edgeImgDiff$ at 512 fixed positions.

\begin{table}[H]
\caption{Real data on an A100 (float64, $T=20$; median of 3 seeds): data-inclusive peak GPU memory (bold: smallest), its reduction by ours, time per iteration, and the PSNR of ours. This PSNR equals that of Raw; we measure it on the unobserved entries for Airport Hall and over the whole cube for HYPSO-1. $^\dagger$The public code uses an inexact oracle and computes other iterates; its PSNR is \edgeFWPubPSNR\,dB. $^\ddagger$Ours is not lighter and is much slower (Section~\ref{sec:five_limits}); a dense-Gram variant needs \edgeDenseGramPeak\,GiB and \edgeDenseGramMs\,ms. HYPSO-1 rows show the Bangladesh scene. The Antarctica scene has the same peaks (within \edgeAntPeakDiff\,MiB) and a PSNR of \edgeAntPSNRThree{} (HOOI) and \edgeAntPSNRZero\,dB (subspace norm).}
\label{tab:edge}
\centering
\scriptsize
\setlength{\tabcolsep}{4pt}
\begin{tabular}{@{}llrrrrrrrr@{}}
\toprule
 & & \multicolumn{3}{c}{Peak (GiB)} & \multicolumn{2}{c}{Reduction} & \multicolumn{2}{c}{Time (ms)} & PSNR \\
\cmidrule(lr){3-5}\cmidrule(lr){6-7}\cmidrule(lr){8-9}
Data & Estimator & Raw & Public & Ours & vs.\ Raw & vs.\ Public & Public & Ours & (dB) \\
\midrule
Airport Hall & ScaledGD & 1.28 & 1.21 & \textbf{0.39} & 3.33$\times$ & 3.14$\times$ & 5.9 & 3.7 & 19.40 \\
($p=0.05$) & Frank--Wolfe$^\dagger$ & 2.13 & 1.25 & \textbf{0.81} & 2.64$\times$ & 1.55$\times$ & 5.9 & 188 & 21.50 \\
 & HaLRTC$^\ddagger$ & \textbf{2.86} & 3.38 & 2.87 & 1.00$\times$ & 1.18$\times$ & 328 & $1.7\times10^{5}$ & 16.09 \\
\midrule
HYPSO-1 & HOOI & 3.71 & 1.19 & \textbf{0.62} & 5.98$\times$ & 1.92$\times$ & 90.2 & 81.0 & 33.89 \\
 & Subspace norm & 11.90 & 11.90 & \textbf{2.46} & 4.83$\times$ & 4.83$\times$ & 302 & 191 & 35.60 \\
\bottomrule
\end{tabular}

\end{table}

\begin{table}[H]
\caption{Runs under a cap on the PyTorch allocator (\texttt{torch.cuda.set\_per\_process\_memory\_fraction}; seed 0, A100). \checkmark{}: completed $T=20$ iterations; OOM: out of memory; --: not run, because the same run was out of memory under a larger cap. Peak as in Table~\ref{tab:edge}. The cap does not cover the \edgeCtx\,GiB held outside the allocator. Although its peak is below the cap, Raw HOOI runs out of memory under the 4\,GiB cap because of fragmentation.}
\label{tab:edge_budget}
\centering
\scriptsize
\begin{tabular}{@{}lllrcccc@{}}
\toprule
 & & & Peak & \multicolumn{4}{c}{Allocator cap (GiB)} \\
\cmidrule(lr){5-8}
Data & Estimator & Code & (GiB) & 0.5 & 1 & 2 & 4 \\
\midrule
Airport Hall & ScaledGD & Raw & 1.28 & -- & OOM & \checkmark & \checkmark \\
 &  & Public & 1.21 & -- & OOM & \checkmark & \checkmark \\
 &  & Ours & 0.39 & \checkmark & \checkmark & \checkmark & \checkmark \\
\addlinespace[2pt]
 & Frank--Wolfe & Raw & 2.13 & -- & -- & OOM & \checkmark \\
 &  & Public & 1.25 & -- & OOM & \checkmark & \checkmark \\
 &  & Ours & 0.81 & OOM & \checkmark & \checkmark & \checkmark \\
\midrule
HYPSO-1 & HOOI & Raw & 3.71 & -- & -- & -- & OOM \\
 &  & Public & 1.19 & -- & OOM & \checkmark & \checkmark \\
 &  & Ours & 0.62 & OOM & \checkmark & \checkmark & \checkmark \\
\addlinespace[2pt]
 & Subspace norm & Raw & 11.90 & -- & -- & -- & OOM \\
 &  & Public & 11.90 & -- & -- & -- & OOM \\
 &  & Ours & 2.46 & -- & -- & OOM & \checkmark \\
\bottomrule
\end{tabular}

\end{table}

\section{HOOI-Type Methods: Parameters and Complete Results}
\label{sec:appendix_hooi_family}

\paragraph{Methods and parameters.} All parameters are the defaults of the authors' code, and all methods except TensorLy are ports. Naive HOOI forms the explicit Kronecker product of the factors and computes full SVDs. TensorLy \texttt{partial\_tucker} runs with \texttt{init="svd"} and \texttt{tol=0}. Tensor Toolbox \texttt{tucker\_als} initializes with the leading eigenvectors of the Gram matrices of modes $2,\dots,N$. Then, for each mode, it applies a chain of mode products and takes the leading eigenvectors of the Gram matrix of the contracted unfolding. It computes these eigenvectors either with a dense symmetric eigensolver (eig) or iteratively (eigs). For eigs, the MATLAB code uses ARPACK (\citealp{lehoucq1998arpack}); since ARPACK has no GPU version, we use a Lanczos method with full reorthogonalization on the GPU. HOQRI \citep{sun2022hoqri} takes one QR step per mode instead of an SVD. HOSVD and ST-HOSVD \citep{vannieuwenhoven2012new} compute truncated SVDs of the unfoldings; ST-HOSVD truncates sequentially. RandHOSVD and RandST-HOSVD \citep{minster2020randomized} use a randomized range finder with oversampling $10$ and one power iteration. TUCKER-TS and TUCKER-TTMTS \citep{malik2018low} use TensorSketch with sketch-size factor $K=10$, tolerance $10^{-3}$, at most 50 iterations and random initialization, as in the authors' dense demo. The one- and two-pass sketches \citep{sun2020lowrank} use Gaussian Khatri--Rao arm sketches of size $k=4r$ and a core sketch of size $s=2k+1$. D-Tucker \citep{jang2020dtucker} computes a randomized SVD of every slice with target rank $4r$ and $10r$ samples, and then runs alternating least squares (ALS) on the compressed slices with tolerance $10^{-4}$ and at most 50 iterations. On the GPU, we solve the least-squares problems of TUCKER-TS and D-Tucker through a thin SVD, because CUDA's least-squares solver handles only full-rank problems; the thin SVD gives the same minimum-norm solution as LAPACK \texttt{gelsd}. We also draw the large random test matrices of RandHOSVD and RandST-HOSVD and the CountSketch of D-Tucker on the GPU. Over 15 seeds, the median errors of the randomized methods are within $0.86$--$1.18\times$ those of the NumPy implementations.

\begin{table}[H]
\caption{Existing exact HOOI implementations relative to ours on an A100 GPU (mean over \famSeeds{} seeds). We divide the end-to-end time and the GPU memory beyond the input tensor of each implementation by those of ours, so values above $1\times$ favor ours; underlined values are below $1\times$. The rows ``Ours (absolute)'' give our absolute values. fail$^\S$: cuSOLVER returns an error for the SVD of the $1000\times10^6$ unfolding. Standard errors and all other methods: Appendix~\ref{sec:appendix_hooi_family}.}
\label{tab:hooi_ratio}
\centering
\scriptsize
\setlength{\tabcolsep}{2.0pt}
\begin{tabular}{lrrrrrrrrrr}
\toprule
 & \multicolumn{6}{c}{order 3, $r=5$} & \multicolumn{4}{c}{order 4, $r=3$} \\
\cmidrule(lr){2-7}\cmidrule(lr){8-11}
 & $n=100$ & $n=200$ & $n=300$ & $n=500$ & $n=800$ & $n=1000$ & $n=20$ & $n=40$ & $n=60$ & $n=80$ \\
\midrule
\multicolumn{11}{l}{\emph{End-to-end time, relative to ours}} \\
\quad Naive HOOI & \underline{0.80$\times$} & 1.29$\times$ & 2.29$\times$ & 7.22$\times$ & 15$\times$ & fail$^\S$ & \underline{0.53$\times$} & \underline{0.62$\times$} & 1.05$\times$ & 1.93$\times$ \\
\quad TensorLy & 2.04$\times$ & 2.58$\times$ & 3.36$\times$ & 7.25$\times$ & 13$\times$ & fail$^\S$ & 2.05$\times$ & 1.64$\times$ & 1.76$\times$ & 2.16$\times$ \\
\quad Tensor Toolbox (eig) & 1.04$\times$ & 1.18$\times$ & 1.33$\times$ & 1.68$\times$ & 1.66$\times$ & 1.50$\times$ & \underline{0.91$\times$} & \underline{0.71$\times$} & \underline{0.59$\times$} & \underline{0.55$\times$} \\
\quad Tensor Toolbox (eigs) & 1.85$\times$ & 1.69$\times$ & 1.44$\times$ & 1.20$\times$ & 1.14$\times$ & 1.05$\times$ & 2.47$\times$ & 1.40$\times$ & 1.04$\times$ & \underline{0.78$\times$} \\
\quad HOQRI & \underline{0.74$\times$} & \underline{0.67$\times$} & \underline{0.65$\times$} & \underline{0.85$\times$} & 1.12$\times$ & 1.15$\times$ & \underline{0.83$\times$} & \underline{0.55$\times$} & \underline{0.42$\times$} & \underline{0.37$\times$} \\
\quad \textbf{Ours (absolute)} & \textbf{0.152\,s} & \textbf{0.135\,s} & \textbf{0.146\,s} & \textbf{0.181\,s} & \textbf{0.383\,s} & \textbf{0.678\,s} & \textbf{0.137\,s} & \textbf{0.201\,s} & \textbf{0.243\,s} & \textbf{0.309\,s} \\
\midrule
\multicolumn{11}{l}{\emph{Memory beyond the input tensor, relative to ours}} \\
\quad Naive HOOI & 8.80$\times$ & 41$\times$ & 107$\times$ & 255$\times$ & 502$\times$ & fail$^\S$ & 35$\times$ & 34$\times$ & 143$\times$ & 205$\times$ \\
\quad TensorLy & 8.86$\times$ & 41$\times$ & 107$\times$ & 256$\times$ & 503$\times$ & fail$^\S$ & 35$\times$ & 34$\times$ & 143$\times$ & 206$\times$ \\
\quad Tensor Toolbox (eig) & 1.37$\times$ & 7.06$\times$ & 19$\times$ & 47$\times$ & 95$\times$ & 122$\times$ & 2.89$\times$ & 3.18$\times$ & 16$\times$ & 25$\times$ \\
\quad Tensor Toolbox (eigs) & 1.17$\times$ & 6.78$\times$ & 18$\times$ & 47$\times$ & 95$\times$ & 121$\times$ & 1.92$\times$ & 3.07$\times$ & 15$\times$ & 25$\times$ \\
\quad HOQRI & 1.37$\times$ & 7.06$\times$ & 19$\times$ & 47$\times$ & 95$\times$ & 122$\times$ & 9.34$\times$ & 3.18$\times$ & 16$\times$ & 25$\times$ \\
\quad \textbf{Ours (absolute)} & \textbf{6.70\,MiB} & \textbf{9.42\,MiB} & \textbf{11.3\,MiB} & \textbf{20.6\,MiB} & \textbf{41.4\,MiB} & \textbf{63.2\,MiB} & \textbf{0.644\,MiB} & \textbf{6.69\,MiB} & \textbf{6.43\,MiB} & \textbf{12.5\,MiB} \\
\midrule
Input tensor $\mathcal{Y}$ & 7.63\,MiB & 61.0\,MiB & 206\,MiB & 954\,MiB & 3{,}906\,MiB & 7{,}629\,MiB & 1.22\,MiB & 19.5\,MiB & 98.9\,MiB & 312\,MiB \\
\bottomrule
\end{tabular}

\end{table}

\paragraph{What makes the difference.} Table~\ref{tab:hooi_ablation} removes one ingredient at a time to show what makes lightweight HOOI fast and light. First, reusing partial contractions shortens a sweep by $\famTreeGain\times$ on $\famNThree^3$ tensors and by $\famTreeGainFour\times$ at order 4. Second, the copy-free Gram initialization is what keeps the peak close to the size of the input. In contrast, the Krylov initialization keeps long Lanczos vectors of length $n^{N-1}$ (peak \famKryInitPeak\,MiB, $\famKryInitTime\times$ slower overall), and forming the unfolding doubles the peak (\famCopyInitPeak\,MiB). Finally, using Krylov extraction also on the small explicit matrices is $\famAllKryTime\times$ slower, as the fixed rule predicts.

\begin{table}[H]
\caption{What makes lightweight HOOI fast and light (mean $\pm$ s.e.\ over \famSeeds{} seeds). ``no reuse'': the $N$ contractions per sweep are computed independently. ``Krylov HOSVD init.'': the initialization applies Algorithm~\ref{alg:component_extraction} to the unfolding operator instead of using the copy-free Gram matrix. ``Krylov everywhere'': Krylov extraction is also used for the small contracted matrices. ``Gram via unfolding copy'': the initialization forms each unfolding explicitly.}
\label{tab:hooi_ablation}
\centering
\scriptsize
\setlength{\tabcolsep}{2.1pt}
\begin{tabular}{lrrrrrr}
\toprule
 & \multicolumn{3}{c}{order 3, $n=800$} & \multicolumn{3}{c}{order 4, $n=80$} \\
\cmidrule(lr){2-4}\cmidrule(lr){5-7}
Configuration & total (s) & per iter.\ (s) & peak (MiB) & total (s) & per iter.\ (s) & peak (MiB) \\
\midrule
\textbf{Ours} & 0.383{\tiny$\pm$0.0033} & 0.0120{\tiny$\pm$0.000025} & 3{,}948{\tiny$\pm$0} & 0.309{\tiny$\pm$0.00047} & 0.00683{\tiny$\pm$0.000052} & 325{\tiny$\pm$0} \\
Ours, no reuse & 0.409{\tiny$\pm$0.0018} & 0.0151{\tiny$\pm$0.000027} & 3{,}947{\tiny$\pm$0} & 0.316{\tiny$\pm$0.0017} & 0.00733{\tiny$\pm$0.00020} & 325{\tiny$\pm$0} \\
Ours, Krylov HOSVD init. & 0.513{\tiny$\pm$0.0025} & 0.0120{\tiny$\pm$0.000019} & 5{,}614{\tiny$\pm$2.62} & 0.251{\tiny$\pm$0.0033} & 0.00672{\tiny$\pm$0.00013} & 1{,}670{\tiny$\pm$0.00078} \\
Ours, Krylov everywhere, no reuse & 0.773{\tiny$\pm$0.0048} & 0.0377{\tiny$\pm$0.00022} & 5{,}614{\tiny$\pm$2.62} & 0.545{\tiny$\pm$0.0048} & 0.0363{\tiny$\pm$0.00024} & 1{,}670{\tiny$\pm$0.00078} \\
Ours, no reuse, Gram via unfolding copy & 0.478{\tiny$\pm$0.0033} & 0.0151{\tiny$\pm$0.000017} & 7{,}848{\tiny$\pm$0} & 0.170{\tiny$\pm$0.0021} & 0.00752{\tiny$\pm$0.00015} & 627{\tiny$\pm$0} \\
\bottomrule
\end{tabular}

\end{table}

\begin{table}[H]
\caption{Exact HOOI implementations (and HOQRI) on an A100: absolute end-to-end time and data-inclusive peak GPU memory (mean $\pm$ s.e.\ over \famSeeds{} seeds; bold: best per column). fail$^\S$: cuSOLVER error in the SVD of the $1000\times10^6$ unfolding. Table~\ref{tab:hooi_ratio} gives the same comparison as ratios.}
\label{tab:hooi_exact}
\centering
\scriptsize
\setlength{\tabcolsep}{3.2pt}
\begin{tabular}{lrrrrrr}
\toprule
 & \multicolumn{6}{c}{order 3, $r=5$} \\
\cmidrule(lr){2-7}
Method & $n=100$ & $n=200$ & $n=300$ & $n=500$ & $n=800$ & $n=1000$ \\
\midrule
\multicolumn{7}{l}{\emph{End-to-end time (s), including each method's own initialization}} \\
Naive HOOI & 0.121{\tiny$\pm$0.0060} & 0.174{\tiny$\pm$0.0042} & 0.335{\tiny$\pm$0.0058} & 1.31{\tiny$\pm$0.0020} & 5.79{\tiny$\pm$0.0038} & fail$^\S$ \\
TensorLy & 0.309{\tiny$\pm$0.0030} & 0.349{\tiny$\pm$0.0024} & 0.492{\tiny$\pm$0.0049} & 1.31{\tiny$\pm$0.0048} & 5.02{\tiny$\pm$0.0032} & fail$^\S$ \\
Tensor Toolbox (eig) & 0.158{\tiny$\pm$0.0020} & 0.159{\tiny$\pm$0.0021} & 0.194{\tiny$\pm$0.0060} & 0.304{\tiny$\pm$0.010} & 0.636{\tiny$\pm$0.0027} & 1.02{\tiny$\pm$0.0063} \\
Tensor Toolbox (eigs) & 0.280{\tiny$\pm$0.0011} & 0.229{\tiny$\pm$0.0026} & 0.210{\tiny$\pm$0.00092} & 0.217{\tiny$\pm$0.0023} & 0.438{\tiny$\pm$0.0042} & 0.709{\tiny$\pm$0.0067} \\
HOQRI & \textbf{0.112{\tiny$\pm$0.0032}} & \textbf{0.0912{\tiny$\pm$0.0026}} & \textbf{0.0949{\tiny$\pm$0.00069}} & \textbf{0.153{\tiny$\pm$0.00079}} & 0.430{\tiny$\pm$0.0027} & 0.782{\tiny$\pm$0.0034} \\
\textbf{Ours} & 0.152{\tiny$\pm$0.0014} & 0.135{\tiny$\pm$0.0023} & 0.146{\tiny$\pm$0.0020} & 0.181{\tiny$\pm$0.0030} & \textbf{0.383{\tiny$\pm$0.0033}} & \textbf{0.678{\tiny$\pm$0.0026}} \\
\midrule
\multicolumn{7}{l}{\emph{Data-inclusive peak GPU memory (MiB)}} \\
Naive HOOI & 66.6{\tiny$\pm$0} & 450{\tiny$\pm$0} & 1{,}415{\tiny$\pm$0} & 6{,}218{\tiny$\pm$0} & 24{,}702{\tiny$\pm$0} & fail$^\S$ \\
TensorLy & 67.0{\tiny$\pm$0} & 451{\tiny$\pm$0} & 1{,}418{\tiny$\pm$0} & 6{,}228{\tiny$\pm$0} & 24{,}726{\tiny$\pm$0} & fail$^\S$ \\
Tensor Toolbox (eig) & 16.8{\tiny$\pm$0} & 128{\tiny$\pm$0} & 418{\tiny$\pm$0} & 1{,}923{\tiny$\pm$0} & 7{,}848{\tiny$\pm$0} & 15{,}315{\tiny$\pm$0} \\
Tensor Toolbox (eigs) & 15.5{\tiny$\pm$0} & 125{\tiny$\pm$0} & 414{\tiny$\pm$0} & 1{,}914{\tiny$\pm$0} & 7{,}827{\tiny$\pm$0} & 15{,}283{\tiny$\pm$0} \\
HOQRI & 16.8{\tiny$\pm$0} & 128{\tiny$\pm$0} & 418{\tiny$\pm$0} & 1{,}923{\tiny$\pm$0} & 7{,}848{\tiny$\pm$0} & 15{,}315{\tiny$\pm$0} \\
\textbf{Ours} & \textbf{14.3{\tiny$\pm$0}} & \textbf{70.5{\tiny$\pm$0}} & \textbf{217{\tiny$\pm$0}} & \textbf{974{\tiny$\pm$0}} & \textbf{3{,}948{\tiny$\pm$0}} & \textbf{7{,}693{\tiny$\pm$0}} \\
\midrule
\multicolumn{7}{l}{\emph{Input tensor $\mathcal{Y}$ (MiB)}} \\
& 7.63 & 61.0 & 206 & 954 & 3{,}906 & 7{,}629 \\
\bottomrule
\end{tabular}

\medskip

\begin{tabular}{lrrrr}
\toprule
 & \multicolumn{4}{c}{order 4, $r=3$} \\
\cmidrule(lr){2-5}
Method & $n=20$ & $n=40$ & $n=60$ & $n=80$ \\
\midrule
\multicolumn{5}{l}{\emph{End-to-end time (s), including each method's own initialization}} \\
Naive HOOI & \textbf{0.0727{\tiny$\pm$0.0051}} & 0.124{\tiny$\pm$0.0015} & 0.256{\tiny$\pm$0.0063} & 0.595{\tiny$\pm$0.0056} \\
TensorLy & 0.280{\tiny$\pm$0.0056} & 0.330{\tiny$\pm$0.0023} & 0.426{\tiny$\pm$0.0033} & 0.667{\tiny$\pm$0.0023} \\
Tensor Toolbox (eig) & 0.124{\tiny$\pm$0.00071} & 0.144{\tiny$\pm$0.0019} & 0.143{\tiny$\pm$0.0017} & 0.171{\tiny$\pm$0.0038} \\
Tensor Toolbox (eigs) & 0.337{\tiny$\pm$0.0019} & 0.282{\tiny$\pm$0.0017} & 0.253{\tiny$\pm$0.0023} & 0.240{\tiny$\pm$0.0022} \\
HOQRI & 0.113{\tiny$\pm$0.0018} & \textbf{0.110{\tiny$\pm$0.00099}} & \textbf{0.101{\tiny$\pm$0.00034}} & \textbf{0.113{\tiny$\pm$0.00092}} \\
\textbf{Ours} & 0.137{\tiny$\pm$0.0015} & 0.201{\tiny$\pm$0.0021} & 0.243{\tiny$\pm$0.0019} & 0.309{\tiny$\pm$0.00047} \\
\midrule
\multicolumn{5}{l}{\emph{Data-inclusive peak GPU memory (MiB)}} \\
Naive HOOI & 23.5{\tiny$\pm$0} & 245{\tiny$\pm$0} & 1{,}016{\tiny$\pm$0} & 2{,}876{\tiny$\pm$0} \\
TensorLy & 23.7{\tiny$\pm$0} & 246{\tiny$\pm$0} & 1{,}021{\tiny$\pm$0} & 2{,}888{\tiny$\pm$0} \\
Tensor Toolbox (eig) & 3.08{\tiny$\pm$0} & 40.8{\tiny$\pm$0} & 199{\tiny$\pm$0} & 627{\tiny$\pm$0} \\
Tensor Toolbox (eigs) & 2.45{\tiny$\pm$0} & 40.0{\tiny$\pm$0} & 198{\tiny$\pm$0} & 625{\tiny$\pm$0} \\
HOQRI & 7.23{\tiny$\pm$0} & 40.8{\tiny$\pm$0} & 199{\tiny$\pm$0} & 627{\tiny$\pm$0} \\
\textbf{Ours} & \textbf{1.86{\tiny$\pm$0}} & \textbf{26.2{\tiny$\pm$0}} & \textbf{105{\tiny$\pm$0}} & \textbf{325{\tiny$\pm$0}} \\
\midrule
\multicolumn{5}{l}{\emph{Input tensor $\mathcal{Y}$ (MiB)}} \\
& 1.22 & 19.5 & 98.9 & 312 \\
\bottomrule
\end{tabular}

\medskip

\end{table}

\begin{table}[H]\centering\scriptsize\setlength{\tabcolsep}{2.3pt}
\caption{All HOOI-type methods, order 3, $n=100$, $r=5$, $\kappa=100$, SNR 20\,dB, 5 seeds (mean $\pm$ s.e.). Input 7.63\,MiB.}
\label{tab:hooi_all_3_100}
\begin{tabular}{llrrrrr}
\toprule
Method & type & total (s) & per iter.\ (s) & peak (MiB) & rel.\ error to $\mathcal{X}^\star$ & vs.\ Ours \\
\midrule
Naive HOOI & exact & 0.121{\tiny$\pm$0.0060} & 0.00523{\tiny$\pm$0.00011} & 66.6{\tiny$\pm$0} & $3.9\times10^{-3}$ & 0.80$\times$ \\
TensorLy & exact & 0.309{\tiny$\pm$0.0030} & -- & 67.0{\tiny$\pm$0} & $3.9\times10^{-3}$ & 2.04$\times$ \\
Tensor Toolbox (eig) & exact & 0.158{\tiny$\pm$0.0020} & 0.00678{\tiny$\pm$0.00019} & 16.8{\tiny$\pm$0} & $3.9\times10^{-3}$ & 1.04$\times$ \\
Tensor Toolbox (eigs) & exact & 0.280{\tiny$\pm$0.0011} & 0.0198{\tiny$\pm$0.000062} & 15.5{\tiny$\pm$0} & $3.9\times10^{-3}$ & 1.85$\times$ \\
HOQRI & same fixed pt. & 0.112{\tiny$\pm$0.0032} & 0.00130{\tiny$\pm$0.000045} & 16.8{\tiny$\pm$0} & $3.9\times10^{-3}$ & 0.74$\times$ \\
\textbf{Ours} & exact & 0.152{\tiny$\pm$0.0014} & 0.00513{\tiny$\pm$0.00014} & 14.3{\tiny$\pm$0} & $3.9\times10^{-3}$ & 1.00$\times$ \\
Ours, no reuse & exact & 0.156{\tiny$\pm$0.0017} & 0.00553{\tiny$\pm$0.00013} & 13.9{\tiny$\pm$0} & $3.9\times10^{-3}$ & 1.03$\times$ \\
Ours, Krylov HOSVD init. & exact & 0.181{\tiny$\pm$0.0018} & 0.00524{\tiny$\pm$0.00013} & 37.0{\tiny$\pm$0.048} & $3.9\times10^{-3}$ & 1.20$\times$ \\
Ours, Krylov everywhere, no reuse & exact & 0.474{\tiny$\pm$0.0031} & 0.0336{\tiny$\pm$0.00033} & 37.0{\tiny$\pm$0.048} & $3.9\times10^{-3}$ & 3.13$\times$ \\
Ours, no reuse, Gram via unfolding copy & exact & 0.151{\tiny$\pm$0.0016} & 0.00537{\tiny$\pm$0.000099} & 16.8{\tiny$\pm$0} & $3.9\times10^{-3}$ & 1.00$\times$ \\
HOSVD & approx. & 0.0492{\tiny$\pm$0.0010} & -- & 66.7{\tiny$\pm$0} & $4.3\times10^{-3}$ & 0.32$\times$ \\
ST-HOSVD & approx. & 0.0495{\tiny$\pm$0.00044} & -- & 66.6{\tiny$\pm$0} & $4.1\times10^{-3}$ & 0.33$\times$ \\
RandHOSVD & approx. & 0.0438{\tiny$\pm$0.0049} & -- & 44.2{\tiny$\pm$0} & $1.1\times10^{-2}$ & 0.29$\times$ \\
RandST-HOSVD & approx. & 0.0333{\tiny$\pm$0.00043} & -- & 44.2{\tiny$\pm$0} & $8.8\times10^{-3}$ & 0.22$\times$ \\
TUCKER-TS & approx. & 0.598{\tiny$\pm$0.0055} & 0.00782{\tiny$\pm$0.00012} & 17.2{\tiny$\pm$0} & $4.2\times10^{-2}$ & 3.95$\times$ \\
TUCKER-TTMTS & approx. & 0.338{\tiny$\pm$0.0029} & 0.00349{\tiny$\pm$0.000048} & 16.2{\tiny$\pm$0} & $3.9\times10^{-1}$ & 2.23$\times$ \\
Two-pass sketch & approx. & 0.0374{\tiny$\pm$0.0049} & -- & 16.9{\tiny$\pm$0} & $8.0\times10^{-2}$ & 0.25$\times$ \\
One-pass sketch & approx. & 0.0863{\tiny$\pm$0.00025} & -- & 16.9{\tiny$\pm$0} & $1.1\times10^{-1}$ & 0.57$\times$ \\
D-Tucker & approx. & 0.521{\tiny$\pm$0.0024} & 0.0350{\tiny$\pm$0.00049} & 16.9{\tiny$\pm$0} & $2.4\times10^{-2}$ & 3.44$\times$ \\
\bottomrule
\end{tabular}
\end{table}

\begin{table}[H]\centering\scriptsize\setlength{\tabcolsep}{2.3pt}
\caption{All HOOI-type methods, order 3, $n=200$, $r=5$, $\kappa=100$, SNR 20\,dB, 5 seeds (mean $\pm$ s.e.). Input 61.0\,MiB.}
\label{tab:hooi_all_3_200}
\begin{tabular}{llrrrrr}
\toprule
Method & type & total (s) & per iter.\ (s) & peak (MiB) & rel.\ error to $\mathcal{X}^\star$ & vs.\ Ours \\
\midrule
Naive HOOI & exact & 0.174{\tiny$\pm$0.0042} & 0.00583{\tiny$\pm$0.00022} & 450{\tiny$\pm$0} & $1.9\times10^{-3}$ & 1.29$\times$ \\
TensorLy & exact & 0.349{\tiny$\pm$0.0024} & -- & 451{\tiny$\pm$0} & $1.9\times10^{-3}$ & 2.58$\times$ \\
Tensor Toolbox (eig) & exact & 0.159{\tiny$\pm$0.0021} & 0.00845{\tiny$\pm$0.00020} & 128{\tiny$\pm$0} & $1.9\times10^{-3}$ & 1.18$\times$ \\
Tensor Toolbox (eigs) & exact & 0.229{\tiny$\pm$0.0026} & 0.0169{\tiny$\pm$0.00023} & 125{\tiny$\pm$0} & $1.9\times10^{-3}$ & 1.69$\times$ \\
HOQRI & same fixed pt. & 0.0912{\tiny$\pm$0.0026} & 0.00125{\tiny$\pm$0.000023} & 128{\tiny$\pm$0} & $1.9\times10^{-3}$ & 0.67$\times$ \\
\textbf{Ours} & exact & 0.135{\tiny$\pm$0.0023} & 0.00532{\tiny$\pm$0.00019} & 70.5{\tiny$\pm$0} & $1.9\times10^{-3}$ & 1.00$\times$ \\
Ours, no reuse & exact & 0.132{\tiny$\pm$0.0013} & 0.00508{\tiny$\pm$0.000096} & 68.9{\tiny$\pm$0} & $1.9\times10^{-3}$ & 0.98$\times$ \\
Ours, Krylov HOSVD init. & exact & 0.169{\tiny$\pm$0.0015} & 0.00510{\tiny$\pm$0.00016} & 172{\tiny$\pm$0.12} & $1.9\times10^{-3}$ & 1.25$\times$ \\
Ours, Krylov everywhere, no reuse & exact & 0.434{\tiny$\pm$0.0033} & 0.0315{\tiny$\pm$0.00020} & 172{\tiny$\pm$0.12} & $1.9\times10^{-3}$ & 3.21$\times$ \\
Ours, no reuse, Gram via unfolding copy & exact & 0.132{\tiny$\pm$0.0019} & 0.00535{\tiny$\pm$0.00017} & 128{\tiny$\pm$0} & $1.9\times10^{-3}$ & 0.97$\times$ \\
HOSVD & approx. & 0.107{\tiny$\pm$0.0024} & -- & 451{\tiny$\pm$0} & $2.0\times10^{-3}$ & 0.79$\times$ \\
ST-HOSVD & approx. & 0.0683{\tiny$\pm$0.0015} & -- & 450{\tiny$\pm$0} & $2.0\times10^{-3}$ & 0.50$\times$ \\
RandHOSVD & approx. & 0.0345{\tiny$\pm$0.0021} & -- & 235{\tiny$\pm$0} & $9.2\times10^{-3}$ & 0.26$\times$ \\
RandST-HOSVD & approx. & 0.0325{\tiny$\pm$0.00016} & -- & 235{\tiny$\pm$0} & $6.9\times10^{-3}$ & 0.24$\times$ \\
TUCKER-TS & approx. & 0.603{\tiny$\pm$0.0035} & 0.00788{\tiny$\pm$0.000064} & 127{\tiny$\pm$0} & $4.6\times10^{-2}$ & 4.45$\times$ \\
TUCKER-TTMTS & approx. & 0.337{\tiny$\pm$0.0024} & 0.00337{\tiny$\pm$0.000040} & 127{\tiny$\pm$0} & $4.2\times10^{-1}$ & 2.49$\times$ \\
Two-pass sketch & approx. & 0.0255{\tiny$\pm$0.00090} & -- & 130{\tiny$\pm$0} & $8.2\times10^{-2}$ & 0.19$\times$ \\
One-pass sketch & approx. & 0.0801{\tiny$\pm$0.00033} & -- & 130{\tiny$\pm$0} & $1.2\times10^{-1}$ & 0.59$\times$ \\
D-Tucker & approx. & 0.822{\tiny$\pm$0.0069} & 0.0401{\tiny$\pm$0.00039} & 81.1{\tiny$\pm$0} & $2.4\times10^{-2}$ & 6.07$\times$ \\
\bottomrule
\end{tabular}
\end{table}

\begin{table}[H]\centering\scriptsize\setlength{\tabcolsep}{2.3pt}
\caption{All HOOI-type methods, order 3, $n=300$, $r=5$, $\kappa=100$, SNR 20\,dB, 5 seeds (mean $\pm$ s.e.). Input 206\,MiB.}
\label{tab:hooi_all_3_300}
\begin{tabular}{llrrrrr}
\toprule
Method & type & total (s) & per iter.\ (s) & peak (MiB) & rel.\ error to $\mathcal{X}^\star$ & vs.\ Ours \\
\midrule
Naive HOOI & exact & 0.335{\tiny$\pm$0.0058} & 0.00885{\tiny$\pm$0.000058} & 1{,}415{\tiny$\pm$0} & $1.3\times10^{-3}$ & 2.29$\times$ \\
TensorLy & exact & 0.492{\tiny$\pm$0.0049} & -- & 1{,}418{\tiny$\pm$0} & $1.3\times10^{-3}$ & 3.36$\times$ \\
Tensor Toolbox (eig) & exact & 0.194{\tiny$\pm$0.0060} & 0.0119{\tiny$\pm$0.00051} & 418{\tiny$\pm$0} & $1.3\times10^{-3}$ & 1.33$\times$ \\
Tensor Toolbox (eigs) & exact & 0.210{\tiny$\pm$0.00092} & 0.0163{\tiny$\pm$0.000086} & 414{\tiny$\pm$0} & $1.3\times10^{-3}$ & 1.44$\times$ \\
HOQRI & same fixed pt. & 0.0949{\tiny$\pm$0.00069} & 0.00124{\tiny$\pm$0.000012} & 418{\tiny$\pm$0} & $1.3\times10^{-3}$ & 0.65$\times$ \\
\textbf{Ours} & exact & 0.146{\tiny$\pm$0.0020} & 0.00539{\tiny$\pm$0.00016} & 217{\tiny$\pm$0} & $1.3\times10^{-3}$ & 1.00$\times$ \\
Ours, no reuse & exact & 0.146{\tiny$\pm$0.0013} & 0.00536{\tiny$\pm$0.000099} & 214{\tiny$\pm$0} & $1.3\times10^{-3}$ & 1.00$\times$ \\
Ours, Krylov HOSVD init. & exact & 0.162{\tiny$\pm$0.0014} & 0.00563{\tiny$\pm$0.00013} & 450{\tiny$\pm$0} & $1.3\times10^{-3}$ & 1.11$\times$ \\
Ours, Krylov everywhere, no reuse & exact & 0.428{\tiny$\pm$0.0026} & 0.0320{\tiny$\pm$0.00018} & 450{\tiny$\pm$0} & $1.3\times10^{-3}$ & 2.92$\times$ \\
Ours, no reuse, Gram via unfolding copy & exact & 0.139{\tiny$\pm$0.0026} & 0.00574{\tiny$\pm$0.00018} & 418{\tiny$\pm$0} & $1.3\times10^{-3}$ & 0.95$\times$ \\
HOSVD & approx. & 0.244{\tiny$\pm$0.0065} & -- & 1{,}416{\tiny$\pm$0} & $1.4\times10^{-3}$ & 1.67$\times$ \\
ST-HOSVD & approx. & 0.114{\tiny$\pm$0.0015} & -- & 1{,}415{\tiny$\pm$0} & $1.3\times10^{-3}$ & 0.78$\times$ \\
RandHOSVD & approx. & 0.0366{\tiny$\pm$0.00054} & -- & 661{\tiny$\pm$0} & $8.5\times10^{-3}$ & 0.25$\times$ \\
RandST-HOSVD & approx. & 0.0251{\tiny$\pm$0.00023} & -- & 661{\tiny$\pm$0} & $6.2\times10^{-3}$ & 0.17$\times$ \\
TUCKER-TS & approx. & 0.568{\tiny$\pm$0.034} & 0.00787{\tiny$\pm$0.000043} & 419{\tiny$\pm$0} & $4.9\times10^{-2}$ & 3.88$\times$ \\
TUCKER-TTMTS & approx. & 0.344{\tiny$\pm$0.0021} & 0.00331{\tiny$\pm$0.000030} & 419{\tiny$\pm$0} & $4.4\times10^{-1}$ & 2.35$\times$ \\
Two-pass sketch & approx. & 0.0220{\tiny$\pm$0.00038} & -- & 426{\tiny$\pm$0} & $8.8\times10^{-2}$ & 0.15$\times$ \\
One-pass sketch & approx. & 0.0750{\tiny$\pm$0.00032} & -- & 426{\tiny$\pm$0} & $1.3\times10^{-1}$ & 0.51$\times$ \\
D-Tucker & approx. & 1.19{\tiny$\pm$0.0044} & 0.0602{\tiny$\pm$0.0012} & 244{\tiny$\pm$0} & $2.4\times10^{-2}$ & 8.12$\times$ \\
\bottomrule
\end{tabular}
\end{table}

\begin{table}[H]\centering\scriptsize\setlength{\tabcolsep}{2.3pt}
\caption{All HOOI-type methods, order 3, $n=500$, $r=5$, $\kappa=100$, SNR 20\,dB, 5 seeds (mean $\pm$ s.e.). Input 954\,MiB.}
\label{tab:hooi_all_3_500}
\begin{tabular}{llrrrrr}
\toprule
Method & type & total (s) & per iter.\ (s) & peak (MiB) & rel.\ error to $\mathcal{X}^\star$ & vs.\ Ours \\
\midrule
Naive HOOI & exact & 1.31{\tiny$\pm$0.0020} & 0.0275{\tiny$\pm$0.000036} & 6{,}218{\tiny$\pm$0} & $7.8\times10^{-4}$ & 7.22$\times$ \\
TensorLy & exact & 1.31{\tiny$\pm$0.0048} & -- & 6{,}228{\tiny$\pm$0} & $7.8\times10^{-4}$ & 7.25$\times$ \\
Tensor Toolbox (eig) & exact & 0.304{\tiny$\pm$0.010} & 0.0208{\tiny$\pm$0.0010} & 1{,}923{\tiny$\pm$0} & $7.8\times10^{-4}$ & 1.68$\times$ \\
Tensor Toolbox (eigs) & exact & 0.217{\tiny$\pm$0.0023} & 0.0155{\tiny$\pm$0.00015} & 1{,}914{\tiny$\pm$0} & $7.8\times10^{-4}$ & 1.20$\times$ \\
HOQRI & same fixed pt. & 0.153{\tiny$\pm$0.00079} & 0.00296{\tiny$\pm$0.000013} & 1{,}923{\tiny$\pm$0} & $7.8\times10^{-4}$ & 0.85$\times$ \\
\textbf{Ours} & exact & 0.181{\tiny$\pm$0.0030} & 0.00697{\tiny$\pm$0.00020} & 974{\tiny$\pm$0} & $7.8\times10^{-4}$ & 1.00$\times$ \\
Ours, no reuse & exact & 0.186{\tiny$\pm$0.0029} & 0.00768{\tiny$\pm$0.00020} & 971{\tiny$\pm$0} & $7.8\times10^{-4}$ & 1.03$\times$ \\
Ours, Krylov HOSVD init. & exact & 0.244{\tiny$\pm$0.0028} & 0.00697{\tiny$\pm$0.00013} & 1{,}624{\tiny$\pm$1.16} & $7.8\times10^{-4}$ & 1.35$\times$ \\
Ours, Krylov everywhere, no reuse & exact & 0.505{\tiny$\pm$0.0020} & 0.0330{\tiny$\pm$0.00014} & 1{,}624{\tiny$\pm$1.16} & $7.8\times10^{-4}$ & 2.79$\times$ \\
Ours, no reuse, Gram via unfolding copy & exact & 0.192{\tiny$\pm$0.0042} & 0.00763{\tiny$\pm$0.00022} & 1{,}923{\tiny$\pm$0} & $7.8\times10^{-4}$ & 1.06$\times$ \\
HOSVD & approx. & 1.03{\tiny$\pm$0.0025} & -- & 6{,}222{\tiny$\pm$0} & $8.0\times10^{-4}$ & 5.72$\times$ \\
ST-HOSVD & approx. & 0.399{\tiny$\pm$0.0042} & -- & 6{,}218{\tiny$\pm$0} & $7.9\times10^{-4}$ & 2.21$\times$ \\
RandHOSVD & approx. & 0.0794{\tiny$\pm$0.0030} & -- & 2{,}862{\tiny$\pm$0} & $7.2\times10^{-3}$ & 0.44$\times$ \\
RandST-HOSVD & approx. & 0.0367{\tiny$\pm$0.00080} & -- & 2{,}597{\tiny$\pm$0} & $4.9\times10^{-3}$ & 0.20$\times$ \\
TUCKER-TS & approx. & 0.623{\tiny$\pm$0.0010} & 0.00781{\tiny$\pm$0.000025} & 1{,}927{\tiny$\pm$0} & $4.7\times10^{-2}$ & 3.44$\times$ \\
TUCKER-TTMTS & approx. & 0.405{\tiny$\pm$0.0040} & 0.00335{\tiny$\pm$0.000035} & 1{,}927{\tiny$\pm$0} & $4.3\times10^{-1}$ & 2.24$\times$ \\
Two-pass sketch & approx. & 0.0289{\tiny$\pm$0.00020} & -- & 1{,}947{\tiny$\pm$0} & $8.3\times10^{-2}$ & 0.16$\times$ \\
One-pass sketch & approx. & 0.0781{\tiny$\pm$0.00040} & -- & 1{,}947{\tiny$\pm$0} & $1.2\times10^{-1}$ & 0.43$\times$ \\
D-Tucker & approx. & 1.91{\tiny$\pm$0.0046} & 0.102{\tiny$\pm$0.00046} & 1{,}054{\tiny$\pm$0} & $2.4\times10^{-2}$ & 10.58$\times$ \\
\bottomrule
\end{tabular}
\end{table}

\begin{table}[H]\centering\scriptsize\setlength{\tabcolsep}{2.3pt}
\caption{All HOOI-type methods, order 3, $n=800$, $r=5$, $\kappa=100$, SNR 20\,dB, 5 seeds (mean $\pm$ s.e.). Input 3{,}906\,MiB.}
\label{tab:hooi_all_3_800}
\begin{tabular}{llrrrrr}
\toprule
Method & type & total (s) & per iter.\ (s) & peak (MiB) & rel.\ error to $\mathcal{X}^\star$ & vs.\ Ours \\
\midrule
Naive HOOI & exact & 5.79{\tiny$\pm$0.0038} & 0.114{\tiny$\pm$0.000028} & 24{,}702{\tiny$\pm$0} & $4.9\times10^{-4}$ & 15.12$\times$ \\
TensorLy & exact & 5.02{\tiny$\pm$0.0032} & -- & 24{,}726{\tiny$\pm$0} & $4.9\times10^{-4}$ & 13.10$\times$ \\
Tensor Toolbox (eig) & exact & 0.636{\tiny$\pm$0.0027} & 0.0377{\tiny$\pm$0.000014} & 7{,}848{\tiny$\pm$0} & $4.9\times10^{-4}$ & 1.66$\times$ \\
Tensor Toolbox (eigs) & exact & 0.438{\tiny$\pm$0.0042} & 0.0223{\tiny$\pm$0.000082} & 7{,}827{\tiny$\pm$0} & $4.9\times10^{-4}$ & 1.14$\times$ \\
HOQRI & same fixed pt. & 0.430{\tiny$\pm$0.0027} & 0.00993{\tiny$\pm$0.0000024} & 7{,}848{\tiny$\pm$0} & $4.9\times10^{-4}$ & 1.12$\times$ \\
\textbf{Ours} & exact & 0.383{\tiny$\pm$0.0033} & 0.0120{\tiny$\pm$0.000025} & 3{,}948{\tiny$\pm$0} & $4.9\times10^{-4}$ & 1.00$\times$ \\
Ours, no reuse & exact & 0.409{\tiny$\pm$0.0018} & 0.0151{\tiny$\pm$0.000027} & 3{,}947{\tiny$\pm$0} & $4.9\times10^{-4}$ & 1.07$\times$ \\
Ours, Krylov HOSVD init. & exact & 0.513{\tiny$\pm$0.0025} & 0.0120{\tiny$\pm$0.000019} & 5{,}614{\tiny$\pm$2.62} & $4.9\times10^{-4}$ & 1.34$\times$ \\
Ours, Krylov everywhere, no reuse & exact & 0.773{\tiny$\pm$0.0048} & 0.0377{\tiny$\pm$0.00022} & 5{,}614{\tiny$\pm$2.62} & $4.9\times10^{-4}$ & 2.02$\times$ \\
Ours, no reuse, Gram via unfolding copy & exact & 0.478{\tiny$\pm$0.0033} & 0.0151{\tiny$\pm$0.000017} & 7{,}848{\tiny$\pm$0} & $4.9\times10^{-4}$ & 1.25$\times$ \\
HOSVD & approx. & 4.66{\tiny$\pm$0.0030} & -- & 24{,}712{\tiny$\pm$0} & $4.9\times10^{-4}$ & 12.17$\times$ \\
ST-HOSVD & approx. & 1.66{\tiny$\pm$0.0016} & -- & 24{,}702{\tiny$\pm$0} & $4.9\times10^{-4}$ & 4.33$\times$ \\
RandHOSVD & approx. & 0.256{\tiny$\pm$0.0038} & -- & 11{,}719{\tiny$\pm$0} & $5.5\times10^{-3}$ & 0.67$\times$ \\
RandST-HOSVD & approx. & 0.0786{\tiny$\pm$0.00079} & -- & 9{,}581{\tiny$\pm$0} & $3.7\times10^{-3}$ & 0.21$\times$ \\
TUCKER-TS & approx. & 0.712{\tiny$\pm$0.0036} & 0.00765{\tiny$\pm$0.000042} & 7{,}859{\tiny$\pm$0} & $4.5\times10^{-2}$ & 1.86$\times$ \\
TUCKER-TTMTS & approx. & 0.606{\tiny$\pm$0.0037} & 0.00323{\tiny$\pm$0.000013} & 7{,}859{\tiny$\pm$0} & $4.5\times10^{-1}$ & 1.58$\times$ \\
Two-pass sketch & approx. & 0.0397{\tiny$\pm$0.00011} & -- & 7{,}911{\tiny$\pm$0} & $8.6\times10^{-2}$ & 0.10$\times$ \\
One-pass sketch & approx. & 0.0857{\tiny$\pm$0.00022} & -- & 7{,}911{\tiny$\pm$0} & $1.3\times10^{-1}$ & 0.22$\times$ \\
D-Tucker & approx. & 3.14{\tiny$\pm$0.010} & 0.159{\tiny$\pm$0.0014} & 4{,}154{\tiny$\pm$0} & $2.4\times10^{-2}$ & 8.21$\times$ \\
\bottomrule
\end{tabular}
\end{table}

\begin{table}[H]\centering\scriptsize\setlength{\tabcolsep}{2.3pt}
\caption{All HOOI-type methods, order 3, $n=1000$, $r=5$, $\kappa=100$, SNR 20\,dB, 5 seeds (mean $\pm$ s.e.). Input 7{,}629\,MiB. OOM: out of GPU memory. fail$^\S$: cuSOLVER error in the SVD of the $n\times n^{N-1}$ unfolding.}
\label{tab:hooi_all_3_1000}
\begin{tabular}{llrrrrr}
\toprule
Method & type & total (s) & per iter.\ (s) & peak (MiB) & rel.\ error to $\mathcal{X}^\star$ & vs.\ Ours \\
\midrule
Naive HOOI & exact & fail$^\S$ & & & & \\
TensorLy & exact & fail$^\S$ & & & & \\
Tensor Toolbox (eig) & exact & 1.02{\tiny$\pm$0.0063} & 0.0570{\tiny$\pm$0.000021} & 15{,}315{\tiny$\pm$0} & $3.9\times10^{-4}$ & 1.50$\times$ \\
Tensor Toolbox (eigs) & exact & 0.709{\tiny$\pm$0.0067} & 0.0316{\tiny$\pm$0.00011} & 15{,}283{\tiny$\pm$0} & $3.9\times10^{-4}$ & 1.05$\times$ \\
HOQRI & same fixed pt. & 0.782{\tiny$\pm$0.0034} & 0.0191{\tiny$\pm$0.0000019} & 15{,}315{\tiny$\pm$0} & $3.9\times10^{-4}$ & 1.15$\times$ \\
\textbf{Ours} & exact & 0.678{\tiny$\pm$0.0026} & 0.0183{\tiny$\pm$0.000012} & 7{,}693{\tiny$\pm$0} & $3.9\times10^{-4}$ & 1.00$\times$ \\
Ours, no reuse & exact & 0.747{\tiny$\pm$0.0029} & 0.0245{\tiny$\pm$0.000024} & 7{,}693{\tiny$\pm$0} & $3.9\times10^{-4}$ & 1.10$\times$ \\
Ours, Krylov HOSVD init. & exact & 0.870{\tiny$\pm$0.0038} & 0.0183{\tiny$\pm$0.000020} & 10{,}293{\tiny$\pm$5.19} & $3.9\times10^{-4}$ & 1.28$\times$ \\
Ours, Krylov everywhere, no reuse & exact & 1.16{\tiny$\pm$0.0055} & 0.0473{\tiny$\pm$0.000100} & 10{,}293{\tiny$\pm$5.19} & $3.9\times10^{-4}$ & 1.72$\times$ \\
Ours, no reuse, Gram via unfolding copy & exact & 0.839{\tiny$\pm$0.0021} & 0.0245{\tiny$\pm$0.000022} & 15{,}315{\tiny$\pm$0} & $3.9\times10^{-4}$ & 1.24$\times$ \\
HOSVD & approx. & fail$^\S$ & & & & \\
ST-HOSVD & approx. & fail$^\S$ & & & & \\
RandHOSVD & approx. & 0.425{\tiny$\pm$0.0014} & -- & 22{,}890{\tiny$\pm$0} & $4.8\times10^{-3}$ & 0.63$\times$ \\
RandST-HOSVD & approx. & 0.130{\tiny$\pm$0.0018} & -- & 18{,}015{\tiny$\pm$0} & $2.9\times10^{-3}$ & 0.19$\times$ \\
TUCKER-TS & approx. & 0.838{\tiny$\pm$0.0059} & 0.00767{\tiny$\pm$0.000048} & 15{,}333{\tiny$\pm$0} & $5.0\times10^{-2}$ & 1.24$\times$ \\
TUCKER-TTMTS & approx. & 0.843{\tiny$\pm$0.0043} & 0.00326{\tiny$\pm$0.000020} & 15{,}333{\tiny$\pm$0} & $4.0\times10^{-1}$ & 1.24$\times$ \\
Two-pass sketch & approx. & 0.0513{\tiny$\pm$0.00027} & -- & 15{,}414{\tiny$\pm$0} & $8.6\times10^{-2}$ & 0.08$\times$ \\
One-pass sketch & approx. & 0.107{\tiny$\pm$0.00014} & -- & 15{,}414{\tiny$\pm$0} & $1.3\times10^{-1}$ & 0.16$\times$ \\
D-Tucker & approx. & 3.96{\tiny$\pm$0.0039} & 0.201{\tiny$\pm$0.0012} & 8{,}013{\tiny$\pm$0} & $2.5\times10^{-2}$ & 5.85$\times$ \\
\bottomrule
\end{tabular}
\end{table}

\begin{table}[H]\centering\scriptsize\setlength{\tabcolsep}{2.3pt}
\caption{All HOOI-type methods, order 4, $n=20$, $r=3$, $\kappa=100$, SNR 20\,dB, 5 seeds (mean $\pm$ s.e.). Input 1.22\,MiB.}
\label{tab:hooi_all_4_20}
\begin{tabular}{llrrrrr}
\toprule
Method & type & total (s) & per iter.\ (s) & peak (MiB) & rel.\ error to $\mathcal{X}^\star$ & vs.\ Ours \\
\midrule
Naive HOOI & exact & 0.0727{\tiny$\pm$0.0051} & 0.00340{\tiny$\pm$0.000044} & 23.5{\tiny$\pm$0} & $4.2\times10^{-3}$ & 0.53$\times$ \\
TensorLy & exact & 0.280{\tiny$\pm$0.0056} & -- & 23.7{\tiny$\pm$0} & $4.2\times10^{-3}$ & 2.05$\times$ \\
Tensor Toolbox (eig) & exact & 0.124{\tiny$\pm$0.00071} & 0.00325{\tiny$\pm$0.000040} & 3.08{\tiny$\pm$0} & $4.2\times10^{-3}$ & 0.91$\times$ \\
Tensor Toolbox (eigs) & exact & 0.337{\tiny$\pm$0.0019} & 0.0249{\tiny$\pm$0.00018} & 2.45{\tiny$\pm$0} & $4.2\times10^{-3}$ & 2.47$\times$ \\
HOQRI & same fixed pt. & 0.113{\tiny$\pm$0.0018} & 0.00183{\tiny$\pm$0.000013} & 7.23{\tiny$\pm$0} & $4.2\times10^{-3}$ & 0.83$\times$ \\
\textbf{Ours} & exact & 0.137{\tiny$\pm$0.0015} & 0.00336{\tiny$\pm$0.000051} & 1.86{\tiny$\pm$0} & $4.2\times10^{-3}$ & 1.00$\times$ \\
Ours, no reuse & exact & 0.138{\tiny$\pm$0.00065} & 0.00361{\tiny$\pm$0.000052} & 1.86{\tiny$\pm$0} & $4.2\times10^{-3}$ & 1.01$\times$ \\
Ours, Krylov HOSVD init. & exact & 0.187{\tiny$\pm$0.00086} & 0.00337{\tiny$\pm$0.000018} & 24.4{\tiny$\pm$0.00022} & $4.2\times10^{-3}$ & 1.37$\times$ \\
Ours, Krylov everywhere, no reuse & exact & 0.553{\tiny$\pm$0.0049} & 0.0394{\tiny$\pm$0.00045} & 24.4{\tiny$\pm$0.00022} & $4.2\times10^{-3}$ & 4.05$\times$ \\
Ours, no reuse, Gram via unfolding copy & exact & 0.130{\tiny$\pm$0.0011} & 0.00381{\tiny$\pm$0.000094} & 3.08{\tiny$\pm$0} & $4.2\times10^{-3}$ & 0.96$\times$ \\
HOSVD & approx. & 0.0308{\tiny$\pm$0.00012} & -- & 23.5{\tiny$\pm$0} & $5.9\times10^{-3}$ & 0.23$\times$ \\
ST-HOSVD & approx. & 0.0314{\tiny$\pm$0.00019} & -- & 23.5{\tiny$\pm$0} & $4.8\times10^{-3}$ & 0.23$\times$ \\
RandHOSVD & approx. & 0.0329{\tiny$\pm$0.0023} & -- & 24.4{\tiny$\pm$0} & $9.7\times10^{-3}$ & 0.24$\times$ \\
RandST-HOSVD & approx. & 0.0285{\tiny$\pm$0.00031} & -- & 24.4{\tiny$\pm$0} & $7.4\times10^{-3}$ & 0.21$\times$ \\
TUCKER-TS & approx. & 0.669{\tiny$\pm$0.0034} & 0.00925{\tiny$\pm$0.000067} & 8.79{\tiny$\pm$0} & $4.4\times10^{-2}$ & 4.90$\times$ \\
TUCKER-TTMTS & approx. & 0.361{\tiny$\pm$0.00059} & 0.00415{\tiny$\pm$0.000031} & 7.58{\tiny$\pm$0} & $4.7\times10^{-1}$ & 2.65$\times$ \\
Two-pass sketch & approx. & 0.0422{\tiny$\pm$0.0059} & -- & 8.03{\tiny$\pm$0} & $8.9\times10^{-2}$ & 0.31$\times$ \\
One-pass sketch & approx. & 0.0884{\tiny$\pm$0.00028} & -- & 11.5{\tiny$\pm$0} & $1.3\times10^{-1}$ & 0.65$\times$ \\
D-Tucker & approx. & 1.10{\tiny$\pm$0.0073} & 0.0722{\tiny$\pm$0.00095} & 8.94{\tiny$\pm$0} & $2.5\times10^{-1}$ & 8.09$\times$ \\
\bottomrule
\end{tabular}
\end{table}

\begin{table}[H]\centering\scriptsize\setlength{\tabcolsep}{2.3pt}
\caption{All HOOI-type methods, order 4, $n=40$, $r=3$, $\kappa=100$, SNR 20\,dB, 5 seeds (mean $\pm$ s.e.). Input 19.5\,MiB.}
\label{tab:hooi_all_4_40}
\begin{tabular}{llrrrrr}
\toprule
Method & type & total (s) & per iter.\ (s) & peak (MiB) & rel.\ error to $\mathcal{X}^\star$ & vs.\ Ours \\
\midrule
Naive HOOI & exact & 0.124{\tiny$\pm$0.0015} & 0.00686{\tiny$\pm$0.000081} & 245{\tiny$\pm$0} & $1.4\times10^{-3}$ & 0.62$\times$ \\
TensorLy & exact & 0.330{\tiny$\pm$0.0023} & -- & 246{\tiny$\pm$0} & $1.4\times10^{-3}$ & 1.64$\times$ \\
Tensor Toolbox (eig) & exact & 0.144{\tiny$\pm$0.0019} & 0.00525{\tiny$\pm$0.00013} & 40.8{\tiny$\pm$0} & $1.4\times10^{-3}$ & 0.71$\times$ \\
Tensor Toolbox (eigs) & exact & 0.282{\tiny$\pm$0.0017} & 0.0199{\tiny$\pm$0.00022} & 40.0{\tiny$\pm$0} & $1.4\times10^{-3}$ & 1.40$\times$ \\
HOQRI & same fixed pt. & 0.110{\tiny$\pm$0.00099} & 0.00183{\tiny$\pm$0.0000023} & 40.8{\tiny$\pm$0} & $1.4\times10^{-3}$ & 0.55$\times$ \\
\textbf{Ours} & exact & 0.201{\tiny$\pm$0.0021} & 0.00686{\tiny$\pm$0.00020} & 26.2{\tiny$\pm$0} & $1.4\times10^{-3}$ & 1.00$\times$ \\
Ours, no reuse & exact & 0.204{\tiny$\pm$0.0032} & 0.00708{\tiny$\pm$0.00017} & 26.1{\tiny$\pm$0} & $1.4\times10^{-3}$ & 1.01$\times$ \\
Ours, Krylov HOSVD init. & exact & 0.210{\tiny$\pm$0.0042} & 0.00684{\tiny$\pm$0.00018} & 195{\tiny$\pm$0.62} & $1.4\times10^{-3}$ & 1.05$\times$ \\
Ours, Krylov everywhere, no reuse & exact & 0.504{\tiny$\pm$0.0018} & 0.0363{\tiny$\pm$0.00019} & 195{\tiny$\pm$0.62} & $1.4\times10^{-3}$ & 2.51$\times$ \\
Ours, no reuse, Gram via unfolding copy & exact & 0.162{\tiny$\pm$0.00087} & 0.00701{\tiny$\pm$0.00011} & 40.8{\tiny$\pm$0} & $1.4\times10^{-3}$ & 0.80$\times$ \\
HOSVD & approx. & 0.0532{\tiny$\pm$0.00069} & -- & 245{\tiny$\pm$0} & $1.9\times10^{-3}$ & 0.26$\times$ \\
ST-HOSVD & approx. & 0.0417{\tiny$\pm$0.0023} & -- & 245{\tiny$\pm$0} & $1.6\times10^{-3}$ & 0.21$\times$ \\
RandHOSVD & approx. & 0.0438{\tiny$\pm$0.0021} & -- & 211{\tiny$\pm$0} & $9.7\times10^{-3}$ & 0.22$\times$ \\
RandST-HOSVD & approx. & 0.0367{\tiny$\pm$0.00022} & -- & 211{\tiny$\pm$0} & $7.0\times10^{-3}$ & 0.18$\times$ \\
TUCKER-TS & approx. & 0.680{\tiny$\pm$0.0046} & 0.00945{\tiny$\pm$0.000092} & 44.6{\tiny$\pm$0} & $4.3\times10^{-2}$ & 3.38$\times$ \\
TUCKER-TTMTS & approx. & 0.366{\tiny$\pm$0.0050} & 0.00417{\tiny$\pm$0.000035} & 44.6{\tiny$\pm$0} & $5.2\times10^{-1}$ & 1.82$\times$ \\
Two-pass sketch & approx. & 0.0390{\tiny$\pm$0.0036} & -- & 65.9{\tiny$\pm$0} & $1.1\times10^{-1}$ & 0.19$\times$ \\
One-pass sketch & approx. & 0.0893{\tiny$\pm$0.0023} & -- & 65.9{\tiny$\pm$0} & $1.5\times10^{-1}$ & 0.44$\times$ \\
D-Tucker & approx. & 3.93{\tiny$\pm$0.0082} & 0.281{\tiny$\pm$0.00060} & 44.4{\tiny$\pm$0} & $1.8\times10^{-2}$ & 19.53$\times$ \\
\bottomrule
\end{tabular}
\end{table}

\begin{table}[H]\centering\scriptsize\setlength{\tabcolsep}{2.3pt}
\caption{All HOOI-type methods, order 4, $n=60$, $r=3$, $\kappa=100$, SNR 20\,dB, 5 seeds (mean $\pm$ s.e.). Input 98.9\,MiB.}
\label{tab:hooi_all_4_60}
\begin{tabular}{llrrrrr}
\toprule
Method & type & total (s) & per iter.\ (s) & peak (MiB) & rel.\ error to $\mathcal{X}^\star$ & vs.\ Ours \\
\midrule
Naive HOOI & exact & 0.256{\tiny$\pm$0.0063} & 0.00926{\tiny$\pm$0.00027} & 1{,}016{\tiny$\pm$0} & $7.6\times10^{-4}$ & 1.05$\times$ \\
TensorLy & exact & 0.426{\tiny$\pm$0.0033} & -- & 1{,}021{\tiny$\pm$0} & $7.6\times10^{-4}$ & 1.76$\times$ \\
Tensor Toolbox (eig) & exact & 0.143{\tiny$\pm$0.0017} & 0.00632{\tiny$\pm$0.00017} & 199{\tiny$\pm$0} & $7.6\times10^{-4}$ & 0.59$\times$ \\
Tensor Toolbox (eigs) & exact & 0.253{\tiny$\pm$0.0023} & 0.0186{\tiny$\pm$0.00020} & 198{\tiny$\pm$0} & $7.6\times10^{-4}$ & 1.04$\times$ \\
HOQRI & same fixed pt. & 0.101{\tiny$\pm$0.00034} & 0.00188{\tiny$\pm$0.000014} & 199{\tiny$\pm$0} & $7.6\times10^{-4}$ & 0.42$\times$ \\
\textbf{Ours} & exact & 0.243{\tiny$\pm$0.0019} & 0.00722{\tiny$\pm$0.00015} & 105{\tiny$\pm$0} & $7.6\times10^{-4}$ & 1.00$\times$ \\
Ours, no reuse & exact & 0.243{\tiny$\pm$0.0019} & 0.00733{\tiny$\pm$0.00020} & 105{\tiny$\pm$0} & $7.6\times10^{-4}$ & 1.00$\times$ \\
Ours, Krylov HOSVD init. & exact & 0.202{\tiny$\pm$0.0026} & 0.00696{\tiny$\pm$0.00015} & 677{\tiny$\pm$1.21} & $7.6\times10^{-4}$ & 0.83$\times$ \\
Ours, Krylov everywhere, no reuse & exact & 0.493{\tiny$\pm$0.0015} & 0.0360{\tiny$\pm$0.00016} & 677{\tiny$\pm$1.21} & $7.6\times10^{-4}$ & 2.03$\times$ \\
Ours, no reuse, Gram via unfolding copy & exact & 0.155{\tiny$\pm$0.0025} & 0.00722{\tiny$\pm$0.00019} & 199{\tiny$\pm$0} & $7.6\times10^{-4}$ & 0.64$\times$ \\
HOSVD & approx. & 0.158{\tiny$\pm$0.0031} & -- & 1{,}016{\tiny$\pm$0} & $9.4\times10^{-4}$ & 0.65$\times$ \\
ST-HOSVD & approx. & 0.0623{\tiny$\pm$0.0014} & -- & 1{,}016{\tiny$\pm$0} & $8.0\times10^{-4}$ & 0.26$\times$ \\
RandHOSVD & approx. & 0.0611{\tiny$\pm$0.0017} & -- & 774{\tiny$\pm$0} & $9.8\times10^{-3}$ & 0.25$\times$ \\
RandST-HOSVD & approx. & 0.0361{\tiny$\pm$0.00040} & -- & 774{\tiny$\pm$0} & $7.2\times10^{-3}$ & 0.15$\times$ \\
TUCKER-TS & approx. & 0.676{\tiny$\pm$0.0048} & 0.00960{\tiny$\pm$0.000098} & 213{\tiny$\pm$0} & $4.3\times10^{-2}$ & 2.78$\times$ \\
TUCKER-TTMTS & approx. & 0.359{\tiny$\pm$0.0022} & 0.00423{\tiny$\pm$0.000034} & 213{\tiny$\pm$0} & $5.1\times10^{-1}$ & 1.48$\times$ \\
Two-pass sketch & approx. & 0.0245{\tiny$\pm$0.00054} & -- & 317{\tiny$\pm$0} & $1.0\times10^{-1}$ & 0.10$\times$ \\
One-pass sketch & approx. & 0.0767{\tiny$\pm$0.00043} & -- & 317{\tiny$\pm$0} & $1.4\times10^{-1}$ & 0.32$\times$ \\
D-Tucker & approx. & 8.73{\tiny$\pm$0.043} & 0.638{\tiny$\pm$0.0056} & 169{\tiny$\pm$0} & $1.8\times10^{-2}$ & 35.97$\times$ \\
\bottomrule
\end{tabular}
\end{table}

\begin{table}[H]\centering\scriptsize\setlength{\tabcolsep}{2.3pt}
\caption{All HOOI-type methods, order 4, $n=80$, $r=3$, $\kappa=100$, SNR 20\,dB, 5 seeds (mean $\pm$ s.e.). Input 312\,MiB.}
\label{tab:hooi_all_4_80}
\begin{tabular}{llrrrrr}
\toprule
Method & type & total (s) & per iter.\ (s) & peak (MiB) & rel.\ error to $\mathcal{X}^\star$ & vs.\ Ours \\
\midrule
Naive HOOI & exact & 0.595{\tiny$\pm$0.0056} & 0.0180{\tiny$\pm$0.000036} & 2{,}876{\tiny$\pm$0} & $4.9\times10^{-4}$ & 1.93$\times$ \\
TensorLy & exact & 0.667{\tiny$\pm$0.0023} & -- & 2{,}888{\tiny$\pm$0} & $4.9\times10^{-4}$ & 2.16$\times$ \\
Tensor Toolbox (eig) & exact & 0.171{\tiny$\pm$0.0038} & 0.00830{\tiny$\pm$0.00034} & 627{\tiny$\pm$0} & $4.9\times10^{-4}$ & 0.55$\times$ \\
Tensor Toolbox (eigs) & exact & 0.240{\tiny$\pm$0.0022} & 0.0168{\tiny$\pm$0.00019} & 625{\tiny$\pm$0} & $4.9\times10^{-4}$ & 0.78$\times$ \\
HOQRI & same fixed pt. & 0.113{\tiny$\pm$0.00092} & 0.00184{\tiny$\pm$0.0000077} & 627{\tiny$\pm$0} & $4.9\times10^{-4}$ & 0.37$\times$ \\
\textbf{Ours} & exact & 0.309{\tiny$\pm$0.00047} & 0.00683{\tiny$\pm$0.000052} & 325{\tiny$\pm$0} & $4.9\times10^{-4}$ & 1.00$\times$ \\
Ours, no reuse & exact & 0.316{\tiny$\pm$0.0017} & 0.00733{\tiny$\pm$0.00020} & 325{\tiny$\pm$0} & $4.9\times10^{-4}$ & 1.02$\times$ \\
Ours, Krylov HOSVD init. & exact & 0.251{\tiny$\pm$0.0033} & 0.00672{\tiny$\pm$0.00013} & 1{,}670{\tiny$\pm$0.00078} & $4.9\times10^{-4}$ & 0.81$\times$ \\
Ours, Krylov everywhere, no reuse & exact & 0.545{\tiny$\pm$0.0048} & 0.0363{\tiny$\pm$0.00024} & 1{,}670{\tiny$\pm$0.00078} & $4.9\times10^{-4}$ & 1.77$\times$ \\
Ours, no reuse, Gram via unfolding copy & exact & 0.170{\tiny$\pm$0.0021} & 0.00752{\tiny$\pm$0.00015} & 627{\tiny$\pm$0} & $4.9\times10^{-4}$ & 0.55$\times$ \\
HOSVD & approx. & 0.420{\tiny$\pm$0.0056} & -- & 2{,}876{\tiny$\pm$0} & $5.8\times10^{-4}$ & 1.36$\times$ \\
ST-HOSVD & approx. & 0.125{\tiny$\pm$0.0034} & -- & 2{,}876{\tiny$\pm$0} & $5.1\times10^{-4}$ & 0.41$\times$ \\
RandHOSVD & approx. & 0.136{\tiny$\pm$0.0043} & -- & 1{,}981{\tiny$\pm$0} & $9.6\times10^{-3}$ & 0.44$\times$ \\
RandST-HOSVD & approx. & 0.0539{\tiny$\pm$0.0011} & -- & 1{,}981{\tiny$\pm$0} & $7.0\times10^{-3}$ & 0.17$\times$ \\
TUCKER-TS & approx. & 0.689{\tiny$\pm$0.0048} & 0.00975{\tiny$\pm$0.000085} & 661{\tiny$\pm$0} & $4.3\times10^{-2}$ & 2.23$\times$ \\
TUCKER-TTMTS & approx. & 0.370{\tiny$\pm$0.0031} & 0.00424{\tiny$\pm$0.000044} & 661{\tiny$\pm$0} & $4.8\times10^{-1}$ & 1.20$\times$ \\
Two-pass sketch & approx. & 0.0255{\tiny$\pm$0.000078} & -- & 984{\tiny$\pm$0} & $1.6\times10^{-1}$ & 0.08$\times$ \\
One-pass sketch & approx. & 0.0786{\tiny$\pm$0.00071} & -- & 984{\tiny$\pm$0} & $2.2\times10^{-1}$ & 0.25$\times$ \\
D-Tucker & approx. & 16.0{\tiny$\pm$0.042} & 1.12{\tiny$\pm$0.0057} & 464{\tiny$\pm$0} & $1.8\times10^{-2}$ & 51.72$\times$ \\
\bottomrule
\end{tabular}
\end{table}

\section{When Algorithm 1 Pays Off}
\label{sec:appendix_crossover}

Table~\ref{tab:crossover} compares Algorithm~\ref{alg:component_extraction} with a dense SVD (\texttt{torch.linalg.svd}) for computing the top $s$ singular triplets of the mode-2 unfolding ($n\times n^2$) of an order-3 tensor. The unfolding is given in three forms: explicit (the unfolding is formed), copy-free (products with the dense tensor, as in HaLRTC) and sparse (the observed entries in compressed sparse row (CSR) form, as in the Frank--Wolfe oracle). For the copy-free and sparse forms, the dense SVD first forms the unfolding, and we include that time and memory. The data are the synthetic tensors of Section~\ref{sec:five} with noise at an SNR of 20\,dB, or tensors with a power-law spectrum $\sigma_i=1/i$. We use $n\in\{50,100,200,300,500\}$, $s$ from 1 to $0.9n$, the Krylov tolerance $10^{-10}$ and \xoSeeds{} seeds (\xoRuns{} runs on the A100, protocol of Appendix~\ref{sec:appendix_experimental_details}). All extracted triplets agree with the dense SVD to $10^{-8}$.

Algorithm~\ref{alg:component_extraction} is faster only when $n\ge\xoNstar$ and very few triplets are needed. Specifically, $s\le5$ in \xoFastFew{} of the \xoFast{} faster cases, and no faster case has $s>\xoSnMax\,n$. For $n\le\xoNsmall$, its fixed cost per call (\xoAlgFixedLo--\xoAlgFixedHi\,ms at $s=1$, against \xoSvdSmallLo--\xoSvdSmallHi\,ms for the dense SVD) makes it slower at every $s$. As $s$ grows, the Krylov dimension approaches $n$. For $s\ge n/2$, the algorithm is therefore $\xoSlowLo$--$\xoSlowHi\times$ slower. Its GPU memory also loses its advantage: for $s\le5$, it is $\xoMemFewLo$--$\xoMemFewHi\times$ less than that of forming the unfolding (copy-free and sparse forms, $n=\xoNmax$), but for $s\ge n/2$, it becomes about the same ($\xoMemHalfLo$--$\xoMemHalfHi\times$). For the small explicit matrices of HOOI and the subspace norm ($n\times r^2$ with $s=r$, $r\in\{5,20\}$), the dense SVD is always faster ($\xoSmallSlowLo$--$\xoSmallSlowHi\times$), as the fixed rule of Section~\ref{sec:method_components} assumes. Therefore, the rule pays off when few triplets are needed, as in the Frank--Wolfe oracle ($s=1$), but not when nearly all are needed, as in HaLRTC at $p=0.05$ (Section~\ref{sec:five_limits}).

\begin{table}[H]
\caption{Algorithm~\ref{alg:component_extraction} against a dense SVD on an A100 for the top $s$ singular triplets of $n\times n^2$ unfoldings ($n=\xoNmax$, float64, median of \xoSeeds{} seeds). Each entry is the time of the dense SVD divided by that of Algorithm~\ref{alg:component_extraction}; for the copy-free and sparse forms, the time of the dense SVD includes forming the unfolding. Values above $1\times$ favor Algorithm~\ref{alg:component_extraction} (bold: at least $1.2\times$). Synthetic: the data of Section~\ref{sec:five} with noise at an SNR of 20\,dB; power law: $\sigma_i=1/i$; sparse: $p\,n^3$ observed entries of the synthetic tensor.}
\label{tab:crossover}
\centering
\scriptsize
\setlength{\tabcolsep}{3pt}
\begin{tabular}{@{}lrrrrrrrr@{}}
\toprule
 & Dense SVD & \multicolumn{7}{c}{Dense SVD time / Algorithm~1 time} \\
\cmidrule(lr){3-9}
Operator ($n=500$) & (ms) & $s=1$ & $5$ & $25$ & $50$ & $125$ & $250$ & $450$ \\
\midrule
Explicit, synthetic & 333 & \textbf{11.12$\times$} & \textbf{7.41$\times$} & 0.22$\times$ & 0.06$\times$ & 0.03$\times$ & 0.02$\times$ & 0.04$\times$ \\
Explicit, power law & 337 & \textbf{8.58$\times$} & \textbf{4.74$\times$} & \textbf{1.57$\times$} & 0.75$\times$ & 0.22$\times$ & 0.04$\times$ & 0.06$\times$ \\
Copy-free, synthetic & 334 & \textbf{12.70$\times$} & \textbf{8.57$\times$} & 0.25$\times$ & 0.07$\times$ & 0.03$\times$ & 0.02$\times$ & 0.04$\times$ \\
Copy-free, power law & 338 & \textbf{9.89$\times$} & \textbf{5.48$\times$} & 0.48$\times$ & 0.90$\times$ & 0.24$\times$ & 0.06$\times$ & 0.06$\times$ \\
Sparse, $p=0.05$ & 329 & \textbf{7.88$\times$} & \textbf{2.30$\times$} & 0.65$\times$ & 0.32$\times$ & 0.11$\times$ & 0.03$\times$ & 0.06$\times$ \\
Sparse, $p=0.2$ & 359 & \textbf{4.36$\times$} & 1.16$\times$ & 0.35$\times$ & 0.15$\times$ & 0.06$\times$ & 0.03$\times$ & 0.05$\times$ \\
\bottomrule
\end{tabular}

\end{table}

\section{Testing the Map on \mxNestCap{} Further Estimators}
\label{sec:appendix_mapext}

We tested the map on \mxNest{} further estimators in $\mathfrak{T}$. Before measuring, we fixed a prediction for them and a rule to score it, and we report the outcome as it came out. The prediction comes from the map that we had drawn from the five estimators: in $\mathfrak{T}$, ours is lighter, and if an update also needs few singular triplets on a large problem, ours is faster. The estimators are HOOI with imputation (hard-impute; public code: TensorLy with a mask), TIHT \citep{rauhut2017low} (no public code; compared with Raw), RGrad \citep{wang2023implicit} and PRGD \citep{zhang2025preconditioned} (ports of the PRGD-Tucker code), SiLRTC \citep{liu2012tensor}, HOOI on sparse observations \citep{xia2021statistically}, and the overlapped and the latent trace-norm ADMM \citep{tomioka2011statistical, tomioka2013convex}. Robust tensor PCA \citep{goldfarb2014robust}, FaLRTC \citep{liu2012tensor} and TMac \citep{xu2015parallel} lie outside $\mathfrak{T}$, so we ran them only in their public form. We follow the protocol of Section~\ref{sec:five} (order 3, $r=5$, $n\in\{100,200,300,500\}$, $p\in\{0.05,0.2\}$, 5 seeds, $T=10$) on an A100-SXM4-80GB. The other experiments used the 40\,GiB card; peak memory does not depend on the card. We calibrate the step sizes of RGrad and PRGD and the threshold of SiLRTC in the same way as $\lambda$ for E0. The ADMM methods use the defaults of the public code. All \mxRuns{} runs completed. The final errors of ours agree with those of the same-iterate implementations to a relative $\mxExactRel$; where TIHT recovers the tensor to below $10^{-6}$, they agree to $\mxExactAbs$ in absolute terms.

\emph{Rule, fixed before the measurements.} The rule scores two predictions at $n=500$. (a) Ours needs at least $1.2\times$ less data-inclusive peak memory than every same-iterate implementation. (b) Ours is at least $1.2\times$ faster (by more than two standard errors) than the fastest of them where we expected an update to need few singular triplets, and not faster where we expected $s/n>0.5$. The map counts as supported if (a) holds in all \mxScored{} cases and at least 80\% of the speed predictions are correct. It counts as rejected if (a) fails once or if fewer than 60\% of them are correct.

\emph{Outcome: rejected} (Table~\ref{tab:mapext}). Condition (a) fails in \mxAfail{} of \mxScored{} cases, and \mxHits{} of \mxScored{} speed predictions are correct.

\emph{What held.} The conditions that rule out a gain held. In a count over all sizes made after the test, ours is never faster where $n=100$ or $s/n>0.5$ (\mxNoGainFast{} of \mxNoGainCells{} cases). In contrast, in the same count, it is faster in \mxGainFast{} of the \mxGainCells{} other cases with sparse-plus-factored states. With $s/n=\mxHighQqLo$--$\mxHighQqHi$ (the ADMM methods), ours is $\mxHighQLo$--$\mxHighQHi\times$ slower, as Appendix~\ref{sec:appendix_crossover} predicts.

\emph{Why the rest failed.} Memory: at its peak, ours needs about \mxWordsObsExt{} words per observed entry (Appendix~\ref{sec:appendix_complexity_mem}), which is about three ambient arrays at $p=0.2$. Therefore, ours is lighter than comparators that hold several dense arrays ($\mxSparseMemLo$--$\mxSparseMemHi\times$ at $p=0.05$ for the \mxSparseN{} estimators with sparse-plus-factored states). However, it is not lighter than HOOI on sparse observations, whose implementations hold one or two dense arrays ($\mxEtenMem\times$ at $p=0.2$), and not where the state stays ambient (the dual variables of the ADMM methods; SiLRTC). Time: where the public code already avoids large SVDs (RGrad and PRGD, which only decompose a core of size $2r$), ours is as fast at $p=0.05$ and $\mxRgSlowLo$--$\mxRgSlowHi\times$ slower at $p=0.2$. We attribute this slowdown to the sparse gathers over $\Omega$, which cost more per entry on a GPU than dense passes over the tensor. For PRGD, a preconditioned method (Corollary~\ref{cor:guarantee_inheritance}), ours is $\mxPrgdMemLo$--$\mxPrgdMemHi\times$ lighter but not faster. In contrast, ours is faster where the comparator forms what ours avoids: TIHT ($\mxTihtLo$--$\mxTihtHi\times$ against the HOSVD of Raw), and HOOI with imputation and on sparse observations at $p=0.05$. For SiLRTC, $s/n=\mxSiqLo$--$\mxSiqHi$ lies within the threshold $s/n\le0.1$ fixed for the test, but beyond the crossover of Appendix~\ref{sec:appendix_crossover}. For the ADMM methods, the prediction at $p=0.2$ assumed low-rank iterates, as for HaLRTC. However, with the default parameters, their iterates stay nearly full rank, and they do not recover the tensor in 10 iterations (relative error $\mxAdmmErrLo$--$\mxAdmmErrHi$). Therefore, Figure~\ref{fig:map} states only necessary conditions, including those found here.

\begin{table}[H]
\caption{Test of the map on \mxNest{} further estimators at $n=500$ (A100-SXM4-80GB, mean of 5 seeds). Memory: data-inclusive peak of the lightest same-iterate implementation divided by that of ours (bold: below $1.2\times$, so condition (a) fails). Time: per-iteration time of the fastest same-iterate implementation divided by that of ours ($\uparrow$ faster, $=$ no difference, $\downarrow$ slower). $s/n$: fraction of singular triplets that an update uses (--: only small explicit matrices are decomposed). The predictions were fixed before the measurements. TIHT has no public code (comparator: Raw).}
\label{tab:mapext}
\centering
\scriptsize
\setlength{\tabcolsep}{4pt}
\begin{tabular}{@{}llrrrll@{}}
\toprule
Estimator & $p$ & $s/n$ & Memory & Time & Prediction & Correct \\
\midrule
HOOI with imputation & 0.05 & -- & 5.78$\times$ & 2.46$\times$ $\uparrow$ & faster & yes \\
 & 0.2 & -- & 1.65$\times$ & 1.21$\times$ $\uparrow$ & faster & yes \\
TIHT & 0.05 & 0.01 & 5.25$\times$ & 13.81$\times$ $\uparrow$ & faster & yes \\
 & 0.2 & 0.01 & 2.16$\times$ & 7.89$\times$ $\uparrow$ & faster & yes \\
RGrad & 0.05 & -- & 4.31$\times$ & 1.03$\times$ $=$ & faster & no \\
 & 0.2 & -- & 1.33$\times$ & 0.54$\times$ $\downarrow$ & faster & no \\
PRGD & 0.05 & -- & 5.78$\times$ & 1.06$\times$ $=$ & faster & no \\
 & 0.2 & -- & 1.66$\times$ & 0.39$\times$ $\downarrow$ & faster & no \\
SiLRTC & 0.05 & 0.07 & 1.22$\times$ & 0.08$\times$ $\downarrow$ & not faster & yes \\
 & 0.2 & 0.09 & \textbf{0.97$\times$} & 0.14$\times$ $\downarrow$ & faster & no, (a) fails \\
HOOI on sparse observations & 0.05 & -- & 2.92$\times$ & 1.77$\times$ $\uparrow$ & faster & yes \\
 & 0.2 & -- & \textbf{0.93$\times$} & 0.95$\times$ $=$ & faster & no, (a) fails \\
Overlapped trace-norm ADMM & 0.05 & 1.00 & \textbf{1.16$\times$} & 0.02$\times$ $\downarrow$ & not faster & yes, (a) fails \\
 & 0.2 & 1.00 & \textbf{1.15$\times$} & 0.02$\times$ $\downarrow$ & faster & no, (a) fails \\
Latent trace-norm ADMM & 0.05 & 0.81 & \textbf{0.77$\times$} & 0.02$\times$ $\downarrow$ & not faster & yes, (a) fails \\
 & 0.2 & 0.83 & \textbf{0.68$\times$} & 0.02$\times$ $\downarrow$ & faster & no, (a) fails \\
\bottomrule
\end{tabular}

\end{table}

\end{document}